\documentclass[12pt]{amsart}
\usepackage{amsbsy,amssymb,amsmath,amsthm,amscd,amsfonts,latexsym,amstext,delarray,
amsmath,graphicx} 
\usepackage{qtree}
\usepackage[margin=.8in]{geometry}
\usepackage{color}

\numberwithin{equation}{section}

\def\F{{\mathbb F}}

\def\R{{\mathbb R}}
\def\C{{\mathbb C}}

\def\P{{\mathbb P}}
\def\A{{\mathbb A}}

\def\cD{{\mathcal D}}

\def\cI{{\mathcal I}}

\def\cL{{\mathcal L}}

\def\cS{{\mathcal S}}

\title[Algebraic Geometry of Indo-European Languages]{Phylogenetics of Indo-European Language families via an Algebro-Geometric Analysis of their Syntactic Structures}
\author[K.Shu, A.Ortegaray, R.C.Berwick, M.Marcolli]{Kevin Shu, Andrew Ortegaray, Robert C.~Berwick and Matilde Marcolli}

\address{California Institute of Technology \\ USA}
\email{kshu@caltech.edu}
\email{aortegar@caltech.edu}
\address{Massachusetts Institute of Technology \\ USA}
\email{berwick@csail.mit.edu}
\address{California Institute of Technology \\ USA \newline \indent
Perimeter Institute for Theoretical Physics \\ Canada \newline \indent
University of Toronto \\ Canada}
\email{matilde@caltech.edu}
\email{matilde@math.utoronto.ca}
\date{}

\begin{document}
\maketitle

\begin{abstract}
Using Phylogenetic Algebraic Geometry, we analyze computationally the phylogenetic tree of subfamilies
of the Indo-European language family, using data of syntactic structures. The two main sources of syntactic data are
the SSWL database and Longobardi's recent data of syntactic parameters.  We compute phylogenetic
invariants and estimates of the Euclidean distance functions for two sets of Germanic languages, a set of Romance languages,
a set of Slavic languages and a set of early Indo-European languages, 
and we compare the results with what is known through historical linguistics.
\end{abstract} 

\section{Introduction}

The use of commutative algebra and algebraic geometry in the study of phylogenetic
trees and networks was developed in recent years in the context of biological applications,
see \cite{PaSturm}, \cite{PaSturm2}.  We argue in this paper that these methods 
have advantages over the other methods of phylogenetic reconstruction, such as
Hamming distance and neighborhood joining, when applied to the computational study
of phylogenetic trees of world languages based on syntactic data. Computational
studies of phylogenetics in Linguistics have been carried out recently in \cite{Barba}, \cite{Warnow},
using lexical and morphological data and in \cite{LongGua}, \cite{LongGua2} using
syntactic data. 

\smallskip

The main advantages of the algebro-geometric approach presented here can be
summarized as follows.

\begin{enumerate}
\item The use of Phylogenetic Algebraic Geometry to select a best candidate tree avoids some
of the well known possible problems (see Chapter~5 of \cite{Warnow-book}) that can occur in 
phylogenetic reconstructions based on Hamming distance and neighborhood-joining
methods. While such methods were used successfully in phylogenetic inference using
syntactic data in \cite{LongGua} and \cite{LongGua2}, we argue that the geometric methods
provide additional useful information, as explained below. 

\smallskip

\item Phylogenetic Algebraic Geometry associates an actual {\em geometric object} 
to a best candidate phylogenetic tree $T$, together with a boundary probability distribution 
at the leaves $P=(p_{i_1\ldots i_n})$ derived from the data. This geometric object consists
of a pair $(V_T, x_{T,P})$ of an algebraic variety $V_T$, which depends on the tree
topology, and a point $x_{T,P}\in V_T$ on it, which depends on both the tree $T$ and
the boundary distribution $P$. Unlike what happens with other phylogenetic methods 
that only provide a best candidate tree $T$, the geometry $(V_T, x_{T,P})$ contains
more information: the position of the point $P$ on the variety $V_T$ encodes information
about the distribution of the binary syntactic features across the language family. For
example, one can have different language families with topologically equivalent phylogenetic
trees. In this case one obtains two different points on the same variety $V_T$ whose
relative positions encode in a quantitative geometric way the difference between how 
the evolution of syntactic feature happened historically in the two families. 

\smallskip

\item The point $x_{T,P}$ is constrained to lie on the locus of real points $V_T(\R)$ of the
complex algebraic variety $V_T$, and in particular on the sublocus $V_T(\R_+)$ of
nonnegative real coordinates, since it is defined by a probability distribution. In several
cases, especially when analyzing sufficiently small trees, $V_T$ turns out to be
a classical and well studied algebraic variety, as in the case of the Secant varieties
of Segre embeddings of products of projective spaces that we encounter in this paper. 
In such cases, there are usually well understood and interesting geometric subvarieties
of $V_T$ and one can gain further insight by understanding when the point $x_{T,P}$
lies on some of these subvarieties, in addition to being contained in the real locus. 
For example, this may suggest compatibility of the boundary distribution $P$ with
respect to certain splitting of the tree into subfamilies and subtrees, which may
provide additional information on the underlying historical linguistics.

\smallskip

\item The algebro-geometric method is compatible with admixtures and with
phylogenetic networks that are not necessarily trees. The algebraic varieties involved
in this setting are different from the phylogenetic varieties of trees $V_T$ discussed
here, but they are analyzed with a similar method. Results on topological analysis of
data of syntactic structures (see \cite{PortMa1}) indicate the presence of nontrivial first
homology in certain language families. This can be seen as supporting evidence for
the use of networks that are not trees for phylogenetic analysis. The algebro-geometric 
formalism necessary to the discussion of more general phylogenetic networks
is discussed in \cite{PaSturm3} and \cite{Cart}.
 \end{enumerate}
 
\medskip
\subsection{Binary variables and syntactic structure}

The idea that the possible syntactic structure of human languages is
governed by certain basic binary variables, or syntactic parameters,
is one of the fundamental ideas underlying the Principles and
Parameters model in linguistics, originally introduced by Chomsky
\cite{Chomsky}, \cite{ChoLa}. The notion of syntactic parameter
underwent successive theoretical reformulation in the context of more
recent minimalist models \cite{Chomsky2}, but the main underlying
conceptual idea remains unchanged. A recent detailed overview of the
state of ongoing research in comparative generative grammar on the
topic of syntactic parameters can be found in the collection of papers
in the volume \cite{LingAn}.  An introduction to syntactic parameters
aimed at a general audience with no prior linguistics background is
given in \cite{Baker}.
 
\smallskip

Interesting questions regarding syntactic parameters include
identifying a minimal set of independent variable completely
determining a language's syntax and obtaining an explicit and complete
description of the dependencies that exist among the known
parameters. A rough analogy is that the set of syntactic parameters
forms a kind of ``basis set'' spanning the space of possible human
languages (alternatively, grammars, since were are attempting to
describe language structure).  Each choice of values for the
parameters in this basis set fixes a distinct possible (presumably
learnable) human language.  Typically, it is assumed that the
parameter values can be learned from data available from positive
example sentences presented to a language learner (i.e., a child).

From a more precise mathematical perspective one can view this as the
question of identifying the correct ``manifold of syntax'' inside a
large ambient space of binary variables.  These binary variables
describing syntactic structures can roughly be thought of as yes/no
answers to questions about whether certain constructions are possible
in a given language or not. For a more precise description of
parameters as instructions for triggering syntactic operations see
\cite{Rizzi}.

\smallskip

There are two existing databases of syntactic structures of world languages that we use
in this paper: the SSWL database \cite{SSWL} and the data of syntactic parameters
collected by Giuseppe Longobardi and the LanGeLin collaboration. The binary variables
recorded in the SSWL database should not be regarded, from the linguistics perspective
as genuine syntactic parameters, although they still provide a very useful collection of
binary variables describing different features of syntactic structures of world languages.
The variables recorded in the SSWL database include a set of $22$ binary variables
describing word order properties, {\em 01}--Subject Verb,$\ldots$, {\em 22}--Noun Pronomial Possessor,
a set of $4$ binary variables {\em A01--A04} describing relations of adjectives to nouns and degree words,
a variable {\em AuxSel01} about the selection of auxiliary verbs, variables {\em C01--C04} 
still related to word order properties on complementarizer and clause and adverbial subordinator
and clause, {\em N201--N211} variables on properties of numerals, 
{\em Neg01--Neg14} variables on negation, 
{\em OrderN301--OrderN312} on word order properties involving demostratives, adjectives, nouns, and
numerals,  {\em Q01--Q15} regarding the structure of questions, {\em Q16Nega--Q18Nega} and
{\em Q19NegQ--Q22NegQ} on answers to negative questions, {\em V201-V202} on declarative and interrogative Verb-Second,  {\em w01a--w01c}  indefinite mass nouns in object position, {\em w02a--w02c} definite mass nouns in object position, {\em w03a--w03d} indefinite singular count nouns in object position, {\em w04a--w04c} definite singular count nouns in object position, {\em w05a--w05c} indefinite
plural count nouns in object position, {\em w06a--w06c} definite plural count nouns in object position,
{\em w07a--w07d} nouns with (intrinsically) unique referents in object position, {\em w08a--w08d}
proper names in object position, {\em w09a--w09b} order of article and proper names in object position,
{\em w10a--w10c} proper names modified by an adjective in object position, {\em w11a--w11b}
order of proper names and adjectives in object position, {\em w12a--w12f} order of definite articles 
and nouns in object position, {\em w20a--w20e} singular count nouns in vocative phrases, {\em w21a--w21e} proper nouns in vocative phrases, {\em w22a--w22e} plural nouns in vocative phrases. 
A detailed description of each of these binary variables can be found on the online site of
the SSWL database, \cite{SSWL}. While these are
certainly not considered to be an exhaustive list of binary variables associated to syntax, they contain 
a considerable amount of information on the variability of syntactic structures across languages. 

\smallskip

The LanGeLin data of Longobardi record a different set of syntactic features, which are
independent of the SSWL data. These variables should be regarded as genuine syntactic parameters
and are based on the general Modularized Global Parameterization
approach developed by Longobardi \cite{Long}, \cite{Long3}, that considers reasonably large
sets of parameters within a single module of grammar, and their expression across a large
number of languages. The LanGeLin data presented in \cite{Long} that we use here include
91 parameters affecting the Determiner Phrases structure. The full list of the LanGeLin
syntactic parameters used in this paper is reported in Appendix D, reproduced from
Appendix A of \cite{Kazakov}. 

\smallskip

 Unlike the SSWL data, which do not record any explicit relations between
 the variables, many explicit relations between the Longobardi
 syntactic parameters are recorded in the LanGeLin data. A more detailed
 analysis of the relations in the LanGeLin data is given in \cite{Kazakov} 
 and in \cite{OBM}. In our analysis here we
 have removed those parameters in the LanGeLin data that are explicitly 
 dependent upon the configuration of other parameters.

\medskip
\subsection{Related Work}

A long-standing, familiar approach to linguistic
phylogenetics is grounded on the use of lexical (including phonemic)
features; see, e.g., \cite{Warnow} for a survey of phylogenetic methods
applying such features on a carefully analyzed Indo-European dataset.
More recently, other researchers have suggested alternatives to bypass
issues with lexical items, such as the non-treelike behavior of lexical
diffusion, sometimes rapid and different time scales for lexical
change, and the like.  For example, Murawaki \cite{Murawaki} used
linguistic typological dependencies such as word order (OV vs. VO, in
the Greenbergian sense) or
grammar type (synthetic vs. analytic), in order to build phylogenies
over longer time scales and across widely different languages.
Murawaki's approach computes latent components from linguistic typological
features in the {\em World Atlas of Languages}, (WALS) and then feeds
these into phylogenetic analysis.  Longobardi
and colleagues have pursued a detailed linguistically-based analysis
of, e.g., Noun Phrases (so-called Determiner structure) across many
different Western European languages to develop a fine-grained explicit
parametric analysis of what distinguishes each of these languages from
the others,
see \cite{LongGua} and subsequent work including the more recent \cite{evolang11}.
In effect, this is a ``hand-tooled'' version of a
statistical, principal-components like approach.   They have used
Jacquard distance metrics as the measure to feed into conventional
distance-based phylogenetic programs.   
The approach presented in the
current work differs from either of these and from other more familiar
phylogenetic methods applied to linguistic datasets (such as maximum
likelihood or Bayesian approaches) in that it adopts a
different approach to the structure of the phylogenetic space itself,
rather than relying on conventional methods, while retaining the
non-lexical, typological information as the basis for describing the
differences among languages. 

\medskip
\subsection{Comments on the data sets}

The two databases used in our analysis, namely the SSWL database \cite{SSWL}
and the recent set of data published by Longobardi and collaborators \cite{Long},
are currently the only existing extensive databases of syntactic structures of world languages.
Therefore any computational analysis of syntax necessarily has to consider these data.

In the process of evaluating phylogenetic trees via the algebro-geometric method,
we also perform a comparative  analysis of the two databases of syntactic
variables that we use. As the extended version of the Longobardi dataset has only
recently become available \cite{Long}, a comparative analysis of this dataset has 
not been previously considered, so the one reported here is novel.  
Other methods of comparative analysis of these two databases of
syntactic structures will be discussed elsewhere. In the cases
analyzed here we see specific examples (such as the
second set of Germanic languages we discuss) where Longobardi's
database appears to be more reliable for phylogenetic reconstructions than the 
SSWL data, even though the latter dataset is larger. 

\medskip
\subsection{Phylogenetics and syntactic data}

The use of syntactic data for phylogenetic reconstruction of language families was developed in previous
work of Longobardi and collaborators, \cite{LongGua}, \cite{LongGua2}, see also \cite{Long2}, \cite{Long3}. 
Computational phylogenetic reconstructions of language family trees based on lexical and morphological 
data were also obtained in \cite{Barba}, \cite{Warnow}. It is well known that the use of lexical data, in the 
form of Swadesh lists, is subject to issues related to synonyms, loan words, and false positives, that may 
affect the measure of proximity between languages. Morphological information is much more
robust, but its encoding into binary data is not always straightforward. Syntactic data, on the other
hand, are usually classified in terms of binary variables (syntactic parameters), and provide a
robust information about language structure. Thus, we believe that syntactic data should be
especially suitable for the use of computational methods in historical linguistics.

\smallskip

In \cite{ShuMa} it was shown that, when using syntactic data of the SSWL database \cite{SSWL} 
with Hamming distances and neighborhood joining methods to construct linguistic phylogenetic trees, several
kinds of errors typically occur. These are mostly due to a combination of two
main factors: 
\begin{itemize}
\item the fact that at present the SSWL data are very non-uniformly mapped across languages;
\item errors
propagated by the use of neighborhood-joining algorithms based on the Hamming distance between the strings of
syntactic variables recorded in the SSWL data. 
\end{itemize}
An additional source of problems is linguistic in nature, namely the
existence of languages lying in historically unrelated families that can have greater similarity than expected at the
level of their syntactic structures. Another possible source of problems is due to the structure of the SSWL database 
itself, where the syntactic binary variable recorded are not what linguists would consider to be actual syntactic
parameter in the sense of the Principles and Parameters model \cite{Chomsky}, \cite{ChoLa}, see also \cite{Rizzi}:
there are conflations of deep and surface structures that make certain subsets of 
the syntactic variables of the SSWL data potentially
problematic from the linguistic perspective. However, it was also shown in \cite{ShuMa} that several of these
problems that occur in a naive use of computational phylogenetic methods can be avoided by a more careful analysis.
Namely, some preliminary evidence is given in \cite{ShuMa} that, when a naive
phylogenetic reconstruction applied simultaneously 
to the entire SSWL database is replaced by a more careful analysis applied to smaller groups of languages
that are more uniformly mapped in the database, the phylogenetic invariants of Phylogenetic Algebraic Geometry 
can identify the correct phylogenetic tree, despite the imperfect nature of the SSWL data. 
The method of Phylogenetic Algebraic Geometry that we refer to here 
was developed in \cite{PaSturm}, \cite{PaSturm2} for applications to
mathematical biology, see also a short survey in \cite{Bocci}.  

\smallskip

In the present paper we focus on certain subfamilies of the Indo-European language family,
in particular the Germanic languages, the Romance languages, and the Slavic languages.
We apply the Phylogenetic Algebraic Geometry method, by computing the phylogenetic
invariants for candidate trees, and the Euclidean distance function. 
We compare the results obtained by applying this method to 
the SSWL data and to a more recent set of data of syntactic parameters collected by 
Longobardi \cite{Long}, which are a largely extended version of the data previously 
available in \cite{LongGua}. 

\smallskip

We list here the specific historical linguistics settings that we analyze in this paper.

\smallskip
\subsection{The Germanic family tree}

We consider the following two sets of Germanic languages:
\begin{enumerate}
\item $\cS_1(G)=\{$ Dutch, German, English, Faroese, Icelandic, Swedish $\}$
\item $\cS_2(G)=\{$ Norwegian, Danish, Icelandic, German, English, Gothic, Old English $\}$.
\end{enumerate}
The first one only consists of modern languages, while in the second one we have included the data
of the two ancient languages Gothic and Old English.
We analyze the first set $\cS_1(G)$ with the SSWL data, and we analyze the second set
first using the new Longobardi data and then using the SSWL data. In both cases we first
generate candidate trees using the software package PHYLIP \cite{PHYL}, then using the
Phylogenetic Algebraic Geometry method we compute the phylogenetic invariants and an estimate of the 
Euclidean distance function for these candidate trees and we select
the best candidate. 

\smallskip

For sufficiently small trees one can expect that other 
methods, including more 
conventional Bayesian analysis, would be able to identify 
the correct candidate tree. However, we see here in specific
examples that the algebro-geometric 
method performs at least better than standard phylogenetic
packages like PHYLIP when applied to the same data.

\smallskip

Given the large number of alternative phylogenetic methods, why use
PHYLIP as a baseline?  There are two main reasons. 
First of all, PHYLIP is selected here as an example of a well known 
and widely used phylogenetic package, hence it is an easy baseline 
for comparison. Moreover, we use 
PHYLIP to preselect a set of candidate trees because likewise parsimony 
method is a standard starting point for Bayesian analysis.
Maximum likelihood inference is generally regarded as a more reliable
method. However, it is worth pointing out here that a form of likelihood 
evaluation is already built into the algebro-geometric method.  Indeed, the
Euclidean distance function is a maximum likelihood estimation,
as shown in \cite{DHOST}. A maximum likelihood degree, which
counts the critical points of the likelihood function on determinantal
varieties, can also be computed, see \cite{MaxLike}, 
but only in sufficiently small cases. 
In this sense then, the method already encompasses a wide variety of
the current classes of phylogenetic approaches.

\smallskip

We show that, for the set $\cS_1(G)$, the phylogenetic invariants suggest the correct tree among the six candidates 
generated by PHYLIP, which is confirmed by a form of likelihood computation achieved via the
computation of the Euclidean distance. The topology of this tree correctly corresponds to 
the known historical subdivision 
of the Germanic languages into West Germanic and North Germanic and the relative proximity 
of the given languages within these subtrees.
In this sense the algebro-geometric method applied to a baseline
dataset can be confirmed, always a key step in advancing a novel
phylogenetic approach as \cite{Warnow}  note.

\smallskip

For the other set $\cS_2(G)$ of seven languages, which are common to both databases, 
we also find that the phylogenetic invariants computed on a subset of the Longobardi syntactic data 
point to the correct best candidate tree, which is confirmed by a lower bound estimate 
of the Euclidean distance. With the SSWL data the phylogenetic invariants computed with
respect to the $\ell^1$ norm still identify the historically correct tree as the best candidate,
but not when computed with respect to the $\ell^\infty$ norm. This confirms in our setting
a general observation of \cite{Casa} on te better reliability of the $\ell^1$ norm in the
computation of phylogenetic invariants. We see here an example where the lower bound
on the Euclidean distance correctly excludes some of the candidates, but fails to assign
the smallest lower bound to the best tree. 
This different behavior of the Longobardi and the
SSWL data on this set of languages presumably reflects the presence of a large
number of dependencies in the SSWL variables.

\smallskip

In the last section of the paper we discuss a possible issue of the direct application
of this algebraic phylogenetic method to syntax, which is caused by neglecting
relations between syntactic parameters and treating them, in this model, like
independent random variables. We suggest possible ways to correct for these
discrepancies, which will be analyzed in future work. We expect that such 
discrepancies may be resolved by a better approach taking syntactic relations into account. 

\smallskip
\subsection{The Romance family tree}

The case of the Romance languages is an interesting example of the limitations of
these methods of phylogenetic reconstructions. We considered as set of
languages Latin, Romanian, Italian, French, Spanish, and Portugues, 
and we used a combination of the SSWL and the Longobardi data, which
are independent sets of data. We find
that PHYLIP produces a unique candidate tree, which is however not the one that
is considered historically correct. We compute the phylogenetic invariants
and the Euclidean distance for both the PHYLIP tree and the historically correct tree. 
The phylogenetic invariants computed with respect to the $\ell^1$ norm
identify the historically correct tree as the favorite candidate, while they do
not give useful information when computed in the $\ell^\infty$ norm. The
estimate of the Euclidean distance also favors the historically correct tree
over the PHYLIP candidate tree.

\smallskip
\subsection{The Slavic family tree}

We also analyze with the same method the phylogenetic tree of a group of Slavic languages for which
we use a combination of SSWL data and the data of \cite{LongGua}: Russian, Polish, Slovenian, Serb-Croatian, Bulgarian.
For this set of languages, PHYLIP applied to the combined syntactic data produces five candidate trees with
inequivalent topologies. Using the phylogenetic invariants computed with the $\ell^1$ norm we
identify the historically correct tree as the best candidate, while the computation in the $\ell^\infty$
norm does not select a unique best candidate. The lower bound estimate of the Euclidean distance
also correctly selects the linguistically accurate tree.

\smallskip
\subsection{The early Indo-European branchings and the Indo-European controversy} 

The use of computational methods in historical linguistics has been the focus of
considerable attention, and controversy, in recent years, due to claims made in
the papers \cite{Gray2}, \cite{Gray} regarding the phylogenetic tree of the Indo-European languages,
based on a computational analysis of trees obtained from distances between binary
data based on lexical lists and cognate words. While this method of computational
analysis of language families has been considered in various contexts (see  \cite{Fos}
for a collection of contributions), the result announced in \cite{Gray2}, \cite{Gray} appeared to
contradict several results obtained by historical linguists by other methods, hence
the ensuing controversy, see \cite{PerLe}. For comparison,
a different reconstruction of the Indo-European tree, carried out by computational methods 
that incorporate lexical, phonological, and morphological data, was obtained by
Ringe, Warnow, and Taylor \cite{RWT}.
Neither of these computational analysis makes any use of syntactic data about the
Indo-European languages. 

\smallskip

We focus here on some specific issues that occur in the phylogenetic tree of
\cite{Gray} compared with that of \cite{RWT}: 
\begin{itemize}
\item The relative positions of the Greco-Armenian subtrees;
\item The position of Albanian in the tree;
\item The relative positions of these languages with respect to the Anatolian-Tocharian subtrees.
\end{itemize}
This means that we neglect several other branches of the Indo-European tree
analyzed in \cite{Gray} and in \cite{RWT} and we focus on a five-leaf binary tree
with leaves corresponding to the languages: Hittite, Tocharian, Albanian, Armenian, and Greek.
We will consider the tree topologies for this subset of languages resulting from the
trees of \cite{Gray} and \cite{RWT} and we will select between them on the basis of
Phylogenetic Algebraic Geometry.  

\smallskip

The set of languages considered here 
(Hittite, Tocharian, Albanian, Armenian, Greek) are listed in the SSWL database \cite{SSWL},
while not all of them are present in the Longobardi data \cite{Long}. Thus, in this case
we have to base our analysis on the SSWL data. With the exception of Armenian and Greek, which
are extensively mapped in the database, the remaining
languages (especially Tocharian and Hittite) are very poorly mapped, and the set of parameters 
that are completely mapped for all of them is very small, hence the resulting analysis should not be
considered very reliable, due to this significant problem. 

\smallskip

Nonetheless, we compute the phylogenetic invariants for the Gray-Atkins tree and for the 
Ringe--Warnow--Taylor tree and we also compute the Euclidean
distance function to the relevant phylogenetic algebraic variety. We find that, while
the evaluation of the phylogenetic invariants with the $\ell^\infty$ norm does not
give useful information, the evaluation in the $\ell^1$ norm favors the 
 linguistically more accurate Ringe--Warnow--Taylor tree. Similarly the estiimate
of the Euclidean distance selects the same Ringe--Warnow--Taylor tree.

\smallskip

The Gray-Atkins tree is {\em not} the one generally agreed upon by linguists, while the
Ringe--Warnow--Taylor tree is considered linguistically more reliable. A more recent 
discussion of the early Indo-European tree, which is also considered linguistically very reliable, 
can be found in \cite{AnthonyRinge}.  However, 
the part of the tree of \cite{AnthonyRinge} that we focus on here agrees with the one of \cite{Warnow}
(though the position of Albanian is not explicitly discussed in \cite{AnthonyRinge}),
hence we refer to \cite{Warnow} in our analysis.

\section{Phylogenetic Algebraic Varieties and Invariants}

Before we proceed to the analysis of the two sets of languages listed above, 
we recall briefly the notation and the results we will be using from Phylogenetic
Algebraic Geometry, see \cite{AllRho}, \cite{PaSturm}, \cite{PaSturm2}.  We also
discuss the limits of the applicability of this method to syntactic data of
languages and some approaches to improve the method accordingly.

\smallskip

In order to apply the algebro-geometric approach, we think of each binary
syntactic variable as a dynamical variable governed by a Markov process on a 
binary tree. 
These binary Markov processes on trees generalize the Jukes--Cantor model,
in the sense that they do not necessarily assume a uniform distribution at the
root of the tree.
The model parameters $(\pi, M^e)$
consist of a probability distribution $(\pi, 1-\pi)$ at the
root vertex (the frequency of expression of the $0$ and $1$ values of the
syntactic binary variables at the root) and bistochastic transition matrices
$$ M^e=\begin{pmatrix} 1-p_e & p_e \\ p_e & 1-p_e \end{pmatrix} $$
along the edges.

\smallskip

For a binary tree with $n$ leaves,
the boundary distribution $P=(p_{i_1,\ldots,i_n})$ counts the frequencies of the
occurrences of binary vectors $(i_1,\ldots, i_n) \in \{ 0,1 \}^n$ of values of the 
binary syntactic variables for the languages $\{ \ell_1, \ldots, \ell_n \}$ 
at the leaves of the tree. If $N$ is the total number of syntactic binary variables
available in the database (counting only those that are completely mapped
for all the $n$ languages consdiered) and  $n_{i_1,\ldots,i_n}$ is the number
of occurrences of the binary vector $(i_1,\ldots, i_n)$ in the list of values of the
$N$ syntactic variables for these $n$ languages, then the frequencies in $P$
are given by $$p_{i_1,\ldots,i_n} = \frac{n_{i_1,\ldots,i_n}}{N}.$$
The boundary distribution is a polynomial function of the model parameters
\begin{equation}\label{Ppol}
 p_{i_1,\ldots, i_n}= \Phi(\pi,M^e)=\sum_{w_v\in \{0,1\}} \pi_{w_{v_r}} \prod_e M^e_{w_{s(e)},w_{t(e)}}, 
\end{equation} 
with a sum over ``histories", that is, paths in the tree. 
This determines a polynomial map of affine spaces
\begin{equation}\label{PhiT}
 \Phi_T: \A^{4n-5} \to \A^{2^n}, 
\end{equation} 
where $4n-5$ is the number of model parameters for a binary tree $T$ with $n$-leaves and
binary variables. Dually, the kernel of the map of polynomial rings
\begin{equation}\label{PsiT}
\Psi_T: \C[z_{i_1,\ldots,i_n}]\to \C[x_1,\ldots,x_{4n-5}]
\end{equation}
defines the phylogenetic ideal $\cI_T$. This corresponds geometrically to 
the phylogenetic algebraic variety $V_T$. 

\smallskip

It is proved in \cite{AllRho} that, for these Markov models on trees with binary variables 
that generalize the Jukes--Cantor model, 
the phylogenetic ideal $\cI_T$ is generated by all the $3\times 3$-minors
of all the {\em flattenings} of the tensor $P=(p_{i_1,\ldots,i_n})$. There is
one such flattening for each internal edge of the binary tree, where each 
internal edge corresponds to a subdivision of the leaves into a disjoint
union of two sets of cardinality $r$ and $n-r$. The flattening is a $2^r \times 2^{n-r}$
matrix defined by setting 
\begin{equation}\label{FlatTP}
 {\rm Flat}_{e,T}(P)(u,v) = P(u_1,\ldots, u_r, v_1, \ldots, v_{n-r}), 
\end{equation} 
where $P$ is the boundary distribution. The terminology corresponds to
the fact that an $n$-tensor $P$ is ``flattened" into a
collection of $2$-tensors (matrices). 

\smallskip

These generators of the phylogenetic ideal can then be used as a test for
the validity of a candidate phylogenetic tree. If the tree is a valid phylogenetic
reconstruction, then the boundary distribution $P=(p_{i_1,\ldots,i_n})$ should
be a zero of all the polynomials in the phylogenetic ideal (or very close to
being a zero, allowing for a small error margin). 

\smallskip

In the case of the binary Jukes--Cantor model, where one assumes a uniform
root distribution, there are additional invariants, as shown in \cite{StuSull}.
For the purpose of linguistic applications it is more natural to work with the
general binary Markov models described above, where the root distribution
$(\pi, 1-\pi)$ is not assumed to be uniform, than with the more restrictive
Jukes--Cantor model. Indeed, there is no reason to assume that  
parameters at the root of a language phylogenetic tree would
have equal frequency of expression of $0$ and $1$: the overall data on
all languages, ancient and modern, contained in the available database
show a clear prevalence of parameters that are expressed (value $1$) rather
than not. (This point was discussed in some detail in \cite{SiTaMa}.)

\smallskip
\subsection{Phylogenetic Invariants}

The generators $\phi_T$ of the phylogenetic ideal $\cI_T$ are given by the 
Allman--Rhodes theorem \cite{AllRho} by all the  $3\times 3$-minors $\det(M)$ 
of the flattening matrices ${\rm Flat}_{e,T}$. 

\smallskip

To every candidate tree, one can also associate a computation of a discrepancy
that measures how much the polynomials $\phi_T$ fail to vanish at the point $P$.
This can be done using different kinds of norms. Generally, one can use either
the $\ell^\infty$ norm and obtain an expression of the form
$$ \| \phi_T(P) \|_{\ell^\infty} =\max_{\stackrel{e\in E(T),}{M(P) \subset {\rm Flat}_{e,T}(P), 3\times 3-\text{minor}}} \,\, | \det(M(P)) |, $$
which we write equivalently in the following as
$$ \| \phi_T(P) \|_{\ell^\infty} =\max_{\phi \in 3\times 3-\text{minors of } {\rm Flat}_{e,T}(P)} |\phi(P)|, $$
where the expression $|\phi(P)|$ stands for the absolute value of the determinant of the $3\times 3$-minor.
It is also natural to use the $\ell^1$ norm and compute
$$ \| \phi_T(P) \|_{\ell^1} =\sum_{\stackrel{e\in E(T),}{M(P) \subset {\rm Flat}_{e,T}(P), 3\times 3-\text{minor}}} \,\, | \det(M(P)) |, $$
equivalently written in the rest of the paper as
$$ \| \phi_T(P) \|_{\ell^1} =\sum_{\phi \in 3\times 3-\text{minors of } {\rm Flat}_{e,T}(P)} |\phi(P)|. $$
One can expect that the $\ell^\infty$ norm will be a very weak invariant, because taking the 
maximum loses a lot of information contained in the phylogenetic invariants $\phi_T(P)$.
Indeed, this turns out to be the case. As analyzed in detail in \cite{Casa}, the $\ell^1$ norm is
a more refined and reliable way to identify best phylogenetic trees on the basis of the
computation of phylogenetic invariants than the $\ell^\infty$ norm. We will see several explicit
examples in the following sections where the $\ell^\infty$ norm does not provide useful
information to identify the correct candidate tree, while the $\ell^1$ norm of the phylogenetic
invariants correctly identifies the unique best candidate tree.

\smallskip

Once the best candidate tree is identifies, the value of these discrepancy measures
$\| \phi_T(P) \|_{\ell^\infty}$ and $\| \phi_T(P) \|_{\ell^1}$ for that tree (which is in general
small but non-zero) can be regarded as a 
possible quantitative measure of how much the observed distribution $P$ of the 
syntactic parameters for the languages at the leaves of
the candidate tree $T$ differ from a distribution obtained by the evolution of identically distributed independent
random variables evolving according to a Markov model on the tree. Since one of the important points we
wish to investigate is how relations between syntactic parameters affect their behavior as random variables
in dynamical models of language change and evolution, we will regard these quantities as one of the
numerical indicators of the discrepancy from the standard independent identically distributed (i.i.d.)
Markov model assumption. The presence
of dependencies between syntactic parameters is expected to cause deviations
from the dynamics of an actual i.i.d.~Markov model
While we do not analyze in the present paper how possible models of parameter
dependencies affect the dynamics and may be reflected in the value of such
invariants, we intend to return to this investigation in future work.

\smallskip
\subsection{Likelihood estimate: Euclidean distance}

As a way to compare different candidate trees and select the best possible candidate, one can use
the Euclidean distance, in an ambient affine space, between the point $P$ given by the
boundary distribution and the variety $V_T$ associated to the candidate tree $T$. 
The tree realizing the smallest distance will be the favorite candidate. 

\smallskip

As we discuss explicitly in \S\S \ref{Var1Sec}, \ref{Var2Sec},  the computation of
the distance to $V_T$ can be estimated in terms of distances of some of the flattening matrices of $T$ 
to certain Segre and Secant varieties, namely determinantal varieties of rank one and two. In some
particular case, like the first set of Germanic languages we analyze, the lower bound
estimate obtained in this way is sharp, under a conditional assumption, which we
discuss more in detail in \S \ref{CondSec} below.

\smallskip

We compute the Euclidean distances of the flattening matrices from the corresponding
determinantal varieties using the Eckart--Young theorem,
as shown in Example~2.3 of \cite{DHOST}, see also \cite{PaSturm}.

\smallskip

The Eckart--Young theorem describes a low-rank approximation problem, namely
minimizing the Euclidean distance
$\| M - M' \|$ between a given $n\times m$ matrix $M$, seen as a vector in $\R^{nm}$,
and an $n\times m$ matrix $M'$ with ${\rm rank}(M')\leq k$, for a given $k \leq n \leq m$. 
One considers the singular decomposition $M = U \Sigma V$ where $\Sigma$
is an $n\times m$ diagonal matrix $\Sigma = {\rm diag}(\sigma_1,\ldots,\sigma_n)$ and 
$\sigma_1 \geq \sigma_2 \geq \cdots \geq \sigma_n \geq 0$, and where $U$ and $V$
are, respectively $n\times n$ and $m\times m$ orthogonal matrices.
Then the minimum of the distance $\| M - M' \|$ is realized by 
$M'=U \Sigma' V$ where $\Sigma'={\rm diag}(\sigma_1,\ldots,\sigma_k, 0,\ldots, 0)$
with the distance given by
$$ \min_{M'} \| M - M' \| =( \sum_{i=k+1}^n \sigma_i^2 )^{1/2} . $$
This can equivalently be stated as the fact that the minimum distance between
a given $n\times m$ matrix $M$ and the determinantal variety $\cD_k(n,m)$
of $n\times m$ matrices of rank $\leq k$ is given by
\begin{equation}\label{EuclDist}
 {\rm dist}(M, \cD_k(n,m)) = \| (\sigma_{k+1},\ldots,\sigma_n) \|, 
\end{equation}
where the $\sigma_i$ are the singular values of $M$. The point $M'$
realizing the minimum is unique iff $\sigma_{k+1} \neq \sigma_k$, with $k$ the rank, \cite{Mirsky}.

\smallskip
\subsection{Conditional cases and distance estimates}\label{CondSec}

In the specific examples we discuss below, we usually consider a list of
pre-selected candidate trees, obtained via the use of the PHYLIP package
and among them we test for the most reliable candidate using the
algebro-geometric methods discussed here. Unlike the case where the 
search happens over all possible interpolating binary trees, in these cases
the pre-selected tree tend to all agree on certain proximity assignments of
some of the leaves. For example, in the first set of Germanic
languages that we discuss below, all the candidate trees agree on the
proximity of Dutch and German and on the proximity of Icelandic and Faroese,
though they disagree in the relative placements of these subtrees with 
respect to the other languages in the set. This agreement among the
candidate trees results in two of the flattening matrices being common
to all of the candidates. 

\smallskip

In a situation like this one it is reasonable to consider 
a ``conditional case" where we assume that the incidence 
condition that these common flattenings lie on the respective
determinantal varieties already holds. We then 
aim at identifying the best candidate tree
among those with these constraints already assumed.

\smallskip

In our implementation this means that, instead of estimating the
Euclidean distance of the point $P$ from certain intersections $V_k \cap W$ of 
subvarieties of the ambient projective space, and searching for a minimum
among the distances $\min_k {\rm dist}(P, V_k \cap W)$, we assume that
it is established that the point already lies on a certain subvariety, $P\in W$,
as the effect of the agreement of all the candidate trees on certain proximity
conditions, and we estimate the minimum $\min_k {\rm dist}(P, V_k)$.
While in general an estimate ${\rm dist}(P, V_1)< {\rm dist}(P, V_2)$ would obviously 
not imply that one also has ${\rm dist}(P, V_1\cap W)< {\rm dist}(P, V_2\cap W)$, if
the incidence $P\in W$ is known, then evaluating and minimizing ${\rm dist}(P, V_k)$ suffices.
We will see that this method provides reliable results in the cases we analyze.

\smallskip

This method provides an evaluation of the Euclidean distance
${\rm dist}(P,V_T)$ in the case of the first set of Germanic languages
that we analyze, since in that case two out of three flattenings are 
common to all trees. In the other cases we consider, where there isn't
so much common agreement between the candidate trees, we can
use a similar method, but this will only provide us with a rough lower bound
on the Euclidean distance ${\rm dist}(P,V_T)$. Indeed, we simply obtain
an estimate using the fact that the lower bound ${\rm dist}(P,V\cap W)
\geq \max\{ {\rm dist}(P,V), {\rm dist}(P,W) \}$, for two subvarieties $V,W$
in the same ambient space. Since this is only a lower bound, which is
in general not expected to be sharp, one can at best hope to use this
estimate to exclude candidates for which the computed 
$\max\{ {\rm dist}(P,V), {\rm dist}(P,W) \}$ is large (within the set
of given candidates), while a small value of this maximum will not
necessarily imply that the corresponding candidate is optimal as
${\rm dist}(P,V\cap W)$ could easily be significantly larger. We see
however that in many cases this lower bound suffices to exclude
most candidates hence it provides a useful estimate.

\smallskip

A more general theoretical discussion of these estimation methods
and their range of validity, compared to other phylogenetic invariants
and tree reconstruction algorithms (such as discussed in 
\cite{Casa}, \cite{Erik}, \cite{Rusinko}) will be discussed elsewhere,
separately from the present application,
since they are not restricted to the specific linguistic setting considered here.

\smallskip
\subsection{Limits of applicability to Syntax}

One of the purposes of this paper is also to better understand the limits of the
applicability of these phylogenetic models to syntactic data. One of the main
assumptions that need to be more carefully questioned is treating syntactic parameters 
as i.i.d.~random variables evolving under the same
Markov model on the tree.  We know that there are relations between syntactic
parameters. While the complete structure of the relations is not known, and is
in fact one of the crucial questions in the field, one can detect the presence of
relations through various computational methods applied to the available 
syntactic data. 

\smallskip

In \cite{Mar} and \cite{ShuMa2}, a quantitative test was devised, aimed at measuring
how the distribution of syntactic parameters over a group of languages differs from
the result of i.i.d.~random variables. Using coding theory,
one associates a binary code to the set of syntactic parameters of a given group
of languages and computes the position of the resulting code in the space
of code parameters (the relative rate of the code and its relative minimum distance).
If the distribution of the syntactic features across languages were the effect of
an evolution of identically distributed independent random variables, one would
expect to find the code points in the region of the space of code parameters
populated by random codes in the Shannon random code ensembles, that is,
in the region below the Gilbert--Varshamov curve. However, what one finds
(see  \cite{ShuMa2}) is the presence of many outliers that are not only above
the Gilbert--Varshamov curve, but even above the symptotic bound and the
Plotkin bound. This provides quantitative evidence for the fact that the evolutionary
process that leads to the boundary distribution $P$ of code parameters may
differ significantly from the hypothesis of the phylogenetic model. 

\smallskip

In \cite{ParkMa} it was shown, using Kanerva networks, that different
syntactic parameters in the SSWL database have different degrees of
recoverability, which can be seen as another numerical indicator of
the presence of relations, with parameters with lower
recoverability counting as closer to being truly independent
variables and those with higher recoverability seen as dependent 
variables. One possible modification of the evolutionary model on
the phylogenetic tree may then be obtained by computing the
observed distribution $P$ at the leaves, by introducing different
weights for the different parameters, which depend on the
recoverability factor, so that parameters that are more likely
to be independent variables would weight more in determining
the boundary distribution and parameters that have higher
recoverability, and are therefore considered dependent variables,
would contribute less to determining $P$. 

\smallskip

A further issue worth mentioning, though we will not discuss it
in this paper, is whether the hypothesis that the evolutionary
dynamics happens on a tree is the best model. There are 
more general phylogenetic reconstruction techniques 
based on graphs that are not trees, see \cite{Gus} and
the algebro-geometric models in \cite{Cart}. It was shown
in \cite{PortMa1} that the persistent topology of the SSWL data 
of some language families (the Indo-European) contain non-trivial
persistent generators of the $H_1$ homology group. 
While the persistent generators of $H_0$ appear to be
related to the structure of a candidate phylogenetic tree,
the presence of a persistent $H_1$ points to the presence
of loops, hence to graphs that are not trees. Persistent
generators of the $H_1$ are also visible in the Longobardi
data. This will be further discussed in \cite{PortMa2}. 

\smallskip

We discuss some possible modifications of the evolutionary Markov
model on the tree in the last section of the paper.

\section{Phylogenetic Algebraic Varieties of the Germanic language family}

As discussed in the Introduction, we first analyze the phylogenetic tree
for the set of Germanic languages $\cS_1(G)$: 
Dutch, German, English, Faroese,  Icelandic, and Swedish. 

\smallskip

These six languages are mapped with different levels of accuracy in the SSWL database: 
we have Dutch (100$\%$), German (75$\%$), English (75$\%$), Faroese (62$\%$), Icelandic (62$\%$),
Swedish (75$\%$). There are 90 syntactic variables that are completely mapped for all of these six
languages: the list is reported in Appendix A. We will use only these 90 variables for the analysis
carried out here. 

\smallskip

We then consider the set $\cS_2(G)$ consisting of seven Germanic languages: 
Norwegian, Danish, Icelandic, German, English, Gothic, Old English. These
are chosen so that they are covered by both the SSWL database \cite{SSWL} and the new
data of Longobardi \cite{Long}, and so that they contain some ancient languages, 
in addition to modern languages situated on both the West and the North Germanic branches.
In this way we can test both the effect of using different syntactic data and the effect of
including ancient languages and their relation to problem of the location of the root vertex 
mentioned above.

\smallskip

The Germanic languages in the set $\cS_2(G)$ have a total of 68 SSWL variables that are completely
mapped for all the seven languages in the set. This is significantly smaller than the 90 variables used for the
set $\cS_1(G)$. This does not depend on the languages being poorly mapped: the levels of accuracy are
comparable with the previous set with Danish (76$\%$), Norwegian (75$\%$), 
German (75$\%$), English (75$\%$), Old English (75$\%$) Icelandic (62$\%$), Gothic (62$\%$). However,
the regions of the overall 115 SSWL variables that are mapped is less uniform across this set of
languages creating a smaller overlap. The set of completely mapped SSWL variables for this
set of languages is reported in Appendix~B. 

\smallskip
\subsection{Candidate PHYLIP trees} 
When using the full but incomplete data for the six Germanic languages in $\cS_1(G)$, we
obtain with PHYLIP a list of six candidate phylogenetic trees, respectively given
(in bracket notation) by
\begin{eqnarray*}
 {\tt pars1} & = & ((\ell_1,\ell_2),(\ell_3,(\ell_4, \ell_5)),\ell_6) \\
 {\tt pars2} & = & ((\ell_3,(\ell_1,\ell_2)), (\ell_4,\ell_5),\ell_6)  \\
 {\tt pars3} & = & (\ell_3,((\ell_1,\ell_2),(\ell_4, \ell_5)),\ell_6)  \\
 {\tt bnb1} & =  & (\ell_6,((\ell_5,\ell_4),(\ell_3,(\ell_2,\ell_1))))  \\
 {\tt bnb2} & = & (\ell_6,(((\ell_5,\ell_4),\ell_3),(\ell_1,\ell_2))) \\
 {\tt bnb3} & = & (\ell_6,(((\ell_5,\ell_4),(\ell_1,\ell_2)),\ell_3)) \\
\end{eqnarray*}
where $\ell_1=$Dutch, $\ell_2=$German, $\ell_3=$English, 
 $\ell_4=$Faroese, $\ell_5=$Icelandic, $\ell_6=$Swedish. 
 The Newick representation of binary trees used by PHYLIP lists the
 leaves in the order specified by the choice of a planar embedding
 of the tree, with brackets and commas indicating the joining together
 of branches. In the rest of the paper, for convenience, we will spell 
 out explicitly the form of the tree graphically, rather than writing them in 
 the Newick bracket notation. In the case of the trees listed here we obtain
 the following.
 
The trees {\tt pars1}, {\tt pars2}, and {\tt pars3} given above in  the Newick representation
have the form
\begin{center}
\Tree  [ [ $\ell_1$  $\ell_2$ ] [ [ $\ell_3$ [ $\ell_4$  $\ell_5$ ] ] $\ell_6$ ] ]  \ \ \ \ \ \ 
\Tree  [  [ $\ell_3$ [ $\ell_1$ $\ell_2$ ] ]  [ $\ell_4$ $\ell_5$ ] $\ell_6$  ]    \ \ \ \ \ \ 
\Tree  [ $\ell_3$ [ [ $\ell_1$ $\ell_2$ ] [ $\ell_4$ $\ell_5$ ]  ] $\ell_6$ ]  
\end{center}
Note that {\tt pars1} is a binary tree, while {\tt pars2} and {\tt pars3} are
not binary trees. We will discuss how to resolve the non-binary structure.
The remaining trees {\tt bnb1}, {\tt bnb2}, and  {\tt bnb3}  are binary trees of the form
\begin{center}
\Tree  [ $\ell_6$ [ [ $\ell_5$ $\ell_4$ ] [ $\ell_3$ [ $\ell_2$ $\ell_1$ ] ] ] ]  \ \ \ \ \ \ 
\Tree  [ $\ell_6$  [ [ [ $\ell_5$ $\ell_4$ ] $\ell_3$ ] [ $\ell_1$ $\ell_2$ ] ] ]  \ \ \ \ \ \ 
\Tree  [ $\ell_6$ [ [ [ $\ell_5$ $\ell_4$ ] [ $\ell_1$ $\ell_2$ ]  ] $\ell_3$ ] ] 
\end{center}
Note how all of these candidate trees agree on the proximity of Dutch and German ($\ell_1$ and $\ell_2$)
and of Faroese and Icelandic ($\ell_4$ and $\ell_5$), while they differ in the relative placement
of these two pairs with respect to one another and with respect to the two remaining languages, English
and Swedish. 

\smallskip

In phylogenetic linguistics the presence of a non-binary tree denotes an ambiguity, which
should eventually be resolved into one of its possible binary splittings.  
As shown in \cite{ERSS}, the phylogenetic algebraic variety of a non-binary tree can be
seen as the intersection of the phylogenetic algebraic varieties of all of its possible binary
splittings. Thus, the phylogenetic ideal (for the binary Jukes-Cantor model) is generated by 
all the $3\times 3$ minors of all the flattening matrices of all the binary splittings of the given 
non-binary tree. Being the intersection of the varieties defined by each of the
binary splittings corresponds exactly to the notion of ambiguity mentioned above. 

\smallskip

The resolution of a non-binary structure of the type shown in {\tt pars2} and {\tt pars3} 
is obtained by replacing the first tree below with the different possibilities given by its three
possible binary splittings that follow:
\begin{center}
\Tree [ A B C ]  \ \ \ \ \ \ 
\Tree [ A [ B C ] ] \ \ \ \ \ \  \Tree [ [ A B ] C ] \ \ \ \ \ \  \Tree [ [ A C ] B ]
\end{center}
Thus, for the tree {\tt pars2} we obtain the three binary trees
\begin{center}
\Tree [ [ $\ell_3$ [ $\ell_1$ $\ell_2$ ] ] [ [ $\ell_4$ $\ell_5$ ]  $\ell_6$ ] ] \ \ \ \ \ \
\Tree [ [ [ $\ell_3$ [ $\ell_1$ $\ell_2$ ] ]  [ $\ell_4$ $\ell_5$ ] ] $\ell_6$ ] \ \ \ \ \ \
\Tree [ [ [ $\ell_3$ [ $\ell_1$ $\ell_2$ ] ] $\ell_6$ ] [ $\ell_4$ $\ell_5$ ] ]
\end{center}
Note, however, that these three binary trees are equivalent up to a shift in
the position of the root, which however does not affect the phylogenetic
invariants, see \cite{AllRho} and Proposition 2.16 in \cite{Bocci}. Thus,
we need only consider one of them for the purpose of computing the
generators of the phylogenetic ideal. 
For the tree {\tt pars3} we obtain the three binary trees
\begin{center}
\Tree [ $\ell_3$ [ [  [ $\ell_1$ $\ell_2$ ] [ $\ell_4$ $\ell_5$ ] ] $\ell_6$ ] ] \ \ \ \ \ \
\Tree [ [ $\ell_3$ [  [ $\ell_1$ $\ell_2$ ] [ $\ell_4$ $\ell_5$ ] ] ] $\ell_6$ ] \ \ \ \ \ \
\Tree [ [ $\ell_3$ $\ell_6$ ] [  [ $\ell_1$ $\ell_2$ ] [ $\ell_4$ $\ell_5$ ] ] ]
\end{center}
Again these three binary trees only differ by a shift of the position of the root,
which does not affect the computation of the phylogenetic invariants, hence
we need only consider one of them for that purpose. Notice, moreover, that
the binary tree {\tt bnb1} is the same as the second binary tree for {\tt pars2}. 
Also the tree {\tt bnb2} has the same topology as the tree {\tt pars1}, up to
a shift in the position of the root, which does not affect the phylogenetic invariants.
Similarly, the tree {\tt bnb3} is the same as the second binary tree of {\tt pars3}.

\smallskip

All of the binary trees considered here have three internal edges, hence all of them
have three flattenings ${\rm Flat}_e(P)$ of the boundary distribution $P=(p_{i_1,\ldots, i_6})$. 

\smallskip

\begin{itemize}
\item The flattenings for {\tt pars1} are given by a $4\times 16$ matrix ${\rm Flat}_{e_1}(P)$,
an $8\times 8$ matrix ${\rm Flat}_{e_2}(P)$ and a $16\times 4$ matrix ${\rm Flat}_{e_3}(P)$.
These correspond to the separating the leaves into two components when deleting
the internal edge $e_i$ according to
$$ e_1 : \{ \ell_1, \ell_2 \} \cup \{ \ell_3, \ell_4, \ell_5, \ell_6 \} $$
$$ e_2 : \{ \ell_1, \ell_2, \ell_6 \} \cup \{ \ell_3, \ell_4, \ell_5 \} $$
$$ e_3 : \{ \ell_1, \ell_2, \ell_3, \ell_6 \} \cup \{ \ell_4, \ell_5 \}.  $$
\item The flattenings for any of the three binary trees for {\tt pars2} are also given by a
$4\times 16$ matrix ${\rm Flat}_{e_1}(P)$, an $8\times 8$ matrix ${\rm Flat}_{e_2}(P)$ 
and a $16\times 4$ matrix ${\rm Flat}_{e_3}(P)$, which in this case correspond to the subdivisions
$$ e_1 : \{ \ell_1, \ell_2 \} \cup \{ \ell_3, \ell_4, \ell_5, \ell_6 \} $$
$$ e_2 : \{ \ell_1, \ell_2, \ell_3 \} \cup \{ \ell_4, \ell_5, \ell_6 \} $$
$$ e_3 : \{ \ell_1, \ell_2, \ell_3, \ell_6 \} \cup \{ \ell_4, \ell_5 \}, $$
which only differ from the previous case in the $e_2$ flattening. 
\item The flattenings for any of the three binary trees for {\tt pars3} are given by a 
$4\times 16$ matrix ${\rm Flat}_{e_1}(P)$, a $16\times 4$ matrix ${\rm Flat}_{e_2}(P)$ 
and a $16\times 4$ matrix ${\rm Flat}_{e_3}(P)$, which 
correspond to the subdivisions
$$ e_1 : \{ \ell_1, \ell_2 \} \cup \{ \ell_3, \ell_4, \ell_5, \ell_6 \} $$
$$ e_2 : \{ \ell_1, \ell_2, \ell_3, \ell_6 \} \cup \{ \ell_4, \ell_5 \} $$
$$ e_3 : \{ \ell_1, \ell_2, \ell_4, \ell_5 \} \cup \{ \ell_3, \ell_6 \}. $$
\item The {\tt bnb1} tree is the same as one of binary trees for {\tt pars2}, hence
their flattenings are also the same.
\item The flattenings for {\tt bnb2} are the same as the flattening of {\tt pars1}, since
the two tree differ only by a shift in the position of the root vertex.
\item The {\tt bnb3} tree is the same as one of binary trees for {\tt pars3}, hence
their flattenings are also the same.
\end{itemize}
Thus, in order to compare the phylogenetic invariants of these various trees,
we need to compute the $3\times 3$ minors of the matrices ${\rm Flat}_e(P)$
for the splittings $\{ \ell_1, \ell_2 \} \cup \{ \ell_3, \ell_4, \ell_5, \ell_6 \}$,
$\{ \ell_1, \ell_2, \ell_6 \} \cup \{ \ell_3, \ell_4, \ell_5 \}$, $\{ \ell_1, \ell_2, \ell_3, \ell_6 \} \cup \{ \ell_4, \ell_5 \}$,
$\{ \ell_1, \ell_2, \ell_3 \} \cup \{ \ell_4, \ell_5, \ell_6 \}$, $\{ \ell_1, \ell_2, \ell_4, \ell_5 \} \cup \{ \ell_3, \ell_6 \}$.
We will compute these in the next subsection.

\smallskip
\subsection{Flattenings}

As discussed above, there are five matrices ${\rm Flat}_e(P)$ that occur in the
computation of the phylogenetic ideals of the candidate phylogenetic trees listed above.
In fact, we do not need to compute all of them, as some occur in all the trees, hence
do not contribute to distinguishing between them. This corresponds to the observation
we already made above, that all the candidate trees agree on the proximity of $\ell_1$
and $\ell_2$ and of $\ell_4$ and $\ell_5$.

\begin{itemize}
\item The $4\times 16$ matrix ${\rm Flat}_{\{ \ell_1, \ell_2 \} \cup \{ \ell_3, \ell_4, \ell_5, \ell_6 \}}(P)$,
contributes to the phylogenetic ideals of all the trees, hence it will not help discriminate between them. 
\item The same is true about the $16 \times 4$ matrix 
${\rm Flat}_{\{ \ell_1, \ell_2, \ell_3, \ell_6 \} \cup \{ \ell_4, \ell_5 \}}(P)$.
\item The $8 \times 8$ matrix ${\rm Flat}_{ \{ \ell_1, \ell_2, \ell_6 \} \cup \{ \ell_3, \ell_4, \ell_5 \} }(P)$
contributes to the phylogenetic invariants of {\tt pars1} and {\tt bnb2}. It is given by  {\small
$$ \left(\begin{array}{cccccccc}
p_{000000} & p_{000100} &  p_{001000} & p_{001100} & p_{000010} & p_{000110} & p_{001010} & p_{001110} \\
p_{010000} & p_{010100} & p_{011000} & p_{011100} & p_{010010} & p_{010110} & p_{011010} & p_{011110} \\
p_{100000} & p_{100100} & p_{101000} & p_{101100} & p_{100010} & p_{100110} & p_{101010} & p_{101110} \\
p_{110000} & p_{110100} & p_{111000} & p_{111100} & p_{110010} & p_{110110} & p_{111010} & p_{111110} \\ 
p_{000001} & p_{000101} &  p_{001001} & p_{001101} & p_{000011} & p_{000111} & p_{001011} & p_{001111} \\
p_{010001} & p_{010101} & p_{011001} & p_{011101} & p_{010011} & p_{010111} & p_{011011} & p_{011111} \\
p_{100001} & p_{100101} & p_{101001} & p_{101101} & p_{100011} & p_{100111} & p_{101011} & p_{101111} \\
p_{110001} & p_{110101} & p_{111001} & p_{111101} & p_{110011} & p_{110111} & p_{111011} & p_{111111} \\ 
\end{array}\right) $$ }
\item The $8 \times 8$ matrix ${\rm Flat}_{ \{ \ell_1, \ell_2, \ell_3 \} \cup \{ \ell_4, \ell_5, \ell_6 \} }(P)$
contributes to the phylogenetic invariants of {\tt pars2} and {\tt bnb1} and it is given by {\small
$$ \left(\begin{array}{cccccccc}
p_{000000} & p_{000010} &  p_{000100} & p_{000110} & p_{000001} & p_{000011} & p_{000101} & p_{000111} \\
p_{010000} & p_{010010} & p_{010100} & p_{010110} & p_{010001} & p_{010011} & p_{010101} & p_{010111} \\
p_{100000} & p_{100010} & p_{100100} & p_{100110} & p_{100001} & p_{100011} & p_{100101} & p_{100111} \\
p_{110000} & p_{110010} & p_{110100} & p_{110110} & p_{110001} & p_{110011} & p_{110101} & p_{110111} \\ 
p_{001000} & p_{001010} &  p_{001100} & p_{001110} & p_{001001} & p_{001011} & p_{001101} & p_{001111} \\
p_{011000} & p_{011010} & p_{011100} & p_{011110} & p_{011001} & p_{011011} & p_{011101} & p_{011111} \\
p_{101000} & p_{101010} & p_{101100} & p_{101110} & p_{101001} & p_{101011} & p_{101101} & p_{101111} \\
p_{111000} & p_{111010} & p_{111100} & p_{111110} & p_{111001} & p_{111011} & p_{111101} & p_{111111} \\ 
\end{array}\right) $$ }
\item The $16\times 4$ matrix ${\rm Flat}_{ \{ \ell_1, \ell_2, \ell_4, \ell_5 \} \cup \{ \ell_3, \ell_6 \} }(P)$
contributes to the phylogenetic invariants of {\tt pars3} and {\tt bnb3} and is given by {\small 
$$ \left(\begin{array}{cccc}
p_{00 0 00 0} & p_{00 0 00 1} & p_{00 1 00 0} & p_{00 1 00 1} \\
p_{01 0 00 0} & p_{01 0 00 1} & p_{01 1 00 0} & p_{01 1 00 1} \\
p_{10 0 00 0} & p_{10 0 00 1} & p_{10 1 00 0} & p_{10 1 00 1} \\
p_{11 0 00 0} & p_{11 0 00 1} & p_{11 1 00 0} & p_{11 1 00 1} \\

p_{00 0 01 0} & p_{00 0 01 1} & p_{00 1 01 0} & p_{00 1 01 1} \\
p_{01 0 01 0} & p_{01 0 01 1} & p_{01 1 01 0} & p_{01 1 01 1} \\
p_{10 0 01 0} & p_{10 0 01 1} & p_{10 1 01 0} & p_{10 1 01 1} \\
p_{11 0 01 0} & p_{11 0 01 1} & p_{11 1 01 0} & p_{11 1 01 1} \\

p_{00 0 10 0} & p_{00 0 10 1} & p_{00 1 10 0} & p_{00 1 10 1} \\
p_{01 0 10 0} & p_{01 0 10 1} & p_{01 1 10 0} & p_{01 1 10 1} \\
p_{10 0 10 0} & p_{10 0 10 1} & p_{10 1 10 0} & p_{10 1 10 1} \\
p_{11 0 10 0} & p_{11 0 10 1} & p_{11 1 10 0} & p_{11 1 10 1} \\

p_{00 0 11 0} & p_{00 0 11 1} & p_{00 1 11 0} & p_{00 1 11 1} \\
p_{01 0 11 0} & p_{01 0 11 1} & p_{01 1 11 0} & p_{01 1 11 1} \\
p_{10 0 11 0} & p_{10 0 11 1} & p_{10 1 11 0} & p_{10 1 11 1} \\
p_{11 0 11 0} & p_{11 0 11 1} & p_{11 1 11 0} & p_{111111} \\
\end{array}\right) $$ }

\end{itemize}

\smallskip
\subsection{Boundary distribution and phylogenetic invariants}

Next we compute the boundary distribution $P=(p_{i_1,\ldots, i_6})$ of the syntactic variables. 
We use only the 90 completely mapped syntactic variables, for which we find occurrences
$$ \begin{array}{cccc}
n_{110111}=3 & n_{000011}=1 & n_{000010}=4 & n_{000000}=40 \\
n_{110000}=2 & n_{001110}=1 & n_{000100}=2 & n_{111111}=22 \\
n_{111110}=1 & n_{000110}=1 & n_{111101}=3 & n_{100000} =2 \\
n_{010000}=1 & n_{111001}=2 & n_{110110}=1 & n_{010111} =1 \\
n_{001000}= 2 & n_{000111}=1 &  & \\
\end{array} $$
while all the remaining cases do not occur, $n_{i_1,\ldots, i_6}=0$ for
$(i_1,\ldots, i_n)$ not in the above list. 

With the boundary distribution determined by the occurrences above the three matrices
of $F1={\rm Flat}_{ \{ \ell_1, \ell_2, \ell_6 \} \cup \{ \ell_3, \ell_4, \ell_5 \} }(P)$,
$F_2={\rm Flat}_{ \{ \ell_1, \ell_2, \ell_3 \} \cup \{ \ell_4, \ell_5, \ell_6 \} }(P)$, and
$F_3={\rm Flat}_{ \{ \ell_1, \ell_2, \ell_4, \ell_5 \} \cup \{ \ell_3, \ell_6 \} }(P)$ are,
respectively, given by
{\small
$$ F_1= \left(\begin{array}{cccccccc}
\frac{4}{9} & \frac{1}{45} &  \frac{1}{45} & 0 & \frac{2}{45} & \frac{1}{90} & 0 & \frac{1}{90}  \\
\frac{1}{90} & 0 & 0 & 0 & 0 & 0 & 0 & 0 \\
\frac{1}{45} & 0 & 0 & 0 & 0 & 0 & 0 & 0 \\
\frac{1}{45}  & 0 & 0 & 0 & 0 & \frac{1}{90} & 0 & \frac{1}{90} \\ 
0 & 0 &  0 & 0 & \frac{1}{90} & \frac{1}{90} & 0 & 0 \\
0 & 0 & 0 & 0 & 0 & \frac{1}{90} & 0 & 0 \\
0 & 0 & 0 & 0 & 0 & 0 & 0 & 0 \\
0 & 0 & \frac{1}{45} & \frac{1}{30} & 0 & \frac{1}{30} & 0 & \frac{11}{45}\\ 
\end{array}\right) $$ 
$$ F_2=\left(\begin{array}{cccccccc}
\frac{4}{9} & \frac{2}{45} &  \frac{1}{45} & \frac{1}{90} & 0 & \frac{1}{90} & 0 & \frac{1}{90} \\
\frac{1}{90} & 0 & 0 & 0 & 0 & 0 & 0 & \frac{1}{90} \\
\frac{1}{45} & 0 & 0 & 0 & 0 & 0 & 0 & 0 \\
\frac{1}{45} & 0 & 0 & \frac{1}{90} & 0 & 0 & 0 & \frac{1}{30} \\ 
\frac{1}{45} & 0 & 0 & \frac{1}{90}  & 0 & 0 & 0 & 0 \\
0 & 0 & 0 & 0 & 0 & 0 & 0 & 0 \\
0 & 0 & 0 & 0 & 0 & 0 & 0 & 0 \\
0 & 0 & 0 & \frac{1}{90} & \frac{1}{45} & 0 & \frac{1}{30} & \frac{11}{45}\\ 
\end{array}\right) $$ 
$$ F_3=\left(\begin{array}{cccc}
\frac{4}{9} & 0 & \frac{1}{45} & 0 \\
\frac{1}{90} & 0 & 0 & 0 \\
\frac{1}{45} & 0 & 0 & 0 \\
\frac{1}{45} & 0 & 0 & \frac{1}{45} \\
\frac{2}{45} & \frac{1}{90} & 0 & 0 \\
0 & 0 & 0 & 0 \\
0 & 0 & 0 & 0 \\
0 & 0 & 0 & 0 \\
\frac{1}{45} & 0 & 0 & 0 \\
0 & 0 & 0 & 0 \\
0 & 0 & 0 & 0 \\
0 & 0 & 0 & \frac{1}{30} \\
\frac{1}{90} & \frac{1}{90} & \frac{1}{90}  & 0  \\
0 & \frac{1}{90} & 0 & 0 \\
0 & 0 & 0 & 0 \\
\frac{1}{90} & \frac{1}{30} & \frac{1}{90} & \frac{11}{45}\\
\end{array}\right) $$ }

\smallskip
\subsection{Phylogenetic invariants}

As we discussed above, the flattening matrices 
$$ {\rm Flat}_{\{ \ell_1, \ell_2 \} \cup \{ \ell_3, \ell_4, \ell_5, \ell_6 \}}(P) \ \ \
\text{ and } \ \ \  {\rm Flat}_{\{ \ell_1, \ell_2, \ell_3, \ell_6 \} \cup \{ \ell_4, \ell_5 \}}(P)$$
occur in all the candidate trees, hence they do not discriminate between
the given candidates (preselected by PHYLIP). Thus it is reasonable to
proceed by assuming that the condition that these two flattenings lie on the 
corresponding determinantal varieties is satisfied and only discriminate 
between the candidate trees on the basis of the position of the remaining
flattenings. There is only one additional flattening
involved in each tree, once these common ones are excluded. Thus,
we estimate the phylogenetic invariants by evaluating the $3\times 3$ minors
of the remaining flattening matrix for each of the trees, using both 
the $\ell^\infty$  and the $\ell^1$ norm. We obtain the following:
\begin{enumerate}
\item For the tree {\tt pars1} (and equivalently {\tt bnb2}) we have
$$ \| \phi_{T_1}(P)\|_{\ell^\infty} =\max_{\phi\in 3\times 3 \text{ minors of } F_1} |\phi(P)| =  \frac{22}{18225} $$
$$ \| \phi_{T_1}(P)\|_{\ell^1} =\sum_{\phi\in 3\times 3 \text{ minors of } F_1} |\phi(P)| = \frac{ 3707}{364500 } $$
\item For the tree {\tt pars2} (equivalently {\tt bnb1}) we have
$$ \| \phi_{T_2}(P)\|_{\ell^\infty} =\max_{\phi\in 3\times 3 \text{ minors of } F_2} |\phi(P)| =  \frac{419}{364500} $$
$$ \| \phi_{T_2}(P)\|_{\ell^1} =\sum_{\phi\in 3\times 3 \text{ minors of } F_2} |\phi(P)| = \frac{ 2719}{364500  } $$
\item For the tree {\tt pars3} (and equivalently {\tt bnb3}) we have
$$ \| \phi_{T_3}(P)\|_{\ell^\infty} =\max_{\phi\in 3\times 3 \text{ minors of } F_3} |\phi(P)| =  \frac{22}{18225} $$
$$ \| \phi_{T_3}(P)\|_{\ell^1} =\sum_{\phi\in 3\times 3 \text{ minors of } F_3} |\phi(P)| = \frac{ 949}{91125 } $$
\end{enumerate}

Thus, in terms of the evaluation of the phylogenetic invariants,
the binary trees of {\tt pars2} and the binary tree {\tt bnb1} are favorite over the other 
possibilities. (We discuss the position of the root vertex below.) Note that the $\ell^\infty$
norm does not distinguish between the other two remaining candidates and only singles
out the preferred candidate {\tt pars2}.
We compute the Euclidean distance function in \S \ref{distGerm1sec}.

\smallskip
\subsection{The problem with the root vertex}\label{rootSec}

As we have seen above, the computation of the phylogenetic invariants helps
selecting between different candidate tree topologies. However, the phylogenetic
invariants by themselves are insensitive to changing the position of the root
in binary trees with the same topology. In terms of phylogenetic inference about
Linguistics, however, it is important to locate more precisely where the root
vertex should be. In the case of languages belonging to a subfamily of the 
Indo-European languages this can be done, as in the example we discussed
in \cite{ShuMa}, by introducing the data of some of the ancient languages
in the same subfamily as a new leaf of the tree, that will help locating more
precisely the root vertex of the original tree based on the modern languages.
For language families for which there are no data of ancient languages available,
however, this kind of phylogenetic analysis will only identify a tree topology as
an unrooted binary tree. We will return to this point in the following
section, where we analyze the set $\cS_2(G)$ which includes two
ancient languages. 

\smallskip

Note that when one or more ancient languages are included in the data (as in the second case of the 
Germanic languages, or the Romance languages discussed here) that suffices to constrain 
the position of the root vertex, while in other cases like the example discussed here, additional 
independent information is needed.

\smallskip
\subsection{Varieties}\label{Var1Sec}

In the discussion above we reduced the question of distinguishing
between the candidate trees to an evaluation of the phylogenetic invariants
coming from the $3\times 3$ minors of one of the three matrices 
${\rm Flat}_{ \{ \ell_1, \ell_2, \ell_6 \} \cup \{ \ell_3, \ell_4, \ell_5 \} }(P)$,
${\rm Flat}_{ \{ \ell_1, \ell_2, \ell_3 \} \cup \{ \ell_4, \ell_5, \ell_6 \} }(P)$, and
${\rm Flat}_{ \{ \ell_1, \ell_2, \ell_4, \ell_5 \} \cup \{ \ell_3, \ell_6 \} }(P)$.
In the first two cases, the phylogenetic ideal defines the $28$-dimensional
determinantal variety of all $8 \times 8$ matrices of rank at most two,
while in the third case the phylogenetic ideal defines the $36$-dimensional
determinantal variety of all $16\times 4$ matrices
of rank at most two, \cite{BruVe}. 
These are not the actual phylogenetic varieties associated to the candidate
trees, which are further cut out by the remaining equations coming from
the $3\times 3$ minors of the other flattenings 
${\rm Flat}_{\{ \ell_1, \ell_2 \} \cup \{ \ell_3, \ell_4, \ell_5, \ell_6 \}}(P)$,
and ${\rm Flat}_{\{ \ell_1, \ell_2, \ell_3, \ell_6 \} \cup \{ \ell_4, \ell_5 \}}(P)$.
The varieties associated to each individual tree are intersections of 
three different determinantal varieties inside a common ambient space
$\A^{2^6}$, or when considered projectively (all the polynomials defining
the phylogenetic ideals are homogeneous) in $\P^{2^6-1}$.

\smallskip

In the case of the trees considered here, two of the three determinantal
varieties stay the same, since the flattenings 
${\rm Flat}_{\{ \ell_1, \ell_2 \} \cup \{ \ell_3, \ell_4, \ell_5, \ell_6 \}}(P)$,
and ${\rm Flat}_{\{ \ell_1, \ell_2, \ell_3, \ell_6 \} \cup \{ \ell_4, \ell_5 \}}(P)$
are common to all candidate trees, while the third component varies
among the three choices determined by the flattenings
${\rm Flat}_{ \{ \ell_1, \ell_2, \ell_6 \} \cup \{ \ell_3, \ell_4, \ell_5 \} }(P)$,
${\rm Flat}_{ \{ \ell_1, \ell_2, \ell_3 \} \cup \{ \ell_4, \ell_5, \ell_6 \} }(P)$, and
${\rm Flat}_{ \{ \ell_1, \ell_2, \ell_4, \ell_5 \} \cup \{ \ell_3, \ell_6 \} }(P)$.

\smallskip

In general, let $\cD_r(n,m)$ denote the determinantal variety of $n\times m$
matrices of rank $\leq r$. As an affine subvariety in $\A^{nm}$ it has dimension 
$r(n+m-r)$. It will be convenient to consider $\cD_r(n,m)$ as a projective
subvariety of $\P^{nm-1}$, though we will maintain the same notation.
In the case $r=1$, the determinantal variety $\cD_1(n,m)$ is the Segre
variety $\cS(n,m)$ given by the embedding $\P^{n-1} \times \P^{m-1} \hookrightarrow \P^{nm-1}$
realized by the Segre map $(x_i, y_j)\mapsto (u_{ij}=x_iy_j)$. In the case $r=2$
the determinantal variety $\cD_2(n,m)$ is the secant variety of lines (chord variety)
${\rm Sec}(\cS(n,m))$ of the Segre variety $\cS(n,m)$, see \S 9 of \cite{Harris}.

\smallskip

Thus, we obtain the following simple geometric description of the three cases
considered above:
\begin{itemize}
\item ${\rm Flat}_{ \{ \ell_1, \ell_2, \ell_6 \} \cup \{ \ell_3, \ell_4, \ell_5 \} }(P)$ (tree topology of {\tt pars1}
and {\tt bnb2}): the relevant variety is the secant variety ${\rm Sec}(\cS(8,8))$ of the Segre variety 
$\cS(8,8)=\P^7\times \P^7$, embedded in $\P^{63}$ via the Segre embedding 
$u_{i_1,\ldots, i_6} = x_{i_1, i_2, i_6} y_{i_3, i_4, i_5}$.
 \item ${\rm Flat}_{ \{ \ell_1, \ell_2, \ell_3 \} \cup \{ \ell_4, \ell_5, \ell_6 \} }(P)$ (tree topology of {\tt pars2}
and {\tt bnb1}): the relevant variety is again ${\rm Sec}(\cS(8,8))$, where $\cS(8,8)$ is embedded in 
$\P^{63}$ via $u_{i_1,\ldots, i_6} = x_{i_1, i_2, i_3} y_{i_4, i_5, i_6}$.
\item ${\rm Flat}_{ \{ \ell_1, \ell_2, \ell_4, \ell_5 \} \cup \{ \ell_3, \ell_6 \} }(P)$ (tree topology of {\tt pars3} 
and {\tt bnb3}): the relevant variety is the secant variety 
${\rm Sec}(\cS(16,4))$ of the Segre variety 
$\cS(16,4)=\P^{15}\times \P^3$, embedded in $\P^{63}$ via the Segre embedding 
$u_{i_1,\ldots, i_6} = x_{i_1, i_2, i_4, i_5} y_{i_3, i_6}$.
\end{itemize}
The evaluation of the phylogenetic invariants at the boundary distribution determined by
the SSWL data selects the second choice, ${\rm Sec}(\cS(8,8))$ with the Segre embedding 
$u_{i_1,\ldots, i_6} = x_{i_1, i_2, i_3} y_{i_4, i_5, i_6}$.

\smallskip

As a general procedure, given a subfamily of languages, $\{ \ell_1, \ldots, \ell_n \}$
and a set of candidate phylogenetic trees $T_1, \ldots, T_m$ produced by
computational methods from the syntactic variables of these $n$ languages,
one can construct with the method above a collection $Y_1, \ldots, Y_m$
of algebraic varieties, where each $Y_k$ associated to the tree $T_k$ is obtained 
by considering the determinantal varieties associated to all those flattenings 
${\rm Flat}_e(P)$ of $T_k$ that are not common to all the other trees $T_j$.

\smallskip

The test for selecting one of the candidate trees, given the boundary distribution
$P=(p_{i_1,\ldots, i_n})$ of the syntactic variables, is then to estimate 
which of the varieties $Y_k$ the point $P$ is closest to, where a suitable
test of closeness is used, for instance through the Euclidean distance function. 
Assuming that this procedure does not result in ambiguities (that is, that there
is a unique closest $Y_k$ to the given distribution $P$), then this method
selects a best candidate $T$ among the $m$ trees $T_k$. It also selects an
associated algebraic variety $Y=Y(T)$, which is larger than the usual 
phylogenetic algebraic variety $X_T$ of $T$, since we have neglected flattenings 
that occur simultaneously in all the $m$ candidate trees $T_k$. 

\smallskip
\subsection{The Euclidean distance}\label{distGerm1sec}

According to the discussion of the previous subsection, on the geometry of
the varieties involved in distinguishing between the candidate trees,
we compute here 
\begin{itemize}
\item the Euclidean distance of the point
${\rm Flat}_{ \{ \ell_1, \ell_2, \ell_6 \} \cup \{ \ell_3, \ell_4, \ell_5 \} }(P)$
and the determinantal variety $\cD_2(8,8)={\rm Sec}(\cS(8,8))$, 
\item the Euclidean  distance of the point 
${\rm Flat}_{ \{ \ell_1, \ell_2, \ell_3 \} \cup \{ \ell_4, \ell_5, \ell_6 \} }(P)$
from the same determinantal variety $\cD_2(8,8)={\rm Sec}(\cS(8,8))$, 
\item the Euclidean distance of the point
${\rm Flat}_{ \{ \ell_1, \ell_2, \ell_4, \ell_5 \} \cup \{ \ell_3, \ell_6 \} }(P)$
from the determinantal variety $\cD_2(16,4) ={\rm Sec}(\cS(16,4))$.
\end{itemize}
Using the Eckart-Young theorem, we compute these distances
using the singular values of these three matrices. These are given by 
$$ \Sigma({\rm Flat}_{ \{ \ell_1, \ell_2, \ell_6 \} \cup \{ \ell_3, \ell_4, \ell_5 \} }(P)) \sim $$
$$ {\rm diag}(0.44940, 0.25001, 0.19237 \times 10^{-1}, 
0.96007 \times 10^{-2}, 0.21595 \times 10^{-2},0.88079 \times 10^{-3}, 4.6239 \times 10^{-19},0) $$
$$ \Sigma({\rm Flat}_{ \{ \ell_1, \ell_2, \ell_3 \} \cup \{ \ell_4, \ell_5, \ell_6 \} }(P)) \sim $$
$$ {\rm diag}( 
0.44956, 0.25018, 0.14729 \times 10^{-1}, 0.44229 \times 10^{-2}, 0.27802  \times 10^{-2},  0.24881 \times 10^{-17}, 0 ) $$
$$ \Sigma({\rm Flat}_{ \{ \ell_1, \ell_2, \ell_4, \ell_5 \} \cup \{ \ell_3, \ell_6 \} }(P)) \sim $$
$$ {\rm diag}( 0.44939, 0.24994, 0.20625  \times 10^{-1}, 0.94442   \times 10^{-2}). $$
Using \eqref{EuclDist} we then obtain
$$ {\rm dist}({\rm Flat}_{ \{ \ell_1, \ell_2, \ell_6 \} \cup \{ \ell_3, \ell_4, \ell_5 \} }(P),{\rm Sec}(\cS(8,8)))^2 =\sigma_3^2 +\cdots +\sigma_8^2=  0.46768 \times 10^{-3}   $$
$$ {\rm dist}({\rm Flat}_{ \{ \ell_1, \ell_2, \ell_3 \} \cup \{ \ell_4, \ell_5, \ell_6 \} }(P),{\rm Sec}(\cS(8,8)))^2 =\sigma_3^2 +\cdots +\sigma_8^2=  0.24424  \times 10^{-3}  $$
$$ {\rm dist}({\rm Flat}_{ \{ \ell_1, \ell_2, \ell_4, \ell_5 \} \cup \{ \ell_3, \ell_6 \} }(P),{\rm Sec}(\cS(16,4)))^2= 
\sigma_3^2 +\sigma_4^2 = 0.51457    \times 10^{-3}   $$
The second Euclidean distance is the smallest, hence this more reliable distance test again favors
the binary trees of {\tt pars2} and the binary tree {\tt bnb1}.

\smallskip

The computation of these Euclidean distances provides 
a selection between the candidate trees in the following way.
The first distance measures how far the point determined by the 
data (in the form of the boundary distribution $P$ and the 
flattening matrix $F_1(P)$) is from the determinantal variety 
$\cD_2(8,8)$ determined by the tree {\tt pars1}. The second
distance measures how far the point determined by the data,
through the flattening $F_2(P)$, is from the determinantal variety 
determined by the tree {\tt pars2}, and the third distance
measures how far the point, through the flattening $F_3(P)$
is from the determinantal variety $\cD_2(16,4)$ determined 
by the tree {\tt pars3}. Since as observed above the remaining
flattenings of $P$ occur in all trees and do not help distinguishing
between them, it suffices to find the best matching condition
between the three possibilities listed here, for which we select
the one realizing the smallest Euclidean distance. 

\smallskip

Unlike the other exmples that we discuss in the rest of the 
paper, where we will only obtain a lower bound estimate for
the Euclidean distance, in this case, under the conditional
assumption that the incidence of the two common flattenings
$$ {\rm Flat}_{\{ \ell_1, \ell_2 \} \cup \{ \ell_3, \ell_4, \ell_5, \ell_6 \}}(P) \ \ \
\text{ and } \ \ \  {\rm Flat}_{\{ \ell_1, \ell_2, \ell_3, \ell_6 \} \cup \{ \ell_4, \ell_5 \}}(P)$$
to the corresponding determinantal variety is realized, the computation
described above provides the actual value of the Euclidean distance
of the point $P$ to the phylogenetic algebraic variety $V_T$, see the
discussion in \S \ref{CondSec} above.

\medskip
\subsection{The West/North Germanic split from SSWL data}
Note that the tree topology selected in this way, which (up to the position of
the root vertex) is equivalent to the tree
\begin{center}
\Tree  [ [ Swedish [ Icelandic Faroese ] ] [ English [ Dutch German ] ] ] 
\end{center}
is also the generally acknowledged correct subdivision of the Germanic languages
into the North Germanic and the West Germanic sub-branches. The North Germanic
in turn splits into a sub-brach that contains Swedish (but also Danish which we
have not included here) and another that contains Icelandic and Faroese (and
also Norwegian, which we have not included, in order to keep the number of
leaves more manageable). The West Germanic branch is split into the Anglo-Frisian
sub-branch (of which here we are only considering English, but which should also
contain Frisian) and the Netherlandic-Germanic branch that contains Dutch and
German. Thus, the analysis through phylogenetic invariants and Euclidean distance has selected the
correct tree topology among the candidates produced by the computational
analysis of the SSWL data obtained with PHYLIP. 

\smallskip

\subsection{Longobardi data and phylogenetic invariants of Germanic Languages}\label{LongoGerSec}

Now we analyze the set $\cS_2(G)$ consisting of  
Norwegian, Danish, Icelandic, German, English, Gothic, and Old English,
using the syntactic parameters collected in the new
data of Longobardi \cite{Long}. 

\smallskip

\begin{center}
\begin{figure}[h]
\includegraphics[scale=0.85]{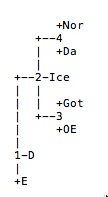} 
\caption{PHYLIP output trees of Germanic languages for the set $\cS_2(G)$ based on the 
Longobardi data. \label{FigS2GLong}}
\end{figure}
\end{center}

The DNA parsimony algorithm of PHYLIP based solely on the new Longobardi data produces
a single candidate phylogenetic tree for the set $\cS_2(G)$ of Germanic languages, shown
in Figure~\ref{FigS2GLong}.

\smallskip
 
 In fact, because of the presence of vertices of higher valence in this tree, one should resolve
 it into the possible binary trees and compare the resulting candidates. Moreover, the placement
 of the ancient languages as ``leaves" of the tree is an artifact, and needs to be resolved into
 the appropriate placement of the root of the binary trees. 
 
 \smallskip
 
 We see here that the fact that ancient languages are treated as leaves in the tree
 although they really are intermediate nodes creates some problems in the reconstruction
 provided by PHYLIP. In the tree of Figure~\ref{FigS2GLong} Gothic and Old English
 are grouped as nearby leaves in the tree, since the reconstruction correctly
 identifies the closer proximity of the two ancient languages with respect to the
 modern ones. However, this causes an error in the proposed tree topology when
 these are placed as two nearby leaves. The standard way of resolving the higher
 valence vertex in Figure~\ref{FigS2GLong}, as discussed in the previous section, 
 would maintain this problem. We propose here a simple method for avoiding
 this problem, via a simple topological move in the resulting trees that restores
 the role of these two languages as intermediate nodes of the tree (and suggests
 a position of the root vertex) while maintaining their relation to the rest of the tree.
 
 \smallskip
 
 In particular, this means that we are going to consider possible candidate trees of the following form,
 where we set $\ell_1=$ Norwegian, $\ell_2=$ Danish, $\ell_3=$ Gothic, $\ell_4=$ Old English,
 $\ell_5=$ Icelandic, $\ell_6=$ English, $\ell_7 =$ German.

 \smallskip
 
 We first visualize the trees obtained by resolving the vertex of Figure~\ref{FigS2GLong}.
 To simplify the picture, let us write $A=\{ \ell_1, \ell_2 \}$ for the end of the tree containing
 this pair of adjacent leaves, and similarly for $B=\{ \ell_3, \ell_4 \}$, $C=\{ \ell_5 \}$,
 $D=\{ \ell_6, \ell_7 \}$, so that we can visualize the the three possible binary splittings of the
 vertex in Figure~\ref{FigS2GLong} as the trees
 \begin{center}
 \Tree [.B [ A [ C D ] ]]  \ \  \ \  \ \  \Tree[.B [ [ A C ] D ]] \ \  \ \  \ \ \Tree [.B [ [ A D ] C ]].
 \end{center}
 We then want to input the extra piece of information concerning the fact that the
 leaves in the set $B=\{ \ell_3, \ell_4 \}$ are not really leaves but inner vertices 
 of the tree, whose proximity is describing the fact that they are in closer proximity 
 to the root of the tree than the other leaves, rather than their proximity as leaves.
 We argue that this can be done effectively by introducing a simple {\em topological
 move} on these trees that achieves exactly this effect, while preserving the relation
 to the rest of the tree, namely the following operation:
 \begin{center}
 \includegraphics[scale=0.25]{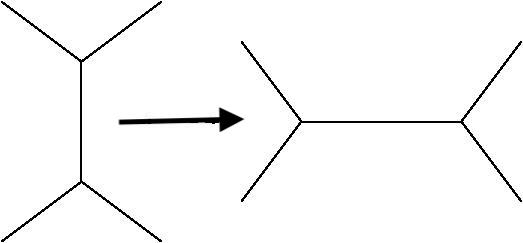} 
 \end{center}
 Applying this operation produces the following list of candidate trees, with
 (1) and (2) derived from the first binary tree above, (3) and (4) from the
 second binary tree above and (5) and (6) from the third one.
 
\begin{enumerate}
\item The first candidate tree $T_1(G)$ has
Icelandic (incorrectly) grouped together with the 
West Germanic (German, English) instead of the North Germanic (Norwegian, Danish) languages.
The labels $\ell_3$ and $\ell_4$ should be thought of not as leaves but as intermediate vertices
placed, respectively, above the $\{ \ell_1, \ell_2 \}$ subtree
and above the $\{ \ell_5, \ell_6, \ell_7 \}$ subtree. 
\begin{center}
\Tree [  [ [ $\ell_1$  $\ell_2$ ] $\ell_3$ ] [ $\ell_4$ [ $\ell_5$ [ $\ell_6$  $\ell_7$ ] ] ] ]
\end{center}

\item The second candidate tree $T_2(G)$
has the same structure as the previous list 
(with the incorrect placement of Icelandic), but with the reversed placement of the
two ancient languages $\ell_3$ and $\ell_4$, this time with Old English placed
at the top of the North Germanic instead of the West Germanic subtree:
\begin{center}
\Tree [ [ $\ell_4$ [ $\ell_1$  $\ell_2$ ] ] [ $\ell_3$ [ $\ell_5$ [ $\ell_6$  $\ell_7$ ] ] ] ]
\end{center}

\item The third candidate tree $T_3(G)$ has the correct placement of Icelandic
in the North Germanic subtree, with Gothic above the North Germanic and Old English
above the West Germanic subtrees:
\begin{center}
\Tree [  [ $\ell_3$ [ $\ell_5$ [ $\ell_1$  $\ell_2$ ]] ] [ $\ell_4$ [ $\ell_6$  $\ell_7$ ] ] ]
\end{center}

\item The fourth candidate tree $T_{4}(G)$ also has the correct placement of Icelandic
in the North Germanic subtree, now with Old English  above the North Germanic and Gothic
above the West Germanic subtrees:
\begin{center}
\Tree [  [ $\ell_4$ [ $\ell_5$ [ $\ell_1$  $\ell_2$ ]] ] [ $\ell_3$ [ $\ell_6$  $\ell_7$ ] ] ]
\end{center}

\item The fifth candidate incorrectly places the sets $\{ \ell_1, \ell_2 \}$ and $\{ \ell_6, \ell_7 \}$
in closer proximity and $\ell_5$ in a separate branch away from the ancient languages $\{ \ell_3, \ell_4\}$,
placing $\ell_4$ as the ancient language in closer proximity to $\ell_5$:
\begin{center}
\Tree [  [ $\ell_5$  $\ell_4$ ] [ $\ell_3$ [ [ $\ell_6$  $\ell_7$ ] [ $\ell_1$  $\ell_2$ ] ] ] ]
\end{center}

\item The sixth candidate tree also incorrectly places $\ell_5$ as a separate branch and
$\{ \ell_1, \ell_2 \}$ and $\{ \ell_6, \ell_7 \}$ in the same branch, while placing $\ell_3$
as the ancient language in closer proximity to $\ell_5$:
\begin{center}
\Tree [  [ $\ell_5$  $\ell_3$ ] [ $\ell_4$ [ [ $\ell_6$  $\ell_7$ ] [ $\ell_1$  $\ell_2$ ] ] ] ]
\end{center}
\end{enumerate}

\subsubsection{Comparison of the first four trees}
We first discuss the candidate trees (1)--(4) as these have a lot of common
structure that simplifies a common analysis. We then show what changes 
for the last two cases.

When considering the new Longobardi data for the purpose of computing phylogenetic
invariants, we need to eliminate from the list all those parameters that have value either
$0$ (undefined in the terminology of Longobardi's data table) or $?$ (unknown). The
reason for eliminating not just the unknown parameters but also those
rendered undefined by entailment relations lies in the fact that the result of \cite{AllRho}
that we use for the computation of the phylogenetic invariants holds for a {\em binary}
Jukes-Cantor model but not for a ternary one. Thus, we stick to only those parameters
that are defined with binary values $\pm 1$ in Longobardi's table, for all the languages
$\ell_1, \ldots, \ell_7$ in our list of Germanic languages. After the change of notation
to binary form, obtained by replacing $1\mapsto 1$ and $-1 \mapsto 0$, we obtain the
following list of parameters {\small
$$ \ell_1=[1,1,1,1,0,1,1,0,1,0,0,1,0,0,1,1,0,0,1,1,1,1,0,0,0,0,0,0,0,0,0,0,1,1,0,1,0,0,0,0,0,0] $$
$$ \ell_2=[1,1,1,1,0,1,1,0,1,0,0,1,0,0,1,1,0,0,1,1,1,1,0,0,0,0,0,0,0,0,0,0,1,1,0,1,0,0,0,0,0,0] $$
$$ \ell_3=[1,1,1,1,0,1,1,0,0,0,0,1,0,0,1,1,0,0,1,1,1,1,0,0,0,0,0,0,0,0,0,0,1,0,1,1,0,0,0,0,0,0] $$
$$ \ell_4=[1,1,1,1,0,1,1,0,1,0,0,1,0,0,1,1,0,0,1,1,1,1,0,0,0,0,0,0,0,0,0,0,1,0,1,1,0,0,0,0,0,0] $$
$$ \ell_5=[1,1,1,1,0,1,1,0,1,0,0,1,0,0,1,1,0,0,1,1,1,1,0,0,0,0,0,0,0,0,0,0,0,1,1,1,0,0,0,0,0,0] $$
$$ \ell_6=[1,1,1,1,0,1,1,0,1,0,0,1,0,0,1,1,0,0,1,0,0,1,0,0,0,0,0,0,0,0,0,0,1,1,1,1,0,0,0,0,0,0] $$
$$ \ell_7=[1,1,1,1,0,1,1,0,1,0,0,1,0,0,1,1,0,0,1,1,0,1,0,0,0,0,0,0,0,0,0,0,1,1,1,1,0,0,0,0,0,0] $$}
Notice how one is left with a shorter list of only $42$ parameters, where most of them have the
same value for all the languages in this group. 
The only non-zero frequencies for binary vectors $(a_1,\ldots,a_7)\in \F_2^7$ that arise
in the boundary distribution at the leaves of the trees are
$$ \begin{array}{cccc} n_{1111111}=12 & n_{0000000}=24 & n_{1101111}=1 & n_{1111101}=1 \\
n_{1111100}=1 & n_{1111011}=1 & n_{1100111}=1 & n_{0011111}=1 \end{array} $$ 
with probabilities
$$ \begin{array}{cccc} p_{1111111}=\frac{2}{7} & p_{0000000}=\frac{4}{7} & p_{1101111}=\frac{1}{42} & p_{1111101}=\frac{1}{42} \\[3mm]
p_{1111100}=\frac{1}{42} & p_{1111011}=\frac{1}{42} & p_{1100111}=\frac{1}{42} & p_{0011111}=\frac{1}{42} \end{array} $$ 
and all other $p_{a_1\cdots a_7}=0$.

We need to consider Flattenings of the boundary tensor $P=(p_{a_1\cdots a_7})$
of the form 
\begin{enumerate}
\item ${\rm Flat}_{\{ \ell_5, \ell_6, \ell_7\}\cup \{ \ell_1, \ell_2, \ell_3, \ell_4\}}$
\item ${\rm Flat}_{\{ \ell_1, \ell_2, \ell_3\}\cup \{ \ell_4, \ell_5, \ell_6, \ell_7\}}$
\item ${\rm Flat}_{\{ \ell_1, \ell_2, \ell_4\}\cup \{ \ell_3, \ell_5, \ell_6, \ell_7\}}$
\item ${\rm Flat}_{\{ \ell_1, \ell_2, \ell_5\}\cup \{ \ell_3, \ell_4, \ell_6, \ell_7\}}$
\item ${\rm Flat}_{\{ \ell_4, \ell_6, \ell_7\}\cup \{ \ell_1, \ell_2, \ell_3, \ell_5\}}$
\item ${\rm Flat}_{\{ \ell_3, \ell_6, \ell_7\}\cup \{ \ell_1, \ell_2, \ell_4, \ell_5\}}$
\end{enumerate}
Note that we do not need to consider the flattenings
${\rm Flat}_{\{ \ell_6, \ell_7\}\cup \{ \ell_1, \ell_2, \ell_3, \ell_4, \ell_5\}}$ and
${\rm Flat}_{\{ \ell_1, \ell_2 \}\cup \{ \ell_3, \ell_4, \ell_5, \ell_6, \ell_6\}}$, as these
are common to all the candidate trees and would not help discriminating between them.

\smallskip

\begin{center}
\begin{figure}[h]
\includegraphics[scale=0.45]{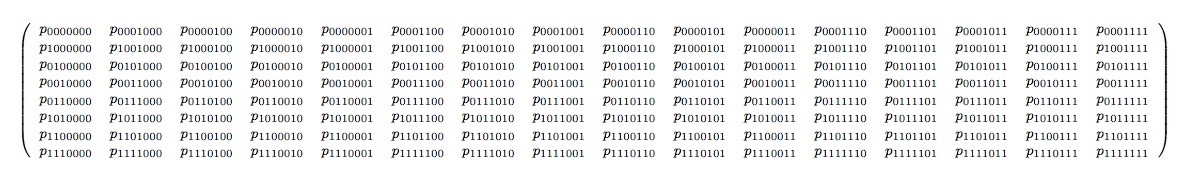} 
\caption{Flattenings $8\times 16$ matrices. \label{MatrGerFig}}
\end{figure}
\end{center}

All the flattenings above correspond to $8 \times 16$ matrices as in Figure~\ref{MatrGerFig},
where in each of the cases listed above the matrix indices $(abcdefg)$ correspond, respectively, to
\begin{enumerate}
\item $(abcdefg)=(a_5 a_6 a_7 a_1 a_2 a_3 a_4)$
\item $(abcdefg)=(a_1 a_2 a_3 a_4 a_5 a_6 a_7)$
\item $(abcdefg)=(a_1 a_2 a_4 a_3 a_5 a_6 a_7)$
\item $(abcdefg)=(a_1 a_2 a_5 a_3 a_4 a_6 a_7)$
\item $(abcdefg)=(a_4 a_6 a_7 a_1 a_2 a_3 a_5)$
\item $(abcdefg)=(a_3 a_6 a_7 a_1 a_2 a_4 a_5)$
\end{enumerate}

The probability distributions corresponding to the permutations listed above
are respectively given by
\begin{enumerate}
\item $n_{1111101}=1$, $n_{1011111}=1$, $n_{1001111}=1$, $n_{0111111}=1$, $n_{1111100}=1$, $n_{1110011}=1$
\item $n_{1101111}=1$, $n_{1111101}=1$, $n_{1111100}=1$, $n_{1111011}=1$, $n_{1100111}=1$, $n_{0011111}=1$
\item $n_{1110111}=1$, $n_{1111101}=1$, $n_{1111100}=1$, $n_{1111011}=1$, $n_{1100111}=1$, $n_{0011111}=1$
\item $n_{1110111}=1$, $n_{1111101}=1$, $n_{1111100}=1$, $n_{1101111}=1$, $n_{1110011}=1$, $n_{0011111}=1$
\item $n_{1111101}=1$, $n_{1011111}=1$, $n_{1001111}=1$, $n_{1111110}=1$, $n_{0111101}=1$, $n_{1110011}=1$
\item $n_{0111111}=1$, $n_{1011111}=1$, $n_{1001111}=1$, $n_{1111110}=1$, $n_{0111101}=1$, $n_{1110011}=1$
\end{enumerate}
while all six cases have the common values $n_{1111111}=12$ and $n_{0000000}=24$.

The corresponding flattening matrices are given by
$$ {\rm Flat}_{\{ \ell_5, \ell_6, \ell_7\}\cup \{ \ell_1, \ell_2, \ell_3, \ell_4\}}(P)=  \left( \begin{array}{cccccccccccccccc}
\frac{4}{7} & 0 & 0 & 0 & 0 &  0 & 0 & 0 & 0 & 0 & 0 & 0 & 0 & 
0 & 0 & 0 \\
0 & 0 & 0 & 0 & 0 &
0 & 0 & 0 & 0 & 0 & 
0 & 0 & 0 & 0 & 0 & \frac{1}{42} \\
0 & 0 & 0 & 0 & 0 & 0 &
0 & 0 & 0 & 0 & 0 & 0 &
0 & 0 & 0 & 0 \\
0 & 0 & 0 & 0 & 0 & 0 &
0 & 0 & 0 & 0 & 0 & 0 &
0 & 0 & 0 & 0 \\
0 & 0 & 0 & 0 & 0 & 0 &
0 & 0 & 0 & 0 & 0 & 0 &
0 & 0 & 0 & \frac{1}{42} \\
0 & 0 & 0 & 0 & 0 & 0 &
0 & 0 & 0 & 0 & 0 & 0 &
0 & 0 & 0 & \frac{1}{42} \\
0 & 0 & 0 & 0 & 0 & 0 &
0 & 0 & 0 & 0 & 0 & 0 &
0 & 0 & 0 & 0 \\
0 & 0 & 0 & 0 & 0 & \frac{1}{42} &
0 & 0 & 0 & 0 & \frac{1}{42} & 0 &
\frac{1}{42} & 0 & 0 & \frac{2}{7} 
\end{array}\right)$$

$${\rm Flat}_{\{ \ell_1, \ell_2, \ell_3\}\cup \{ \ell_4, \ell_5, \ell_6, \ell_7\}}(P)=  \left( \begin{array}{cccccccccccccccc}
\frac{4}{7} & 0 & 0 & 0 & 0 &  0 & 0 & 0 & 0 & 0 & 0 & 0 & 0 & 
0 & 0 & 0 \\
0 & 0 & 0 & 0 & 0 &
0 & 0 & 0 & 0 & 0 & 
0 & 0 & 0 & 0 & 0 & 0 \\
0 & 0 & 0 & 0 & 0 & 0 &
0 & 0 & 0 & 0 & 0 & 0 &
0 & 0 & 0 & 0 \\
0 & 0 & 0 & 0 & 0 & 0 &
0 & 0 & 0 & 0 & 0 & 0 &
0 & 0 & 0 & \frac{1}{42} \\
0 & 0 & 0 & 0 & 0 & 0 &
0 & 0 & 0 & 0 & 0 & 0 &
0 & 0 & 0 & 0 \\
0 & 0 & 0 & 0 & 0 & 0 &
0 & 0 & 0 & 0 & 0 & 0 &
0 & 0 & 0 & 0 \\
0 & 0 & 0 & 0 & 0 & 0 &
0 & 0 & 0 & 0 & 0 & 0 &
0 & 0 & \frac{1}{42} & \frac{1}{42} \\
0 & 0 & 0 & 0 & 0 & \frac{1}{42} &
0 & 0 & 0 & 0 & 0 & 0 &
\frac{1}{42} & \frac{1}{42} & 0 & \frac{2}{7}
\end{array}\right) $$

$${\rm Flat}_{\{ \ell_1, \ell_2, \ell_4\}\cup \{ \ell_3, \ell_5, \ell_6, \ell_7\}}(P) =  \left( \begin{array}{cccccccccccccccc}
\frac{4}{7} & 0 & 0 & 0 & 0 &  0 & 0 & 0 & 0 & 0 & 0 & 0 & 0 & 
0 & 0 & 0 \\
0 & 0 & 0 & 0 & 0 &
0 & 0 & 0 & 0 & 0 & 
0 & 0 & 0 & 0 & 0 & 0 \\
0 & 0 & 0 & 0 & 0 & 0 &
0 & 0 & 0 & 0 & 0 & 0 &
0 & 0 & 0 & 0 \\
0 & 0 & 0 & 0 & 0 & 0 &
0 & 0 & 0 & 0 & 0 & 0 &
0 & 0 & 0 & \frac{1}{42} \\
0 & 0 & 0 & 0 & 0 & 0 &
0 & 0 & 0 & 0 & 0 & 0 &
0 & 0 & 0 & 0 \\
0 & 0 & 0 & 0 & 0 & 0 &
0 & 0 & 0 & 0 & 0 & 0 &
0 & 0 & 0 & 0 \\
0 & 0 & 0 & 0 & 0 & 0 &
0 & 0 & 0 & 0 & 0 & 0 &
0 & 0 & \frac{1}{42} & 0 \\
0 & 0 & 0 & 0 & 0 & \frac{1}{42} &
0 & 0 & 0 & 0 & 0 & 0 &
\frac{1}{42} & \frac{1}{42} & \frac{1}{42} & \frac{2}{7}
\end{array}\right) $$ 

$${\rm Flat}_{\{ \ell_1, \ell_2, \ell_5\}\cup \{ \ell_3, \ell_4, \ell_6, \ell_7\}}(P) =  \left( \begin{array}{cccccccccccccccc}
\frac{4}{7} & 0 & 0 & 0 & 0 &  0 & 0 & 0 & 0 & 0 & 0 & 0 & 0 & 
0 & 0 & 0 \\
0 & 0 & 0 & 0 & 0 &
0 & 0 & 0 & 0 & 0 & 
0 & 0 & 0 & 0 & 0 & 0 \\
0 & 0 & 0 & 0 & 0 & 0 &
0 & 0 & 0 & 0 & 0 & 0 &
0 & 0 & 0 & 0 \\
0 & 0 & 0 & 0 & 0 & 0 &
0 & 0 & 0 & 0 & 0 & 0 &
0 & 0 & 0 & \frac{1}{42} \\
0 & 0 & 0 & 0 & 0 & 0 &
0 & 0 & 0 & 0 & 0 & 0 &
0 & 0 & 0 & 0 \\
0 & 0 & 0 & 0 & 0 & 0 &
0 & 0 & 0 & 0 & 0 & 0 &
0 & 0 & 0 & 0 \\
0 & 0 & 0 & 0 & 0 & 0 &
0 & 0 & 0 & 0 & 0 & 0 &
0 & 0 & 0 & \frac{1}{42} \\
0 & 0 & 0 & 0 & 0 & \frac{1}{42} &
0 & 0 & 0 & 0 & \frac{1}{42} & 0 &
\frac{1}{42} & 0 & \frac{1}{42} & \frac{2}{7} 
\end{array}\right)$$

$${\rm Flat}_{\{ \ell_4, \ell_6, \ell_7\}\cup \{ \ell_1, \ell_2, \ell_3, \ell_5\}}(P) =  \left( \begin{array}{cccccccccccccccc}
\frac{4}{7} & 0 & 0 & 0 & 0 &  0 & 0 & 0 & 0 & 0 & 0 & 0   & 0 &  0 & 0 & 0 \\
0 & 0 & 0 & 0 & 0 &
0 & 0 & 0 & 0 & 0 & 
0 & 0 & 0 & 0 & 0 & \frac{1}{42} \\
0 & 0 & 0 & 0 & 0 & 0 &
0 & 0 & 0 & 0 & 0 & 0 &
0 & 0 & 0 & 0  \\
0 & 0 & 0 & 0 & 0 & 0 &
0 & 0 &0 & 0 & 0 & 0 &
0 & 0 & 0 & 0 \\
0 & 0 & 0 & 0 & 0 & 0 &
0& 0 & 0 & 0 & 0 & 0 &
0 & 0& 0& 0 \\
0 & 0 & 0 & 0 & 0 &0 &
0 & 0&0& 0 & 0 & 0 &
0 & 0 & 0 & \frac{1}{42} \\
0 & 0 & 0 & 0 & 0 & 0 &
0 &0 & 0 & 0& 0 & 0 &
0 & 0 & 0 & 0 \\
0 & 0 &0 & 0& 0& \frac{1}{42} &
0 & 0 & 0 & 0& \frac{1}{42} & \frac{1}{42} &
\frac{1}{42} & 0& 0 & \frac{2}{7} 
\end{array}\right)$$

$${\rm Flat}_{\{ \ell_3, \ell_6, \ell_7\}\cup \{ \ell_1, \ell_2, \ell_4, \ell_5\}}(P) =  \left( \begin{array}{cccccccccccccccc}
\frac{4}{7} & 0 & 0 & 0 & 0 & 0 & 0 & 0 & 0 & 0 & 0 & 0 & 0 & 0 & 0 & 0 \\
              0 & 0 & 0 & 0 & 0 & 0 & 0 & 0 & 0 & 0 & 0 & 0 & 0 & 0 & 0 & \frac{1}{42}  \\
              0 & 0 & 0 & 0 & 0 & 0 & 0 & 0 & 0 & 0 & 0 & 0 & 0 & 0 & 0 & 0 \\
              0 & 0 & 0 & 0 & 0 & 0 & 0 & 0 & 0 & 0 & 0 & 0 & 0 & 0 & 0 & 0 \\
              0 & 0 & 0 & 0 & 0 & 0 & 0 & 0 & 0 & 0 & 0 & 0 & \frac{1}{42} & 0 & 0 & \frac{1}{42} \\
              0  & 0 & 0 & 0 & 0 & 0 & 0 & 0 & 0 & 0 & 0 & 0 & 0 & 0 & 0 & \frac{1}{42} \\
              0 & 0 & 0 & 0 & 0 & 0 & 0 & 0 & 0 & 0 & 0 & 0 & 0 & 0 & 0 & 0 \\
              0 & 0 & 0 & 0 & 0 & 0 & 0 & 0 & 0 & 0 & \frac{1}{42} & \frac{1}{42} & 0 & 0 & 0 & \frac{2}{7} 
\end{array}\right) $$

\subsubsection{Comparison of the remaining two trees}

The trees $T_5(G)$ and $T_6(G)$ have a slightly different structure, since in addition to placing
in closest proximity the pairs $\{ \ell_1, \ell_2 \}$ and $\{ \ell_6, \ell_7 \}$ like all other trees they
also identify pairs $\{ \ell_4, \ell_5 \}$ in the case of $T_5(G)$ and $\{ \ell_3, \ell_5 \}$ in the case
of $T_6(G)$. Thus, while these two trees also have the flattenings 
${\rm Flat}_{\{ \ell_6, \ell_7\}\cup \{ \ell_1, \ell_2, \ell_3, \ell_4, \ell_5\}}$ and
${\rm Flat}_{\{ \ell_1, \ell_2 \}\cup \{ \ell_3, \ell_4, \ell_5, \ell_6, \ell_6\}}$ common to all the other trees,
they also have a flattening
$$ {\rm Flat}_{\{\ell_3,\ell_4,\ell_5 \}\cup \{ \ell_1,\ell_2,\ell_6,\ell_7\}} $$
common to both trees $T_5(G)$ and $T_6(G)$ and
$$ \begin{array}{lll}
F_5:= &{\rm Flat}_{\{ \ell_4, \ell_5 \}\cup \{ \ell_1,\ell_2,\ell_3,\ell_6,\ell_7 \}} & \text{ for } \ T_5(G) \\[2mm]
 F_6:= & {\rm Flat}_{\{\ell_3,\ell_5\} \cup \{ \ell_1,\ell_2,\ell_4,\ell_6,\ell_7 \}}  & \text{ for } \ T_6(G).
 \end{array} $$
 We have as corresponding matrices 
 $$ {\rm Flat}_{\{\ell_3,\ell_4,\ell_5 \}\cup \{ \ell_1,\ell_2,\ell_6,\ell_7\}}(P) = 
 \left( \begin{array}{cccccccccccccccc}
 \frac{4}{7} &0 &0&0&0&0&0&0&0&0&0&0&0&0&0&0 \\
 0&0&0&0&0&0&0&0&0&0&0&0&0&0&0&0 \\
 0&0&0&0&0&0&0&0&0&0&0&0&0&0&0&0 \\
 0&0&0&0&0&0&0&0&0&0&0&0&0&0&0&\frac{1}{42} \\
 0&0&0&0&0&0&0&0&0&0&0&0&0&0&0&\frac{1}{42} \\
 0&0&0&0&0&0&0&0&0&0&0&0&0&0&0&\frac{1}{42} \\
 0&0&0&0&0&0&0&0&0&0&0&0&0&0&0&0 \\
 0&0&0& \frac{1}{42} & 0& \frac{1}{42} & 0&0& 0&0&0&0&0&0& \frac{1}{42} & \frac{2}{7}
 \end{array}\right) $$
while the matrices (written in transpose form) for $F_5$ and $F_6$ are given in Appendix~C.

\smallskip
\subsection{Computation of the phylogenetic invariants}

We compute the phylogenetic invariants using the $\ell^\infty$ and the $\ell^1$ norm.
\begin{enumerate}
\item $T_1(G)$ with flattenings 
$M_1={\rm Flat}_{\{ \ell_5, \ell_6, \ell_7\}\cup \{ \ell_1, \ell_2, \ell_3, \ell_4\}}$ and
$M_2={\rm Flat}_{\{ \ell_1, \ell_2, \ell_3\}\cup \{ \ell_4, \ell_5, \ell_6, \ell_7\}}$  gives:
$$ \| \phi_{T_1}(P)\|_{\ell^\infty} =\max\{ \max_{\substack{\phi \in \in 3\times 3 \text{minors}
\\ \text{ of } M_1}} \left| \phi(P) \right| \, , \, \max_{\substack{\phi \in \in 3\times 3 \text{minors}
\\ \text{ of } M_2}} \left| \phi(P) \right| \}= \frac{4}{1029} $$
$$ \| \phi_{T_1}(P)\|_{\ell^1} =\sum_{\substack{\phi \in \in 3\times 3 \text{minors}
\\ \text{ of } M_1}} \left| \phi(P) \right| + \sum_{\substack{\phi \in \in 3\times 3 \text{minors}
\\ \text{ of } M_2}} \left| \phi(P) \right| =\frac{ 83}{8232} $$

\item $T_2(G)$ with flattenings  
$M_1={\rm Flat}_{\{ \ell_5, \ell_6, \ell_7\}\cup \{ \ell_1, \ell_2, \ell_3, \ell_4\}}$ and
$M_3={\rm Flat}_{\{ \ell_1, \ell_2, \ell_4\}\cup \{ \ell_3, \ell_5, \ell_6, \ell_7\}}$ gives
$$ \| \phi_{T_2}(P)\|_{\ell^\infty} =\max\{ \max_{\substack{\phi \in \in 3\times 3 \text{minors}
\\ \text{ of } M_1}} \left| \phi(P) \right| \, , \, \max_{\substack{\phi \in \in 3\times 3 \text{minors}
\\ \text{ of } M_3}} \left| \phi(P) \right| \}=  \frac{ 4}{1029 } $$
$$ \| \phi_{T_2}(P)\|_{\ell^1} =\sum_{\substack{\phi \in \in 3\times 3 \text{minors}
\\ \text{ of } M_1}} \left| \phi(P) \right| + \sum_{\substack{\phi \in \in 3\times 3 \text{minors}
\\ \text{ of } M_3}} \left| \phi(P) \right| = \frac{ 233}{24696 } $$

\item $T_3(G)$ with flattenings
$M_4={\rm Flat}_{\{ \ell_1, \ell_2, \ell_5\}\cup \{ \ell_3, \ell_4, \ell_6, \ell_7\}}$
and $M_5={\rm Flat}_{\{ \ell_4, \ell_6, \ell_7\}\cup \{ \ell_1, \ell_2, \ell_3, \ell_5\}}$ gives
$$ \| \phi_{T_3}(P)\|_{\ell^\infty} =\max\{ \max_{\substack{\phi \in \in 3\times 3 \text{minors}
\\ \text{ of } M_4}} \left| \phi(P) \right| \, , \, \max_{\substack{\phi \in \in 3\times 3 \text{minors}
\\ \text{ of } M_5}} \left| \phi(P) \right| \}=  \frac{ 1}{3087 } $$
$$ \| \phi_{T_3}(P)\|_{\ell^1} =\sum_{\substack{\phi \in \in 3\times 3 \text{minors}
\\ \text{ of } M_4}} \left| \phi(P) \right| + \sum_{\substack{\phi \in \in 3\times 3 \text{minors}
\\ \text{ of } M_5}} \left| \phi(P) \right| = \frac{16}{3087} $$

\item $T_4(G)$ with flattenings 
$M_4={\rm Flat}_{\{ \ell_1, \ell_2, \ell_5\}\cup \{ \ell_3, \ell_4, \ell_6, \ell_7\}}$ and
$M_6={\rm Flat}_{\{ \ell_3, \ell_6, \ell_7\}\cup \{ \ell_1, \ell_2, \ell_4, \ell_5\}}$ gives 
$$ \| \phi_{T_4}(P)\|_{\ell^\infty} =\max\{ \max_{\substack{\phi \in \in 3\times 3 \text{minors}
\\ \text{ of } M_4}} \left| \phi(P) \right| \, , \, \max_{\substack{\phi \in \in 3\times 3 \text{minors}
\\ \text{ of } M_6}} \left| \phi(P) \right| \}=  \frac{ 4}{1029 }   $$
$$ \| \phi_{T_4}(P)\|_{\ell^1} =\sum_{\substack{\phi \in \in 3\times 3 \text{minors}
\\ \text{ of } M_4}} \left| \phi(P) \right| + \sum_{\substack{\phi \in \in 3\times 3 \text{minors}
\\ \text{ of } M_6}} \left| \phi(P) \right| = \frac{181}{18522} $$

\item $T_5(G)$ with flattenings $F_5$ of Appendix C and 
$M_7={\rm Flat}_{\{\ell_3,\ell_4,\ell_5 \}\cup \{ \ell_1,\ell_2,\ell_6,\ell_7\}}$ gives
$$ \| \phi_{T_5}(P)\|_{\ell^\infty} =\max\{ \max_{\substack{\phi \in \in 3\times 3 \text{minors}
\\ \text{ of } F_5}} \left| \phi(P) \right| \, , \, \max_{\substack{\phi \in \in 3\times 3 \text{minors}
\\ \text{ of } M_7}} \left| \phi(P) \right| \}=   \frac{ 4}{1029 }   $$
$$ \| \phi_{T_5}(P)\|_{\ell^1} =\sum_{\substack{\phi \in \in 3\times 3 \text{minors}
\\ \text{ of } F_5}} \left| \phi(P) \right| + \sum_{\substack{\phi \in \in 3\times 3 \text{minors}
\\ \text{ of } M_7}} \left| \phi(P) \right| = \frac{ 233}{24696 } $$

\item $T_6(G)$ with flattenings $F_6$ of Appendix C and 
$M_7={\rm Flat}_{\{\ell_3,\ell_4,\ell_5 \}\cup \{ \ell_1,\ell_2,\ell_6,\ell_7\}}$ gives
$$ \| \phi_{T_6}(P)\|_{\ell^\infty} =\max\{ \max_{\substack{\phi \in \in 3\times 3 \text{minors}
\\ \text{ of } F_6}} \left| \phi(P) \right| \, , \, \max_{\substack{\phi \in \in 3\times 3 \text{minors}
\\ \text{ of } M_7}} \left| \phi(P) \right| \}=   \frac{ 4}{1029 }   $$
$$ \| \phi_{T_6}(P)\|_{\ell^1} =\sum_{\substack{\phi \in \in 3\times 3 \text{minors}
\\ \text{ of } F_6}} \left| \phi(P) \right| + \sum_{\substack{\phi \in \in 3\times 3 \text{minors}
\\ \text{ of } M_7}} \left| \phi(P) \right| = \frac{ 83}{8232 } $$
\end{enumerate}

In this case we see that both the $\ell^\infty$ and the $\ell^1$ norm provide a good test
that selects the historically correct tree $T_3(G)$. Note that the $\ell^\infty$ 
has the same value $4/1029$ on all the other candidates and the lower value 
$1/3087$ only for the correct tree $T_3(G)$.

\smallskip
\subsection{Estimates of Euclidean distance for the $\cS_2(G)$ Germanic languages}\label{Var2Sec}

We obtain an estimate of the likelihood of the candidate trees based on 
computing a lower bound for the Euclidean distance in terms of distances between
the flattening matrices ${\rm Flat}_e(P)$ of the boundary distribution $P$ and the determinantal
varieties they are expected to lie on. More concretely, we have the following:
\begin{enumerate}
\item The Euclidean distance estimate for the tree $T_1(G)$ is given by ${\rm dist}(P,V_{T_1})\geq L_1$
$$ L_1=\max\{ d({\rm Flat}_{\{ \ell_1, \ell_2, \ell_3\}\cup \{ \ell_4, \ell_5, \ell_6, \ell_7\}}(P),\cD_2(8,16)),
d({\rm Flat}_{\{ \ell_5, \ell_6, \ell_7\}\cup \{ \ell_1, \ell_2, \ell_3, \ell_4\}}(P),\cD_2(8,16))\} $$
\item The Euclidean distance estimate of $T_2(G)$ is given by ${\rm dist}(P,V_{T_2})\geq L_2$
$$ L_2=\max \{ d({\rm Flat}_{\{ \ell_1, \ell_2, \ell_4\}\cup \{ \ell_3, \ell_5, \ell_6, \ell_7\}}(P),\cD_2(8,16)),
d({\rm Flat}_{\{ \ell_5, \ell_6, \ell_7\}\cup \{ \ell_1, \ell_2, \ell_3, \ell_4\}}(P),\cD_2(8,16)) \} $$
\item The Euclidean distance estimate of $T_3(G)$ is given by ${\rm dist}(P,V_{T_3})\geq L_3$
$$ L_3=\max \{ d(  {\rm Flat}_{\{ \ell_1, \ell_2, \ell_5\}\cup \{ \ell_3, \ell_4, \ell_6, \ell_7\}}(P), \cD_2(8,16)),
d({\rm Flat}_{\{ \ell_4, \ell_6, \ell_7\}\cup \{ \ell_1, \ell_2, \ell_3, \ell_5\}}(P) ,\cD_2(8,16)) \} $$
\item The Euclidean distance estimate of $T_4(G)$ is given by ${\rm dist}(P,V_{T_4})\geq L_4$
$$  L_4=\max \{ d({\rm Flat}_{\{ \ell_1, \ell_2, \ell_5\}\cup \{ \ell_3, \ell_4, \ell_6, \ell_7\}}(P), \cD_2(8,16)),
d({\rm Flat}_{\{ \ell_3, \ell_6, \ell_7\}\cup \{ \ell_1, \ell_2, \ell_4, \ell_5\}}(P), \cD_2(8,16))\} $$
\end{enumerate}

The singular value decomposition of the flattening matrices gives $\Sigma={\rm diag}(\sigma_1,\ldots,\sigma_8)$ with
$$ \Sigma({\rm Flat}_{\{ \ell_5, \ell_6, \ell_7\}\cup \{ \ell_1, \ell_2, \ell_3, \ell_4\}}(P)) \sim $$
$$ {\rm diag}( 0.57143, 0.291548, 0.58333 \times 10^{-2}, 0.12240 \times 10^{-17}, 
0.10572 \times 10^{-34}, 0.16149 \times 10^{-51}, 0.63652 \times 10^{-68}, 0) $$
$$ \Sigma({\rm Flat}_{\{ \ell_1, \ell_2, \ell_3\}\cup \{ \ell_4, \ell_5, \ell_6, \ell_7\}})(P)) \sim $$
$$ {\rm diag}( 0.57143, 0.29059, 0.23973 \times 10^{-1}, 0.33558  \times 10^{-2},
0.64145 \times 10^{-19}, 0.60260 \times 10^{-31}, 0, 0) $$
$$ \Sigma({\rm Flat}_{\{ \ell_1, \ell_2, \ell_4\}\cup \{ \ell_3, \ell_5, \ell_6, \ell_7\}}(P)) \sim $$
$$ {\rm diag}( 0.57143, 0.29061, 0.23809 \times 10^{-1}, 0.33787 \times 10^{-2}, 0,0,0,0) $$
$$ \Sigma({\rm Flat}_{\{ \ell_1, \ell_2, \ell_5\}\cup \{ \ell_3, \ell_4, \ell_6, \ell_7\}}(P)) \sim $$
$$ {\rm diag}( 0.57143, 0.29155, 0.54996  \times 10^{-2},0,0,0,0,0) $$
$$ \Sigma({\rm Flat}_{\{ \ell_4, \ell_6, \ell_7\}\cup \{ \ell_1, \ell_2, \ell_3, \ell_5\}}(P)) \sim $$
$$ {\rm diag}( 0.57143, 0.29155, 0.54996 \times 10^{-2},0,0,0,0,0) $$
$$ \Sigma({\rm Flat}_{\{ \ell_3, \ell_6, \ell_7\}\cup \{ \ell_1, \ell_2, \ell_4, \ell_5\}}(P)) \sim $$
$$ {\rm diag}( 0.57143, 0.29059, 0.23892 \times 10^{-1}, 0.38881 \times 10^{-2},
0.12435 \times 10^{-17}, 0.73417 \times 10^{-19}, 0.32257 \times 10^{-34}, 0). $$

By the Eckart-Young theorem we then have
$$ d({\rm Flat}_{\{ \ell_5, \ell_6, \ell_7\}\cup \{ \ell_1, \ell_2, \ell_3, \ell_4\}}(P),\cD_2(8,16))^2=\sigma_3^2+\cdots+\sigma_8^2= 0.34027 \times 10^{-4} $$
$$ d({\rm Flat}_{\{ \ell_1, \ell_2, \ell_3\}\cup \{ \ell_4, \ell_5, \ell_6, \ell_7\}})(P),\cD_2(8,16))^2=\sigma_3^2+\cdots+\sigma_8^2= 0.58597 \times 10^{-3} $$
$$ d({\rm Flat}_{\{ \ell_1, \ell_2, \ell_4\}\cup \{ \ell_3, \ell_5, \ell_6, \ell_7\}}(P),\cD_2(8,16))^2=\sigma_3^2+\cdots+\sigma_8^2= 0.57831 \times 10^{-3} $$
$$ d({\rm Flat}_{\{ \ell_1, \ell_2, \ell_5\}\cup \{ \ell_3, \ell_4, \ell_6, \ell_7\}}(P),\cD_2(8,16))^2= \sigma_3^2+\cdots+\sigma_8^2= 0.30245  \times 10^{-4} $$
$$ d({\rm Flat}_{\{ \ell_4, \ell_6, \ell_7\}\cup \{ \ell_1, \ell_2, \ell_3, \ell_5\}}(P),\cD_2(8,16))^2=\sigma_3^2+\cdots+\sigma_8^2= 0.30245 \times 10^{-4} $$
$$ d({\rm Flat}_{\{ \ell_3, \ell_6, \ell_7\}\cup \{ \ell_1, \ell_2, \ell_4, \ell_5\}}(P),\cD_2(8,16))^2=\sigma_3^2+\cdots+\sigma_8^2= 0.58595 \times 10^{-3} .$$
Thus, we obtain
$$ L_1 = 0.58597 \times 10^{-3}, \ \ \  L_2= 0.57831 \times 10^{-3}, \ \ \ L_3 = 0.30245 \times 10^{-4}, \ \ \ 
L_4 = 0.58595 \times 10^{-3} . $$

\smallskip

Thus, both the computation of the phylogenetic invariants and the Euclidean distance estimate
select the tree $T_3$ as the preferred candidate
phylogenetic tree, which is indeed the closest to what is regarded as the correct linguistic
phylogenetic tree.

\subsubsection{Euclidean distance for  $T_5$ and $T_6$} We discuss the Euclidean distance estimate for the 
two remaining trees. To this purpose, we compute the Euclidean distance of the 
flattening ${\rm Flat}_{\{\ell_3,\ell_4,\ell_5 \}\cup \{ \ell_1,\ell_2,\ell_6,\ell_7\}}(P)$
from  the determinantal variety $\cD(8,16)$, with $P$ given
by the distribution at the leaves based on the Longobardi data. We find
$$ \Sigma({\rm Flat}_{\{\ell_3,\ell_4,\ell_5 \}\cup \{ \ell_1,\ell_2,\ell_6,\ell_7\}}(P)) \sim $$
$$ {\rm diag}( 
0.57143 , 0.29155 , 0.58333 \times 10^{-2}  , 0.18608 \times 10^{-17}, 0.32093 \times 10^{-33},
0,0,0) $$
$$ d({\rm Flat}_{\{\ell_3,\ell_4,\ell_5 \}\cup \{ \ell_1,\ell_2,\ell_6,\ell_7\}}(P),\cD_2(8,16))^2=
\sigma_3^2+\cdots+\sigma_8^2= 0.34027 \times 10^{-4}. $$

\smallskip

We then compute the distances between the flattenings $F_5$ and $F_6$ and the
determinantal variety $\cD_2(4,32)$. 
The singular values of $F_5$ are given by
$$ \Sigma(F_5(P))=( 0.57143 , 0.29061 , 0.23809 \times 10^{-1} , 0.33787 \times 10^{-2}  ) $$
which give the distance
$$ d(F_5(P),\cD_2(4,32))^2 = \sigma_3^2 + \sigma_4^2 = 0.57831 \times 10^{-3}. $$
The singular values for $F_6$ are
$$ \Sigma(F_5(P))=( 0.57143 , 0.29060 , 0.23973 \times 10^{-1}, 0.33558 \times 10^{-2}) $$
which gives the distance
$$ d(F_6(P),\cD_2(4,32))^2 = \sigma_3^2 + \sigma_4^2 = 0.58597  \times 10^{-3} $$
Thus, the lower bound for the Euclidean distance for the tree $T_5(G)$ is given by
$$ \max\{ d({\rm Flat}_{\{\ell_3,\ell_4,\ell_5 \}\cup \{ \ell_1,\ell_2,\ell_6,\ell_7\}}(P),\cD_2(8,16))^2, 
d(F_5(P),\cD_2(4,32))^2  \} = 0.57831 \times 10^{-3} $$
and similarly the lower bound for the Euclidean distance for the tree $T_6(G)$ is given by
$$ \max\{ d({\rm Flat}_{\{\ell_3,\ell_4,\ell_5 \}\cup \{ \ell_1,\ell_2,\ell_6,\ell_7\}}(P),\cD_2(8,16))^2, 
d(F_6(P),\cD_2(4,32))^2  \} =0.58597  \times 10^{-3}. $$

\smallskip

When we compare the estimates for the trees $T_5(G)$ and $T_6(G)$ with the
ones obtained for the previous trees $T_1(G),\ldots, T_4(G)$ we see that these lower
bounds are comparable to those of $T_1(G),T_2(G),T_4(G)$ and only $T_3(G)$
has a significantly smaller estimate. Thus, this criterion also suggests $T_3(G)$
as the most favorable candidate.

\smallskip
\subsection{Comparison with SSWL data}

\begin{center}
\begin{figure}[h]
\includegraphics[scale=0.85]{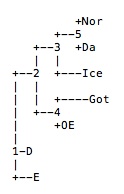} \  \   \  \   \
\includegraphics[scale=0.85]{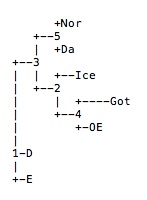}
\caption{PHYLIP output trees of Germanic languages for the set $\cS_2(G)$ based on combined
Longobardi and SSWL data. \label{FigS2G}}
\end{figure}
\end{center}

The DNA parsimony algorithm of PHYLIP produced two candidate phylogenetic trees for the
set $\cS_2(G)$ of Germanic languages based on the combination of the Longobardi data 
and the SSWL data. They are shown in Figure~\ref{FigS2G}.

In this case, the inclusion of the additional SSWL data resolves the ambiguity of the tree of
Figure~\ref{FigS2GLong}. In terms of our treatment of the positioning of the ancient
languages, 
the trees shown in Figure~\ref{FigS2G} should be regarded
as corresponding to the possible trees in cases (3) and (4) discussed above in \S \ref{LongoGerSec}, for
the first tree and cases (5) and (6) for the second one.

\smallskip

Thus, the set of possible binary trees we should consider for a comparison between
the phylogenetic invariants evaluated on the Longobardi and on the SSWL data, consists of the
trees $T_3(G)$ and $T_4(G)$ 
and $T_5(G)$ and $T_6(G)$
of the previous section. We will
evaluate here the phylogenetic invariants and estimate the Euclidean distance function
of these candidate trees (including for completeness also $T_1(G)$ and $T_2(G)$
of the previous section) using the boundary distribution based on the SSWL data.

\smallskip
\subsection{Boundary distribution for $\cS_2(G)$ based on SSWL data}

The Germanic languages in the set $\cS_2(G)$ have a total of 68 SSWL variables that are completely
mapped for all the seven languages in the set. This is significantly smaller than the 90 variables used for the
set $\cS_1(G)$. This does not depend on the languages being poorly mapped: the levels of accuracy are
comparable with the previous set with Danish (76$\%$), Norwegian (75$\%$), 
German (75$\%$), English (75$\%$), Old English (75$\%$) Icelandic (62$\%$), Gothic (62$\%$). However,
the regions of the overall 115 SSWL variables that are mapped is less uniform across this set of
languages creating a smaller overlap. The set of completely mapped SSWL variables for this
set of languages is reported in Appendix~B. 

\medskip

The occurrences of binary vectors at the leaves is given by
$$ \begin{array}{lll}
n_{0,0,0,0,0,0,0} =26 & n_{1,1,1,1,1,1,1}=16 & n_{0,0,1,1,0,0,1}=2 \\
n_{0,0,1,0,0,0,0}=3    & n_{1,1,0,1,0,0,0}=1   & n_{0,0,1,1,1,1,0}=1 \\
n_{0,0,1,1,1,0,0}=1    & n_{0,0,1,0,1,0,0}=1   & n_{1,1,0,1,0,1,1}=2 \\
n_{1,0,1,1,1,0,0}=1    & n_{1,1,1,1,1,0,1}=1   & n_{1,1,1,1,1,0,0}=1 \\
n_{1,1,1,1,0,1,1}=3    & n_{1,1,0,1,1,0,1}=1   & n_{0,0,0,0,1,0,0}=1 \\
n_{1,1,0,0,1,1,1}=1    & n_{0,0,0,0,0,1,0}=1   & n_{0,0,0,1,0,0,0}=2 \\
n_{0,0,0,0,0,0,1}=1    & n_{0,0,1,1,0,0,0}=1   & n_{1,1,0,1,1,1,1}=1 
\end{array} $$
Thus, the boundary probability distribution for the SSWL data for these seven Germanic languages is given by
$$ \begin{array}{lll}
p_{0,0,0,0,0,0,0} =\frac{13}{34} & p_{1,1,1,1,1,1,1}=\frac{4}{17}   & p_{0,0,1,1,0,0,1}=\frac{1}{34} \\[2mm]
p_{0,0,1,0,0,0,0}=\frac{3}{68}    & p_{1,1,0,1,0,0,0}=\frac{1}{68}   & p_{0,0,1,1,1,1,0}=\frac{1}{68} \\[2mm]
p_{0,0,1,1,1,0,0}=\frac{1}{68}    & p_{0,0,1,0,1,0,0}=\frac{1}{68}   & p_{1,1,0,1,0,1,1}=\frac{1}{34} \\[2mm]
p_{1,0,1,1,1,0,0}=\frac{1}{68}    & p_{1,1,1,1,1,0,1}=\frac{1}{68}   & p_{1,1,1,1,1,0,0}=\frac{1}{68} \\[2mm]
p_{1,1,1,1,0,1,1}=\frac{3}{68}    & p_{1,1,0,1,1,0,1}=\frac{1}{68}   & p_{0,0,0,0,1,0,0}=\frac{1}{68} \\[2mm]
p_{1,1,0,0,1,1,1}=\frac{1}{68}    & p_{0,0,0,0,0,1,0}=\frac{1}{68}   & p_{0,0,0,1,0,0,0}=\frac{1}{34} \\[2mm]
p_{0,0,0,0,0,0,1}=\frac{1}{68}    & p_{0,0,1,1,0,0,0}=\frac{1}{68}   & p_{1,1,0,1,1,1,1}=\frac{1}{68} 
\end{array} $$

\medskip

The six flattening matrices corresponding to the different trees of the previous section are in this case of the following form.

$$ {\rm Flat}_{\{ \ell_5, \ell_6, \ell_7\}\cup \{ \ell_1, \ell_2, \ell_3, \ell_4\}}(P)=   
\left( \begin{array}{cccccccccccccccc}
 \frac{13}{34} & 0 & 0 & \frac{3}{68} & \frac{1}{34}  &  0 & 0 & 0 & 0 & 0 & \frac{1}{68} & 0 & \frac{1}{68} & 0 & 0 & 0 \\[2mm]
\frac{1}{68} & 0 & 0 & \frac{1}{68}  & 0 & 0 & 0 & 0 & 0 & 0 & \frac{1}{68} & 0 & 0 & \frac{1}{68}  & 0 & \frac{1}{68} \\[2mm]
\frac{1}{68}  & 0 & 0 & 0 & 0 & 0 & 0 & 0 & 0 & 0 & 0 & 0 & 0 & 0 & 0 & 0 \\[2mm]
\frac{1}{68}  & 0 & 0 & 0 & 0 & 0 & 0 & 0 & 0 & 0 & \frac{1}{34} & 0 & 0 & 0 & 0 & 0 \\[2mm]
0 & 0 & 0 & 0 & 0 & 0 & 0 & 0 & 0 & 0 & 0 & 0 & \frac{1}{34}  & 0 & 0 &   \frac{3}{68}	 \\[2mm]
0 & 0 & 0 & 0 & 0 & 0 & 0 & 0 & 0 & 0 & 0 & 0 & \frac{1}{68} & 0 & 0 & \frac{1}{68}  \\[2mm]
0 & 0 & 0 & 0 & 0 & 0  & 0 & 0 & 0 & 0 & \frac{1}{68} & 0 & 0 & 0  & 0 & 0 \\[2mm]
0 & 0 & 0 & 0 & 0 & \frac{1}{68} & 0 & 0 & 0 & 0 & 0 & 0 & \frac{1}{68} & 0 & 0 & \frac{4}{17}
\end{array}\right)    $$

$${\rm Flat}_{\{ \ell_1, \ell_2, \ell_3\}\cup \{ \ell_4, \ell_5, \ell_6, \ell_7\}}(P)=      
 \left( \begin{array}{cccccccccccccccc}
 \frac{13}{34} & \frac{1}{34} & \frac{1}{68}  & \frac{1}{68}  & \frac{1}{68} &  0 & 0 & 0 & 0 & 0 & 0 & 0 & 0 & 0 & 0 & 0 \\[2mm]
0 & 0 & 0 & 0 & 0 & 0 & 0 & 0 & 0 & 0 & 0 & 0 & 0 & 0 & 0 & 0 \\[2mm]
0 & 0 & 0 & 0 & 0 & 0 & 0 & 0 & 0 & 0 & 0 & 0 & 0 & 0 & 0 & 0 \\[2mm]
\frac{3}{68} & \frac{1}{68}  & \frac{1}{68} & 0 & 0 & \frac{1}{68} & 0 & \frac{1}{34} & 0 & 0 & 0 & \frac{1}{68}  & 0 & 0 & 0 & 0 \\[2mm]
0  & 0 & 0 & 0 & 0 & 0 & 0 & 0 & 0 & 0 & 0 & 0 & 0 & 0 & 0 & 0 \\[2mm]
0 & 0 & 0 & 0 & 0 & \frac{1}{68}  & 0 & 0 & 0 & 0 & 0 & 0 & 0 & 0 & 0 & 0 \\[2mm]
0 & \frac{1}{68} & 0 & 0 & 0 & 0 & 0 & 0 & 0 & 0 & 0 & 0  & \frac{1}{68}  & \frac{1}{34}  & \frac{1}{68} & \frac{1}{68}  \\[2mm]
0 & 0 & 0  & 0 & 0 & \frac{1}{68} & 0 & 0 & 0 & 0 & 0 & 0 & \frac{1}{68}  &  \frac{3}{68} & 0 & \frac{4}{17}
\end{array}\right) $$

$${\rm Flat}_{\{ \ell_1, \ell_2, \ell_4\}\cup \{ \ell_3, \ell_5, \ell_6, \ell_7\}}(P) =   
\left( \begin{array}{cccccccccccccccc}
 \frac{13}{34} &  \frac{3}{68} & \frac{1}{68}  & \frac{1}{68} & \frac{1}{68} & \frac{1}{68} & 0 & 0 & 0 & 0 & 0 & 0 & 0 & 0 & 0 & 0 \\[2mm]
0 & 0 & 0 & 0 & 0 & 0 & 0 & 0 & 0 & 0 & 0 & 0 & 0 & 0 & 0 & 0 \\[2mm]
0 & 0 & 0 & 0 & 0 & 0 & 0 & 0 & 0 & 0 & 0 & 0 & 0 & 0 & 0 & 0 \\[2mm]
\frac{1}{34} & \frac{1}{68}  & 0 & 0 & 0 & \frac{1}{68}  & 0 & \frac{1}{34} & 0 & 0 & 0 & \frac{1}{68} & 0 & 0 & 0 & 0 \\[2mm]
0 & 0 & 0 & 0 & 0 & 0 & 0 & 0 & 0 & 0 & 0 & 0 & 0 & 0 & 0 & 0 \\[2mm]
0 & 0 & 0 & 0 & 0 & \frac{1}{68}  & 0 & 0 & 0 & 0 & 0 & 0 & 0 & 0 & 0 & 0 \\[2mm]
0 & 0 & 0 & 0 & 0 & 0 & 0 & 0 & 0 & 0 & 0 & 0 & 0 & 0 & \frac{1}{68}  & 0 \\[2mm]
\frac{1}{68} & 0 & 0 & 0 & 0 & \frac{1}{68} &
0 & 0 & 0 & \frac{1}{68} & \frac{1}{34} & 0 &
\frac{1}{68}  &  \frac{3}{68} & \frac{1}{68}  & \frac{4}{17}
\end{array}\right)  $$

$${\rm Flat}_{\{ \ell_1, \ell_2, \ell_5\}\cup \{ \ell_3, \ell_4, \ell_6, \ell_7\}}(P) =  
\left( \begin{array}{cccccccccccccccc}
 \frac{13}{34} &  \frac{3}{68} & \frac{1}{34} & \frac{1}{68}  & \frac{1}{68} & \frac{1}{68}   & 0 & 0 & 0 & 0 & 0 & 0 & \frac{1}{34} & 0 & 0 & 0 \\[2mm]
0 & 0 & 0 & 0 & 0 & 0 & 0 & 0 & 0 & 0 & 0 & 0 & 0 & 0 & 0 & 0 \\[2mm]
0 & 0 & 0 & 0 & 0 & 0 & 0 & 0 & 0 & 0 & 0 & 0 & 0 & 0 & 0 & 0 \\[2mm]
\frac{1}{68} & \frac{1}{68} & 0 & 0 & 0 & \frac{1}{68} & 0  & 0 & 0 & 0 & 0 & \frac{1}{68}  & 0 & 0 & 0 & 0 \\[2mm]
0 & 0 & 0 & 0 & 0 & 0 & 0 & 0 & 0 & 0 & 0 & 0 & 0 & 0 & 0 & 0 \\[2mm]
0 & 0 & 0 & 0 & 0 & \frac{1}{68} & 0 & 0 & 0 & 0 & 0 & 0 & 0 & 0 & 0 & 0 \\[2mm]
0 & 0 & \frac{1}{68}  & 0 & 0 & 0 & 0 & 0 & 0 & 0 & 0 & 0 & 0 & 0 & \frac{1}{34}  & \frac{3}{68} \\[2mm]
0 & 0 & 0 & 0 & 0 & \frac{1}{68}  &
0 & 0 & 0 & \frac{1}{68} & \frac{1}{68} & 0  &
\frac{1}{68} & 0  & \frac{1}{68} & \frac{4}{17}
\end{array}\right)    $$

$${\rm Flat}_{\{ \ell_4, \ell_6, \ell_7\}\cup \{ \ell_1, \ell_2, \ell_3, \ell_5\}}(P) =   
\left( \begin{array}{cccccccccccccccc}
 \frac{13}{34} & 0 & 0 & \frac{3}{68} & \frac{1}{68}  &  0 & 0 & 0 & 0 & 0 & \frac{1}{68}  & 0 & 0 & 0 & 0 & 0 \\[2mm]
\frac{1}{34}  & 0 & 0 & \frac{1}{68} & 0 & \frac{1}{68} & 0 & 0 & 0 & 0 & \frac{1}{68} & 0 & 0 & \frac{1}{68}  & 0 & \frac{1}{68}  \\[2mm]
\frac{1}{68} & 0 & 0 & 0 & 0 & 0 & 0 & 0 & 0 & 0 & 0 & 0 & 0 & 0 & 0 & 0 \\[2mm]
\frac{1}{68} & 0 & 0 & 0 & 0 & 0 & 0 & 0 & 0 & 0 & 0 & 0 & 0 & 0 & 0 & 0 \\[2mm]
0 & 0 & 0 & 0 & 0 & 0 & 0 & 0 & 0 & 0 & 0 & 0 & \frac{1}{68} & 0 & 0 & 0 \\[2mm]
0 & 0 & 0 & \frac{1}{34} & 0 & 0 & 0 & 0 & 0 & 0 & 0 & 0 & \frac{1}{68} & 0 & 0 & \frac{1}{68} \\[2mm]
0 & 0 & 0 & 0 & 0 & 0 & 0 & 0 & 0 & 0 & \frac{1}{68} & 0 & 0 & 0 & 0 & 0 \\[2mm]
0 & 0 & 0 & 0 & 0 & \frac{1}{34}  & 0 & 0 & 0 & 0 & 0 & \frac{3}{68} & \frac{1}{68}  & 0 & 0 & \frac{4}{17}
\end{array}\right)   $$

$${\rm Flat}_{\{ \ell_3, \ell_6, \ell_7\}\cup \{ \ell_1, \ell_2, \ell_4, \ell_5\}}(P) =   
\left( \begin{array}{cccccccccccccccc}
 \frac{13}{34} & 0 & 0 & \frac{1}{34}  & \frac{1}{68} & 0 & 0 & 0 & 0 & 0 & 0 & \frac{1}{68}  & 0 & 0 & 0 & 0 \\[2mm]
 \frac{3}{68} & 0 & 0 & \frac{1}{68}  & \frac{1}{68} & 0 & 0 & 0 & 0 & 0 & \frac{1}{68} & 0 & 0 & \frac{1}{68} & 0 & \frac{1}{68}  \\[2mm]
\frac{1}{68}  & 0 & 0 & 0 & 0 & 0 & 0 & 0 & 0 & 0 & 0 & 0 & 0 & 0 & 0 & 0 \\[2mm]
\frac{1}{68} & 0 & 0 & 0 & 0 & 0 & 0 & 0 & 0 & 0 & 0 & 0 & 0 & 0 & 0 & \frac{1}{68}  \\[2mm]
0 & 0 & 0 & 0 & 0 & 0 & 0 & 0 & 0 & 0 & 0 & \frac{1}{34}   & \frac{1}{68} & 0 & 0 & \frac{1}{68}     \\[2mm]
0 & 0 & 0 & \frac{1}{34}  & 0 & 0 & 0 & 0 & 0 & 0 & 0 & 0 & 0 & 0 & 0 & \frac{1}{68} \\[2mm]
0 & 0 & 0 & 0 & 0 & 0 & 0 & 0 & 0 & 0 & \frac{1}{68}   & 0 & 0 & 0 & 0 & 0 \\[2mm]
0 & 0 & 0 & 0 & 0 & 0 & 0 & 0 & 0 & 0 & 0 & \frac{3}{68} & 0 & 0 & 0 & \frac{4}{17} 
\end{array}\right)     $$

\smallskip
\subsubsection{The trees $T_5$ and $T_6$}

For the two remaining trees we have the flattening matrix
$$ {\rm Flat}_{\{\ell_3,\ell_4,\ell_5 \}\cup \{ \ell_1,\ell_2,\ell_6,\ell_7\}}(P) = 
 \left( \begin{array}{cccccccccccccccc}
 \frac{13}{34} & 0&0&0& \frac{1}{68} & \frac{1}{68} & 0 & 0 \\
 \frac{1}{34} & 0 & 0 & \frac{1}{68} & 0 & 0 & 0 & 0 \\
 \frac{3}{68} & 0 & 0 & 0 & 0 & 0 & 0 & 0 \\
 \frac{1}{68} & 0&0&0&0 & \frac{1}{34} & 0 &0 \\
 \frac{1}{68} & 0 & 0 & 0 & 0 & 0 & 0 & 0 \\
 0 & 0 & 0 & 0 & 0 & 0 & 0 & 0 \\
  \frac{1}{68} & 0 & 0 & 0 & 0 & 0 & 0 & 0 \\ 
  \frac{1}{68} & \frac{1}{68} & 0 & \frac{1}{68} & \frac{1}{68} & 0 & 0 & 0 \\
   0 & 0 & 0 & 0 & 0 & 0 & 0 & 0 \\
    0 & 0 & 0 & 0 & 0 & 0 & 0 & \frac{1}{34} \\
     0 & 0 & 0 & 0 & 0 & 0 & 0 & 0 \\
      0 & 0 & 0 & 0 & 0 & 0 & 0 & \frac{3}{68} \\
       0 & 0 & 0 & 0 & 0 & 0 & 0 & \frac{1}{68} \\
          0 & 0 & 0 & 0 & 0 & 0 & \frac{1}{68} & \frac{1}{68} \\
        0 & 0 & 0 & 0 & 0 & 0 & 0 & 0 \\
         0 & 0 & 0 & 0 & 0 & 0 &  \frac{1}{68} & \frac{4}{17} \\
 \end{array}\right) $$
 and the matrices for the flattenings $F_5$ and $F_6$ given in the Appendix~C.

\smallskip
\subsection{Phylogenetic invariants}

We compute the phylogenetic invariants, using either the $\ell^\infty$ or the $\ell^1$ norm.
This case shows, as observed already in \cite{Casa}, that the $\ell^1$ norm gives more
reliable results than the $\ell^\infty$ norm.

\begin{itemize}
\item For the first tree $T_1(G)$ we consider all $3\times 3$ minors of the flattenings
$$ M_1 = {\rm Flat}_{\{ \ell_5, \ell_6, \ell_7\}\cup \{ \ell_1, \ell_2, \ell_3, \ell_4\}}(P) \ \ \ \text {and } \ \ \ M_2 = {\rm Flat}_{\{ \ell_1, \ell_2, \ell_3\}\cup \{ \ell_4, \ell_5, \ell_6, \ell_7\}}(P)$$
and we obtain
$$ \| \phi_{T_1}(P) \|_{\ell^\infty} = \max_{3\times 3 \text{minors of } M_1,M_2} | \phi(P) | = \frac{13}{4913} $$
$$  \| \phi_{T_1}(P) \|_{\ell^1} =\sum_{3\times 3 \text{minors of } M_1,M_2} | \phi(P) | 
 =\frac{8811}{157216}    $$

\item For the second tree $T_2(G)$ we consider all $3\times 3$ minors of the flattenings
$$ M_1 = {\rm Flat}_{\{ \ell_5, \ell_6, \ell_7\}\cup \{ \ell_1, \ell_2, \ell_3, \ell_4\}}(P) \ \ \ \text{and} \ \ \  
M_3 = {\rm Flat}_{\{ \ell_1, \ell_2, \ell_4\}\cup \{ \ell_3, \ell_5, \ell_6, \ell_7\}}(P)$$
and we obtain
$$ \| \phi_{T_2}(P) \|_{\ell^\infty} = \max_{3\times 3 \text{minors of } M_1,M_3} | \phi(P) | = \frac{13}{4913}  $$
$$  \| \phi_{T_2}(P) \|_{\ell^1} =\sum_{3\times 3 \text{minors of } M_1,M_3} | \phi(P) |  = \frac{7103}{157216} $$

\item For the third tree $T_3(G)$ we consider all $3\times 3$ minors of the flattenings
$$ M_4 = {\rm Flat}_{\{ \ell_1, \ell_2, \ell_5\}\cup \{ \ell_3, \ell_4, \ell_6, \ell_7\}}(P) \ \ \ \text{ and } \ \ \ 
M_5 = {\rm Flat}_{\{ \ell_4, \ell_6, \ell_7\}\cup \{ \ell_1, \ell_2, \ell_3, \ell_5\}}(P)$$
and we obtain
$$ \| \phi_{T_3}(P) \|_{\ell^\infty} = \max_{3\times 3 \text{minors of } M_4,M_5} | \phi(P) | =\frac{13}{4913}   $$
$$  \| \phi_{T_3}(P) \|_{\ell^1} =\sum_{3\times 3 \text{minors of } M_4,M_5} | \phi(P) |  = \frac{5439}{157216} $$

\item For the fourth tree $T_4(G)$ we consider all $3\times 3$ minors of the flattenings
$$ M_4 = {\rm Flat}_{\{ \ell_1, \ell_2, \ell_5\}\cup \{ \ell_3, \ell_4, \ell_6, \ell_7\}}(P) \ \ \ \text{ and } M_6 = {\rm Flat}_{\{ \ell_3, \ell_6, \ell_7\}\cup \{ \ell_1, \ell_2, \ell_4, \ell_5\}}(P)$$
and we obtain
$$ \| \phi_{T_4}(P) \|_{\ell^\infty} = \max_{3\times 3 \text{minors of } M_4,M_6} | \phi(P) | =\frac{13}{4913}   $$
$$  \| \phi_{T_4}(P) \|_{\ell^1} =\sum_{3\times 3 \text{minors of } M_4,M_6} | \phi(P) |  = \frac{5739}{157216} $$

\item For the fifth tree $T_5(G)$ we consider all $3\times 3$ minors of the flattenings
$$ M_7={\rm Flat}_{\{\ell_3,\ell_4,\ell_5 \}\cup \{ \ell_1,\ell_2,\ell_6,\ell_7\}}(P) \ \ \ \text{and} \ \ \ F_5 \ \text{ (as in Appendix C)} $$
and we obtain
$$ \| \phi_{T_5}(P) \|_{\ell^\infty} = \max_{3\times 3 \text{minors of } M_7,F_5} | \phi(P) | = \frac{13}{4913}  $$
$$  \| \phi_{T_5}(P) \|_{\ell^1} =\sum_{3\times 3 \text{minors of } M_7,F_5} | \phi(P) |  = \frac{25}{578}   $$

\item For the sixth tree $T_6(G)$ we consider all $3\times 3$ minors of the flattenings
$$ M_7 ={\rm Flat}_{\{\ell_3,\ell_4,\ell_5 \}\cup \{ \ell_1,\ell_2,\ell_6,\ell_7\}}(P) \ \ \  \text{and}  \ \ \
F_6 \ \text{ (as in Appendix C)} $$
and we obtain
$$ \| \phi_{T_6}(P) \|_{\ell^\infty} = \max_{3\times 3 \text{minors of } M_7,F_6} | \phi(P) | = \frac{207}{78608} $$
$$  \| \phi_{T_6}(P) \|_{\ell^1} =\sum_{3\times 3 \text{minors of } M_7,F_6} | \phi(P) |  = \frac{11795}{314432} $$

\end{itemize}

When we evaluate the minimum among these candidate trees we see that using
the $\ell^\infty$ norm in this case would incorrectly select the tree $T_6(G)$ as
the best candidate, while using the $\ell^1$ norm correctly selects $T_3(G)$
$$ \min_T \| \phi_T(P) \|_{\ell^\infty} =   \frac{207}{78608}   = \| \phi_{T_6}(P) \|_{\ell^\infty} $$
$$ \min_T \| \phi_T(P) \|_{\ell^1} =   \frac{5439}{157216}   = \| \phi_{T_3}(P) \|_{\ell^1} . $$
The $\ell^\infty$ norm also does not distinguish at all between the trees $T_1(G),\ldots, T_5(G)$.

\smallskip
\subsection{Euclidean distance function}

The Euclidean distance lower bound 
estimate can be obtained as in \S \ref{Var2Sec} by replacing the boundary
probability based on the Longobardi data with the one based on SSWL data. We obtain
the following.

\smallskip

The singular value decompositions $\Sigma={\rm diag}(\sigma_k)$ are now of the form 
$$ \begin{array}{rl} \Sigma(M_1)= & (0.38754,  0.24162, 0.36255 \times 10^{-1}, 0.29457 \times 10^{-1},  \\[2mm] & 0.17913 \times 10^{-1},    0.18822 \times 10^{-2}, 0.44554 \times 10^{-3},  0.81454 \times 10^{-18} )
\end{array} $$
$$ \begin{array}{rl} \Sigma(M_2)= & (     
0.38705  , 0.24121 , 0.40755 \times 10^{-1}  , 0.35206 \times 10^{-1} ,      \\[2mm]
& 0.13458 \times 10^{-1} , 0.25922  \times 10^{-17}, 0.30537 \times 10^{-18}  , 0.12727 \times 10^{-32} )
 \end{array} $$
$$ \begin{array}{rl} \Sigma(M_3)= & (
0.38779, 0.24265, 0.37646 \times 10^{-1}  , 0.14679  \times 10^{-1} , \\[2mm] 
& 0.13520  \times 10^{-1}, 0.72298  \times 10^{-17} , 0.10019  \times 10^{-18} , 0.15015 \times 10^{-30}  
) \end{array} $$
$$ \begin{array}{rl} \Sigma(M_4)=  & (
0.38833, 0.23760, 0.54943 \times 10^{-1}, 0.25989 \times 10^{-1} , \\[2mm] 
& 0.11091 \times 10^{-1}, 0.37355 \times 10^{-17} , 0.11876 \times 10^{-18}  , 0.41814 \times 10^{-32}
) \end{array} $$
$$ \begin{array}{rl} \Sigma(M_5)=   & (
0.38730, 0.24267 , 0.35401 \times 10^{-1}, 0.25107 \times 10^{-1}, \\[2mm]
& 0.13409 \times 10^{-1} , 0.10671 \times 10^{-1}, 0.83305 \times 10^{-3}, 0.63417 \times 10^{-18} 
) \end{array} $$
$$ \begin{array}{rl} \Sigma(M_6)= & (
0.38735 , 0.24147, 0.34918 \times 10^{-1} , 0.29212 \times 10^{-1} , \\[2mm] 
& 0.23098 \times 10^{-1} , 0.10765  \times 10^{-1} , 0.17668 \times 10^{-2} , 0.31311 \times 10^{-3} 
) \end{array} $$
$$ \begin{array}{rl}  \Sigma(M_7)= & (
0.38775 , 0.24257 , 0.29048 \times 10^{-1}, 0.26515 \times 10^{-1} , \\[2mm] 
& 0.14181 \times 10^{-1}  , 0.11708 \times 10^{-1}, 0.13047 \times 10^{-2}, 0.60234 \times 10^{-18} 
) \end{array}  $$
$$ \Sigma(F_5)=( 0.38710 , 0.24296 , 0.44347 \times 10^{-1} , 0.15179 \times 10^{-1}) $$
$$ \Sigma(F_6)=( 0.39170, 0.23723, 0.30854 \times 10^{-1}, 0.20237 \times 10^{-1}) $$

One obtains from these the Euclidean distances
$$ d(M_1,\cD_2(8,16))^2=\sigma_3^2+\cdots+\sigma_8^2= 0.25068 \times 10^{-2} $$
$$ d(M_2,\cD_2(8,16))^2=\sigma_3^2+\cdots+\sigma_8^2= 0.30816 \times 10^{-2}$$
$$ d(M_3,\cD_2(8,16))^2=\sigma_3^2+\cdots+\sigma_8^2= 0.18155 \times 10^{-2} $$
$$ d(M_4,\cD_2(8,16))^2=\sigma_3^2+\cdots+\sigma_8^2= 0.38172 \times 10^{-2} $$
$$ d(M_5,\cD_2(8,16))^2=\sigma_3^2+\cdots+\sigma_8^2= 0.21780 \times 10^{-2} $$
$$ d(M_6,\cD_2(8,16))^2=\sigma_3^2+\cdots+\sigma_8^2= 0.27252 \times 10^{-2}$$
$$ d(M_7,\cD_2(8,16))^2=\sigma_3^2+\cdots+\sigma_8^2= 0.18867 \times 10^{-2} $$
$$ d(F_5,\cD_2(4,32))^2=\sigma_3^2+\sigma_4^2 = 0.21971 \times 10^{-2} $$
$$ d(F_6,\cD_2(4,32))^2=\sigma_3^2+\sigma_4^2 = 0.13615 \times 10^{-2}. $$

Thus, we find that, in the case of the SSWL data for these Germanic languages,
the lower bound on the Euclidean distance gives a less reliable answer. While it
correctly excludes the candidates $T_1(G),T_2(G),T_4(G),T_5(G)$, it assigns
the lowest value to the tree $T_6(G)$ rather than to the correct tree $T_3(G)$
selected by the phylogenetic invariants (computed with the $\ell^1$-norm). Thus,
we see here an example where the lower bound is an unreliable predictor of the
actual Euclidean distance. This example confirms the expectation that Longobardi's
LanGeLin data behave better for phylogenetic reconstruction than the SSWL data. 

\smallskip 

A possible explanation for this phenomenon
lies in the fact that, although the list of SSWL variables for this set of languages is longer than the list
of variables in the Longobardi data, there is a high degree of dependency between the SSWL data.
This was also observed in \cite{ParkMa} where the dependencies between SSWL variables were studied
using Kanerva networks. Thus, the actual number of independent variables that contribute to the
boundary distribution may be smaller in the use of the SSWL data. The fact that the languages in
the set $\cS_2(G)$ have a smaller overlap in the regions of the SSWL variables that are uniformly
mapped for all languages, compared to those in the set $\cS_1(G)$ further explains why the $\ell^\infty$-phylogenetic invariants and the Euclidean distance evaluated on the boundary distribution of 
SSWL data correctly identify the best tree in the
$\cS_1(G)$ case but not in the $\cS_2(G)$ case and the $\ell^1$-phylogenetic invariant 
identifies the correct tree in the case of $\cS_2(G)$ only by a small margin. 
We will return to discuss this 
point in \S \ref{RelationsSec} below.

\section{Phylogenetic Algebraic Varieties of the Romance Languages}

We consider here the case of the Romance subfamily of the Indo-European language
family. In particular, we focus of the relative position of the languages $\ell_1=$ Latin,
$\ell_2 =$ Romanian, $\ell_3= $ French, $\ell_4 =$ Italian, $\ell_5 =$ Spanish, and
$\ell_6 =$ Portuguese. We use the combined data of the SSWL and the
Longobardi databases for this phylogenetic analysis, where we retain only those 
features of the SSWL database that are completely mapped for all of these languages.

\smallskip

When run on this set of syntactic data, the PHYLIP phylogenetic program produces a
unique most parsimonious tree candidate, which is given by the tree $T_1$
\begin{center}
\Tree [ . $\ell_1$  [ $\ell_2$ [ $\ell_5$ [ $\ell_6$ [ $\ell_3$  $\ell_4$ ] ] ] ] ]
\end{center}
with the additional linguistic information that $\ell_1$ (Latin) should be considered 
as the root vertex, since the tree produced by PHYLIP is unrooted. There is clearly
a problem with this tree, since the topology one expects based on historical linguistics
is instead given by the tree $T_2$
\begin{center}
 \Tree [ . $\ell_1$ [ $\ell_2$ [ $\ell_4$ [ $\ell_3$ [ $\ell_5$ $\ell_6$ ] ] ] ] ]
\end{center}

\subsection{Flattening matrices of the PHYLIP tree}
There are three flattening matrices associated to the tree $T_1$, given by the
three possible splittings $e_1=\{ \ell_1, \ell_2 \}\cup \{ \ell_3, \ell_4, \ell_5, \ell_6\}$,
$e_2=\{ \ell_1, \ell_2, \ell_5 \}\cup \{ \ell_3, \ell_4, \ell_6 \}$ and $e_3=\{ \ell_1, \ell_2, \ell_5, \ell_6 \}\cup
\{ \ell_3, \ell_4 \}$.
With the boundary probability distribution given by the combined SSWL and Longobardi data, these
are given by 
$$ {\rm Flat}_{T_1, e_1} =
\left( \begin{array}{cccc} 0.2 & 0.0121 & 0.0606 & 0.0121 \\
0 & 0 & 0 & 0.0061 \\
0 & 0 & 0.0061 & 0 \\
0 & 0 & 0.0061 & 0 \\
0 & 0 & 0 & 0.0061 \\
0 & 0 & 0 & 0 \\
0 & 0 & 0 & 0 \\
0 & 0 & 0 & 0.0182 \\
0.0242 & 0 & 0.0182 & 0 \\
0 & 0.0061 & 0 & 0 \\
0 & 0 & 0 & 0 \\
0 & 0 & 0.0061 & 0 \\
0.0061 & 0 & 0 & 0.0061 \\
0 & 0 & 0 & 0 \\
0.0061 & 0 & 0 & 0.0061 \\
0.0364 & 0.1091 & 0.0364 & 0.4121 
\end{array}\right)  
$$

$$ {\rm Flat}_{T_1, e_2} = \left( \begin{array}{cccccccc}
0.2 & 0 & 0.0121 & 0 & 0.0606 & 0 & 0.0121 & 0.0061 \\
0 & 0 & 0 & 0 & 0.0061 & 0.0061 & 0 & 0 \\
0 & 0 & 0 & 0 & 0 & 0 & 0.0061 & 0 \\
0 & 0 & 0 & 0 & 0 & 0 & 0 & 0.0182 \\
0.0242 & 0 & 0 & 0.0061 & 0.0182 & 0 & 0  & 0 \\
0 & 0 & 0 & 0 & 0 & 0.0061 & 0 & 0 \\
0.0061 & 0 & 0 & 0 & 0 & 0 & 0.0061 & 0 \\
0.0061 & 0.0364 & 0 & 0.1091 & 0 & 0.0364 & 0.0061 & 0.4121 
\end{array} \right)
$$
while the third flattening $ {\rm Flat}_{T_1, e_3}$ is given by {\tiny
$$\left( \begin{array}{cccccccccccccccc}
0.2 & 0 & 0.0121 & 0 & 0 & 0 & 0 & 0 & 0.0606 & 0 & 0.0121 & 0.0061 & 0.0061 & 0.0061 & 0 & 0 \\
0 & 0 & 0 & 0 & 0 & 0 & 0 & 0 & 0 & 0 & 0.0061 & 0 & 0 & 0 & 0 & 0.0182 \\
0.0242 & 0 & 0 & 0.0061 & 0 & 0 & 0 & 0 & 0.0182 & 0 & 0 & 0 & 0 & 0.0061 & 0 & 0 \\
0.0061 & 0 & 0 & 0 & 0.0061 & 0.0364 & 0 & 0.1091 & 0 & 0 & 0.0061 & 0 & 0 & 0.0364 & 0.0061 & 0.4121
\end{array} \right)
$$}

\smallskip
\subsection{Flattening matrices of the historically correct tree}
When we consider the linguistically correct tree $T_2$, instead of the tree $T_1$ computed by PHYLIP,
using the same syntactic data for the boundary distribution, we find the flattening matrices 
$$ {\rm Flat}_{T_2,e_1} = \begin{pmatrix}
0.2 & 0 & 0 & 0  \\
0.0121 & 0 & 0 & 0 \\
0.0606 & 0 & 0.0061 & 0.0061 \\
0.0121 & 0.0061 & 0 & 0 \\
0 & 0 & 0 & 0 \\
0 & 0 & 0 & 0 \\
0 & 0 & 0 & 0 \\
0.0061 & 0 & 0 & 0.0182 \\
0.0242 & 0 & 0 & 0 \\
0 & 0.0061 & 0 & 0 \\
0.0182 & 0 & 0 & 0.0061 \\
0 & 0 & 0 & 0 \\
0.0061 & 0 & 0.0061 & 0.0364 \\
0 & 0 & 0 & 0.1091 \\
0 & 0 & 0 & 0.0364 \\
0.0061 & 0 & 0.0061 & 0.4121
\end{pmatrix} $$
$$ {\rm Flat}_{T_2,e_2}=\begin{pmatrix}
0.2 & 0 & 0 & 0 & 0.0242 & 0 & 0 & 0 \\
0.0121 & 0 & 0 & 0 & 0 & 0.0061 & 0 & 0 \\
0.0606 & 0 & 0.0061 & 0.0061 & 0.0182 & 0 & 0 & 0.0061 \\
0.0121 & 0.0061 & 0 & 0 & 0 & 0 & 0 & 0 \\
0 & 0 & 0 & 0 & 0.0061 & 0 & 0.0061 & 0.0364 \\
0 & 0 & 0 & 0 & 0 & 0 & 0 & 0.1091 \\
0 & 0 & 0 & 0 & 0 & 0 & 0 & 0.0364 \\
0.0061 & 0 & 0 & 0.0182 & 0.0061 & 0 & 0.0061 & 0.4121 
\end{pmatrix} $$
and with the third flattening matrix ${\rm Flat}_{T_2,e_3}$ given by {\tiny
$$ \left( \begin{array}{cccccccccccccccc}
0.2 & 0 & 0 & 0 & 0 & 0 & 0 & 0 & 0.0242 & 0 & 0 & 0 & 0.0061 & 0 & 0.0061 & 0.0364 \\
0.0121 & 0 & 0 & 0 & 0 & 0 & 0 & 0 & 0 & 0.0061 & 0 & 0 & 0 & 0 & 0 & 0.1091 \\
0.0606 & 0 & 0.0061 & 0.0061 & 0 & 0 & 0 & 0 & 0.0182 & 0 & 0 & 0.0061 & 0 & 0 & 0 & 0.0364 \\
0.0121 & 0.0061 & 0 & 0 & 0.0061 & 0 & 0 & 0.0182 & 0 & 0 & 0 & 0 & 0.0061 & 0 & 0.0061 & 0.4121
\end{array} \right) $$ }

\smallskip
\subsection{Phylogenetic invariants}

We compare the phylogenetic invariants of these two trees computed with
respect to the $\ell^\infty$ and the $\ell^1$ norm.

\begin{enumerate}
\item from the PHYLIP tree $T_1$ we obtain:
$$ \| \Phi_{T_1}(P) \|_{\ell^\infty}= \max\{ \max_{\substack{\phi \in 3\times 3 \text{ minors} \\\text{ of } {\rm Flat}_{T_1,e_1} }} |\phi(P)|, \, \max_{\substack{\phi \in 3\times 3 \text{ minors} \\\text{ of } {\rm Flat}_{T_1,e_2} } } |\phi(P)| , \max_{\substack{\phi \in 3\times 3 \text{ minors} \\\text{ of } {\rm Flat}_{T_1,e_3} }} |\phi(P)|
\} = 0.89579   \times 10^{-3} $$
$$  \| \Phi_{T_1}(P) \|_{\ell^1}= \sum_{\substack{\phi \in 3\times 3 \text{ minors} \\\text{ of } {\rm Flat}_{T_1,e_1} }} |\phi(P)| \, + \, 
\sum_{\substack{\phi \in 3\times 3 \text{ minors} \\\text{ of } {\rm Flat}_{T_1,e_2} }} |\phi(P)|\, +\,
\sum_{\substack{\phi \in 3\times 3 \text{ minors} \\\text{ of } {\rm Flat}_{T_1,e_3} }} |\phi(P)| =
0.24790  \times 10^{-1} $$

\item for the historically correct tree $T_2$ we find:
$$ \| \Phi_{T_2}(P) \|_{\ell^\infty}= \max\{ \max_{\substack{\phi \in 3\times 3 \text{ minors} \\\text{ of } {\rm Flat}_{T_2,e_1} }} |\phi(P)|, \, \max_{\substack{\phi \in 3\times 3 \text{ minors} \\\text{ of } {\rm Flat}_{T_2,e_2} } } |\phi(P)| , \max_{\substack{\phi \in 3\times 3 \text{ minors} \\\text{ of } {\rm Flat}_{T_2,e_3} }} |\phi(P)|
\} = 0.89579   \times 10^{-3} $$
$$  \| \Phi_{T_2}(P) \|_{\ell^1}= \sum_{\substack{\phi \in 3\times 3 \text{ minors} \\\text{ of } {\rm Flat}_{T_2,e_1} }} |\phi(P)| \, + \, 
\sum_{\substack{\phi \in 3\times 3 \text{ minors} \\\text{ of } {\rm Flat}_{T_2,e_2} }} |\phi(P)|\, +\,
\sum_{\substack{\phi \in 3\times 3 \text{ minors} \\\text{ of } {\rm Flat}_{T_2,e_3} }} |\phi(P)| =
0.22681  \times 10^{-1} $$
\end{enumerate}

Once again we see that the $\ell^1$ norm reliably distinguishes the historically correct tree $T_2$
over the incorrect PHYLIP candidate, while the $\ell^\infty$ norm gives the same result for both
candidate trees and does not help distinguishing them. 

\smallskip
\subsection{Estimate of the Euclidean distance}

We also compute a lower bound estimate on the Euclidean distance. In the case of the
first tree $T_1$ The Euclidean distances of the flattening matrices from the respective
determinantal varieties are given by
$$ D_{1,1} = {\rm dist}({\rm Flat}_{T_1, e_1} , \cD_2(4,16)), \ \ \ \ D_{1,2}= {\rm dist}({\rm Flat}_{T_1, e_2} , \cD_2(8,8)),
\ \ \  D_{1,3}={\rm dist}({\rm Flat}_{T_1, e_3} , \cD_2(16,4)). $$
The singular values of the flattening matrices are given, respectively, by 
$$ \Sigma({\rm Flat}_{T_1, e_1})= (0.4320, 0.2075,0.14766\times 10^{-1}, 0.8211 \times 10^{-2}) $$
while the singular values of ${\rm Flat}_{T_1, e_2}$ are given by 
$$ (0.4299,
0.2115,
0.1390 \times 10^{-1},
0.8586 \times 10^{-2},
0.7806 \times 10^{-2},
0.4896 \times 10^{-2},
0.8464 \times 10^{-3},
0.1867  \times 10^{-3}   )
$$
and
$$ \Sigma({\rm Flat}_{T_1, e_3})= (
0.4299,
0.2118,
0.1332 \times 10^{-1}  ,
0.7593 \times 10^{-2}). $$
Thus, the Euclidean distances are given, respectively, by
$$ D_{1,1}^2 = 0.2854\times 10^{-3} $$
$$ D_{1,2}^2 = 0.3525 \times 10^{-3} $$
$$ D_{1,3}^2 = 0.2351 \times 10^{-3} $$

For the second tree $T_2$ the Euclidean distances of the flattening
matrices to the corresponding determinantal varieties are given by 
$$ D^2_{2,1} = 0.1390 \times 10^{-3}, $$
which is computed using the singular values
$$ \Sigma({\rm Flat}_{T_2,e_1})=( 0.4300,  0.2119,  0.8567 \times 10^{-2},  0.8102 \times 10^{-2}) ,  $$
$$ D^2_{2,2} = 0.3390  \times 10^{-3} $$
computed using the singular values $\Sigma({\rm Flat}_{T_2,e_2})$ given by {\small 
$$  ( 0.4299 , 0.2115, 0.14218 \times 10^{-1} , 
0.6889  \times 10^{-2} , 0.6061 \times 10^{-2}  , 0.6007 \times 10^{-2} , 0.4070 \times 10^{-2} ,
0.7823 \times 10^{-19} ) $$ }
and 
$$ D^2_{2,3} = 0.2854 \times 10^{-3} $$
with singular values 
$$ \Sigma({\rm Flat}_{T_2,e_3}) =( 0.4320, 0.2075 , 0.1477 \times 10^{-1},  0.8211  \times 10^{-2}). $$

Thus if we compare the likelihood of the two models $T_1$ and $T_2$ using the maximum between the
distances as a lower bound for the Euclidean distance to the phylogenetic variety we find 
$$ L_1 = \max \{ D^2_{1,1}, D^2_{1,2}, D^2_{1,3} \} = 0.3525 \times 10^{-3}  $$
$$ L_2 = \max \{ D^2_{2,1}, D^2_{2,2}, D^2_{2,3} \} = 0.3390  \times 10^{-3} , $$
hence $L_2< L_1$, which also favors the historically correct tree $T_2$:
\begin{center}
 \Tree [ .Latin [ Romanian [ Italian [ French [ Spanish Portuguese ] ] ] ] ]
\end{center}

\section{Phylogenetic Algebraic Varieties of the Slavic Languages}

We then consider a set of Slavic languages: $\ell_1=$ Russian, $\ell_2=$ Polish,
$\ell_3=$ Bulgarian, $\ell_4=$ Serb-Croatian, $\ell_5=$ Slovenian, for which
we again use a combination of SSWL and Longobardi data.  The PHYLIP 
most parsimonious trees algorithm produces in this case five candidate trees
when run on this combination of syntactic data. We use additional linguistic
information on where the root vertex should be placed, separating the West-Slavic
branch where Polish resides from the part of the tree that contains both the East-Slavic
branch and the South-Slavic branch.

We see then that the
candidate trees are respectively given by

\begin{center}
$T_1=$ \hbox{\Tree [ $\ell_2$  [ $\ell_1$  [  $\ell_5$  [ $\ell_4$ $\ell_3$ ]] ]]} \ \ \ \ \
$T_2=$ \hbox{\Tree [ $\ell_2$ [ $\ell_1$ [ $\ell_4$ [ $\ell_5$ $\ell_3$ ] ] ] ]} \ \ \ \ \
$T_3=$ \hbox{\Tree [ $\ell_2$ [ $\ell_1$ [ $\ell_3$ [ $\ell_4$ $\ell_5$ ] ] ] ] } 
\end{center}
\begin{center}
$T_4=$ \hbox{\Tree  [[ $\ell_2$ $\ell_3$ ]  [ $\ell_1$ [ $\ell_4$ $\ell_5$  ] ] ]} \ \ \ \ \ 
$T_5=$ \hbox{\Tree  [ $\ell_2$ [ [ $\ell_1$ $\ell_3$ ] [ $\ell_4$ $\ell_5$  ] ] ]}
\end{center}

\begin{enumerate}
\item The first tree $T_1$ incorrectly places Bulgarian in closer proximity to
Serb-Croatian than Slovenian.
\item The second tree $T_2$ has a similar misplacement, with Bulgarian
appearing to be in greater proximity to Slovenian than Serb-Croatian.
\item The third tree $T_3$ correctly places Slovenian and Serb-Croatian in closest proximity,
and it also correctly places Bulgarian in the same South-Slavic subbranch with the pair of 
Slovenian and Serb-Croatian, so it corresponds to the correct tree topology that matches what is
known from historical linguistics.
\item The fourth tree $T_4$ misplaces Bulgarian in the West-Slavic branch 
with Polish instead of placing it in the South-Slavic branch.
\item The fifth tree $T_5$ misplaces Bulgarian in the East-Slavic branch with Russian
instead of placing it in the South-Slavic branch.
\end{enumerate}

\smallskip
\subsection{Flattening matrices}
The flattening matrices for these trees are given by the following
\begin{enumerate}
\item For the tree $T_1$ the flattening matrices are {\small
$$ {\rm Flat}_{T_1,e_1}=\begin{pmatrix} 0.5122 & 0.0 & 0.0122 & 0.0 \\
0.0 & 0.0 & 0.0 & 0.0\\
0.0 & 0.0 & 0.0 & 0.0\\
0.0122 & 0.0 & 0.0 & 0.0610 \\
0.0854  & 0.0 & 0.0 & 0.0 \\
0.0 & 0.0 & 0.0 & 0.0122 \\
0.0 & 0.0 & 0.0 & 0.0 \\
0.0 & 0.0 & 0.0 & 0.3049 \end{pmatrix} $$
$$ {\rm Flat}_{T_1,e_2}=\begin{pmatrix} 
0.5122 & 0.0 & 0.0 & 0.0 & 0.0122 & 0.0 & 0.0 & 0.0 \\
0.0 & 0.0122 & 0.0 & 0.0 & 0.0 & 0.0 & 0.0 & 0.0610 \\
0.0854 & 0.0 & 0.0 & 0.0 & 0.0 & 0.0 & 0.0 & 0.0122 \\
0.0 & 0.0 & 0.0 & 0.0 & 0.0 & 0.0 & 0.0 & 0.3049
\end{pmatrix} $$ }
\item For the tree $T_2$ the flattening matrices are {\small
$$ {\rm Flat}_{T_2,e_1}=\begin{pmatrix}
0.5122 & 0.0 & 0.0122 & 0.0\\
0.0 & 0.0 & 0.0 & 0.0 \\
0.0 & 0.0 & 0.0 & 0.0 \\
0.0122 & 0.0 & 0.0 & 0.0610 \\
0.0854 & 0.0 & 0.0 & 0.0 \\
0.0 & 0.0 & 0.0 & 0.0122 \\
0.0 & 0.0 & 0.0 & 0.0 \\
0.0 & 0.0 & 0.0 & 0.3049
\end{pmatrix} $$
$$ {\rm Flat}_{T_2,e_2}=\begin{pmatrix}
0.5122 & 0.0 & 0.0854 & 0.0 \\
0.0 & 0.0122 & 0.0 & 0.0 \\
0.0 & 0.0 & 0.0 & 0.0 \\
0.0 & 0.0 & 0.0 & 0.0 \\
0.0122 & 0.0 & 0.0 & 0.0 \\
0.0 & 0.0 & 0.0 & 0.0 \\
0.0 & 0.0 & 0.0 & 0.0122 \\
0.0 & 0.0610 & 0.0 & 0.3049
\end{pmatrix} $$ }

\item For the tree $T_3$ the flattening matrices are {\small
$$ {\rm Flat}_{T_3,e_1}=\begin{pmatrix}
0.5122 & 0.0 & 0.0122 & 0.0 \\
0.0 & 0.0 & 0.0 & 0.0 \\
0.0 & 0.0 & 0.0 & 0.0 \\
0.0122 & 0.0 & 0.0 & 0.0610 \\
0.0854 & 0.0 & 0.0 & 0.0 \\
0.0 & 0.0 & 0.0 & 0.0122\\
0.0 & 0.0 & 0.0 & 0.0 \\
0.0 & 0.0 & 0.0 & 0.3049
\end{pmatrix} $$
$$ {\rm Flat}_{T_3,e_2}=\begin{pmatrix}
0.5122 & 0.0 & 0.0122 & 0.0 & 0.0854 & 0.0 & 0.0 & 0.0 \\
0.0 & 0.0 & 0.0 & 0.0 & 0.0 & 0.0 & 0.0 & 0.0122 \\
0.0 & 0.0 & 0.0 & 0.0 & 0.0 & 0.0 & 0.0 & 0.0 \\
0.0122 & 0.0 & 0.0 & 0.0610 & 0.0 & 0.0 & 0.0 & 0.3049
\end{pmatrix} $$ }

\item For the tree $T_4$ the flattening matrices are {\small
$$ {\rm Flat}_{T_4,e_1}=\begin{pmatrix}
0.5122 & 0.0122 & 0.0854 & 0.0\\
0.0 & 0.0 & 0.0 & 0.0 \\
0.0 & 0.0 & 0.0 & 0.0 \\
0.0122 & 0.0 & 0.0 & 0.0 \\
0.0 & 0.0 & 0.0 & 0.0 \\
0.0 & 0.0 & 0.0 & 0.0122 \\
0.0 & 0.0 & 0.0 & 0.0 \\
0.0 & 0.0610 & 0.0 & 0.3049 
\end{pmatrix} $$
$$ {\rm Flat}_{T_4,e_2}=\begin{pmatrix}
0.5122 & 0.0 & 0.0122 & 0.0 & 0.0854 & 0.0 & 0.0 & 0.0 \\
0.0 & 0.0 & 0.0 & 0.0 & 0.0 & 0.0 & 0.0 & 0.0122 \\
0.0 & 0.0 & 0.0 & 0.0 & 0.0 & 0.0 & 0.0 & 0.0 \\
0.0122 & 0.0 & 0.0 & 0.0610 & 0.0 & 0.0 & 0.0 & 0.3049 
\end{pmatrix} $$ }

\item For the tree $T_5$ the flattening matrices are {\small
$$ {\rm Flat}_{T_5,e_1}=\begin{pmatrix}
0.5122 & 0.0 & 0.0854 & 0.0 \\
0.0 & 0.0 & 0.0 & 0.0 \\
0.0 & 0.0 & 0.0 & 0.0 \\
0.0122 & 0.0 & 0.0 & 0.0 \\
0.0122 & 0.0 & 0.0 & 0.0 \\
0.0 & 0.0 & 0.0 & 0.0122 \\
0.0 & 0.0 & 0.0 & 0.0 \\
0.0 & 0.0610 & 0.0 & 0.3049 
\end{pmatrix} $$
$$ {\rm Flat}_{T_5,e_2}=\begin{pmatrix}
0.5122 & 0.0 & 0.0122 & 0.0 & 0.0854 & 0.0 & 0.0 & 0.0 \\
0.0 & 0.0 & 0.0 & 0.0 & 0.0 & 0.0 & 0.0 & 0.0122 \\
0.0 & 0.0 & 0.0 & 0.0 & 0.0 & 0.0 & 0.0 & 0.0 \\
0.0122 & 0.0 & 0.0 & 0.0610 & 0.0 & 0.0 & 0.0 & 0.3049
\end{pmatrix} $$ }
\end{enumerate}

\smallskip
\subsection{Phylogenetic invariants}

When evaluating the phylogenetic invariant 
for the boundary probability distribution given by the combination of the
SSWL and Longobardi data we have the following result
\begin{enumerate}
\item For the tree $T_1$:
$$ \| \Phi_{T_1}(P) \|_{\ell^\infty}= \max\{ \max_{\substack{\phi \in 3\times 3 \text{ minors} \\\text{ of } {\rm Flat}_{T_1,e_1} }} |\phi(P)|, \, \max_{\substack{\phi \in 3\times 3 \text{ minors} \\\text{ of } {\rm Flat}_{T_1,e_2} } } |\phi(P)| \} = 
0.19043 \times 10^{-2} $$
$$  \| \Phi_{T_1}(P) \|_{\ell^1}= \sum_{\substack{\phi \in 3\times 3 \text{ minors} \\\text{ of } {\rm Flat}_{T_1,e_1} }} |\phi(P)| \, + \, 
\sum_{\substack{\phi \in 3\times 3 \text{ minors} \\\text{ of } {\rm Flat}_{T_1,e_2} }} |\phi(P)| = 0.31794  \times 10^{-2} $$
\item For the tree $T_2$:
$$ \| \Phi_{T_2}(P) \|_{\ell^\infty}= \max\{ \max_{\substack{\phi \in 3\times 3 \text{ minors} \\\text{ of } {\rm Flat}_{T_2,e_1} }} |\phi(P)|, \, \max_{\substack{\phi \in 3\times 3 \text{ minors} \\\text{ of } {\rm Flat}_{T_2,e_2} } } |\phi(P)| \} =   0.19043  \times 10^{-2}
 $$
$$  \| \Phi_{T_2}(P) \|_{\ell^1}= \sum_{\substack{\phi \in 3\times 3 \text{ minors} \\\text{ of } {\rm Flat}_{T_2,e_1} }} |\phi(P)| \, + \, 
\sum_{\substack{\phi \in 3\times 3 \text{ minors} \\\text{ of } {\rm Flat}_{T_2,e_2} }} |\phi(P)| = 
0.36582   \times 10^{-2}
$$
\item For the tree $T_3$:
$$ \| \Phi_{T_3}(P) \|_{\ell^\infty}= \max\{ \max_{\substack{\phi \in 3\times 3 \text{ minors} \\\text{ of } {\rm Flat}_{T_3,e_1} }} |\phi(P)|, \, \max_{\substack{\phi \in 3\times 3 \text{ minors} \\\text{ of } {\rm Flat}_{T_3,e_2} } } |\phi(P)| \} =     0.38087 \times 10^{-3}
 $$
$$  \| \Phi_{T_3}(P) \|_{\ell^1}= \sum_{\substack{\phi \in 3\times 3 \text{ minors} \\\text{ of } {\rm Flat}_{T_3,e_1} }} |\phi(P)| \, + \, 
\sum_{\substack{\phi \in 3\times 3 \text{ minors} \\\text{ of } {\rm Flat}_{T_3,e_2} }} |\phi(P)| = 
   0.90864 \times 10^{-3}
$$
\item For the tree $T_4$:
$$ \| \Phi_{T_4}(P) \|_{\ell^\infty}= \max\{ \max_{\substack{\phi \in 3\times 3 \text{ minors} \\\text{ of } {\rm Flat}_{T_4,e_1} }} |\phi(P)|, \, \max_{\substack{\phi \in 3\times 3 \text{ minors} \\\text{ of } {\rm Flat}_{T_4,e_2} } } |\phi(P)| \} =     0.38087  \times 10^{-3}
 $$
$$  \| \Phi_{T_4}(P) \|_{\ell^1}= \sum_{\substack{\phi \in 3\times 3 \text{ minors} \\\text{ of } {\rm Flat}_{T_4,e_1} }} |\phi(P)| \, + \, 
\sum_{\substack{\phi \in 3\times 3 \text{ minors} \\\text{ of } {\rm Flat}_{T_4,e_2} }} |\phi(P)| = 
   0.13621 \times 10^{-2}
$$

\item For the tree $T_5$:
$$ \| \Phi_{T_5}(P) \|_{\ell^\infty}= \max\{ \max_{\substack{\phi \in 3\times 3 \text{ minors} \\\text{ of } {\rm Flat}_{T_5,e_1} }} |\phi(P)|, \, \max_{\substack{\phi \in 3\times 3 \text{ minors} \\\text{ of } {\rm Flat}_{T_5,e_2} } } |\phi(P)| \} =     0.38087 \times 10^{-3}
 $$
$$  \| \Phi_{T_5}(P) \|_{\ell^1}= \sum_{\substack{\phi \in 3\times 3 \text{ minors} \\\text{ of } {\rm Flat}_{T_5,e_1} }} |\phi(P)| \, + \, 
\sum_{\substack{\phi \in 3\times 3 \text{ minors} \\\text{ of } {\rm Flat}_{T_5,e_2} }} |\phi(P)| = 
   0.17175 \times 10^{-2}
$$
\end{enumerate}
For this set of languages we see again, as observed in \cite{Casa}, that
the $\ell^1$ norm is a better test than the $\ell^\infty$ norm for the evaluation of the
phylogenetic invariants. While the $\ell^\infty$ norm does not distinguish between the
trees $T_3$, $T_4$, $T_5$, the $\ell^1$ norm correctly singles out $T_3$ as the 
preferred candidate.

\smallskip
\subsection{Estimates of Euclidean distance}

The matrix $A= {\rm Flat}_{T_1,e_1}= {\rm Flat}_{T_2,e_1}= {\rm Flat}_{T_3,e_1}$ has singular values \newline
$\Sigma(A)=(0.5195, 0.3111, 0.2023\times 10^{-2}, 0.2577\times 10^{-17}, 0, 0, 0, 0)$. \newline The matrix
$B= {\rm Flat}_{T_3,e_2}= {\rm Flat}_{T_4,e_2}= {\rm Flat}_{T_5,e_2}$ has singular values \newline
$\Sigma(B)=(0.5196, 0.3110, 0.2391\times 10^{-2}, 0)$. The remaining matrices have \newline
$\Sigma({\rm Flat}_{T_1,e_2})=(0.5194, 0.3112, 0.1196 \times 10^{-1}, 0.2003 \times 10^{-2})$, \newline
$\Sigma({\rm Flat}_{T_2,e_2})=(0.5194,0.3112,  0.1220 \times 10^{-1},   0.2004\times 10^{-2},0,0,0,0   )$, \newline
$\Sigma({\rm Flat}_{T_4,e_1})=(  0.5195,  0.3111 ,  0.2438\times 10^{-2}, 0.1964\times 10^{-2} , 0,0,0,0  )$, \newline
$\Sigma({\rm Flat}_{T_5,e_1})=(  0.5195  , 0.3111 ,  0.2834 \times 10^{-2} , 0.2390 \times 10^{-2}, 0 ,0,0,0 )$.

The computation of the Euclidean distances then gives 
\begin{enumerate}
\item For the tree $T_1$
$$ {\rm dist}({\rm Flat}_{T_1,e_1},  \cD_2(4,8))^2=\sigma_3^2+\cdots \sigma_8^2 = 0.4094  \times 10^{-5} $$
$$ {\rm dist}({\rm Flat}_{T_1,e_2}, \cD_2(8,4))^2=\sigma_3^2+\sigma_4^2 = 0.1470 \times 10^{-3} $$

\item For the tree $T_2$ 
$$ {\rm dist}({\rm Flat}_{T_2,e_1},  \cD_2(4,8))^2=\sigma_3^2+\cdots \sigma_8^2 = 0.4094 \times 10^{-5} $$
$$ {\rm dist}({\rm Flat}_{T_2,e_2},  \cD_2(4,8))^2=\sigma_3^2+\cdots \sigma_8^2 = 0.1527 \times 10^{-3} $$

\item For the tree $T_3$
$$ {\rm dist}({\rm Flat}_{T_3,e_1},  \cD_2(4,8))^2=\sigma_3^2+\cdots \sigma_8^2 = 0.4094 \times 10^{-5} $$
$$ {\rm dist}({\rm Flat}_{T_3,e_2},  \cD_2(4,8))^2=\sigma_3^2+\cdots \sigma_8^2 = 0.5718 \times 10^{-5}  $$

\item For the tree $T_4$
$$ {\rm dist}({\rm Flat}_{T_4,e_1},  \cD_2(4,8))^2=\sigma_3^2+\cdots \sigma_8^2 =  0.9803 \times 10^{-5} $$
$$ {\rm dist}({\rm Flat}_{T_4,e_2},  \cD_2(4,8))^2=\sigma_3^2+\cdots \sigma_8^2 = 0.5718  \times 10^{-5}  $$

\item For the tree $T_5$
$$ {\rm dist}({\rm Flat}_{T_5,e_1},  \cD_2(4,8))^2=\sigma_3^2+\cdots \sigma_8^2 = 0.1374 \times 10^{-4}  $$
$$ {\rm dist}({\rm Flat}_{T_5,e_2},  \cD_2(4,8))^2=\sigma_3^2+\cdots \sigma_8^2 = 0.5718 \times 10^{-5}  $$
\end{enumerate}

\smallskip

The lower bounds on the Euclidean distance function obtained above indicate 
as preferred candidate the tree $T_3$, which is the correct linguistic tree:
\begin{center}
\Tree [ Polish [ Russian [ Bulgarian [ Serb-Croatian Slovenian ] ] ] ] 
\end{center}

\section{Phylogenetic Algebraic Varieties of the early Indo-European tree}\label{EarlySec}

We now discuss the last phylogenetic problem listed in the Introduction,
namely the early branchings of the Indo-European tree involving
the set of languages Hittite, Tocharian,
Albanian, Armenian, and Greek. We analyze here the difference
between the trees of \cite{Gray} and \cite{RWT}, when seen from the
point of view of Phylogenetic Algebraic Geometry. 

\smallskip
\subsection{Trees and phylogenetic invariants}
Once we restrict our attention to the five languages listed above, 
the trees of \cite{Gray} and \cite{RWT}
that we wish to compare result in the smaller five-leaf trees
\begin{center}
\Tree [ Hittite [ [ Tocharian Armenian ] [ Albanian Greek ] ] ]
\end{center}
for the case computed by \cite{Gray}, and the tree
\begin{center}
\Tree[ Hittite [ Tocharian [ Albanian [ Armenian Greek ] ] ] ]
\end{center}
for the case computed by \cite{RWT}.

\smallskip

Forgetting momentarily the position of the root vertex (which is in both trees adjacent to
the Anatolian branch), we are comparing two trees of the form 
\begin{center}
\includegraphics[scale=0.5]{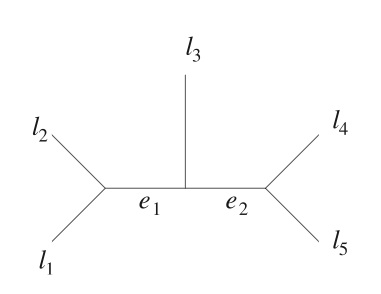}
\includegraphics[scale=0.5]{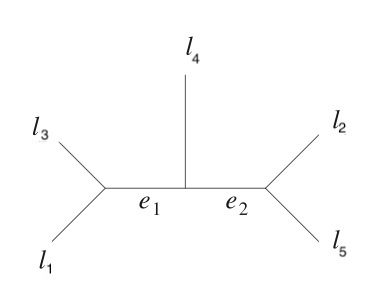}
\end{center}
where we have $\ell_1=$ Tocharian, $\ell_2=$ Armenian, $\ell_3 =$ Hittite,
$\ell_4=$ Albanian, $\ell_5=$ Greek. 

\smallskip

In both cases the flattenings of the tree are given by the matrices
\begin{center}
\includegraphics[scale=0.75]{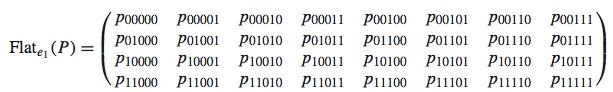}
\includegraphics[scale=0.75]{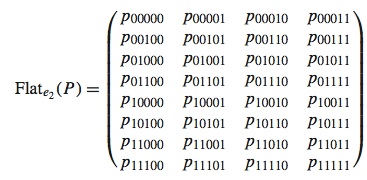}
\end{center}
and the phylogenetic ideal of the tree is generated by all the $3\times 3$ minors
of these two matrices. 

\smallskip

In order to compare the two possibilities then, we evaluate the phylogenetic
invariants (the generators of the phylogenetic ideal obtained in this way) on
the boundary distribution obtained from the data of SSWL variables for the
five languages, distributed in the leaves of the tree in one of the two ways
described above, and we compute the Euclidean distance function. 

\smallskip

\subsection{Syntactic structures and boundary distributions} 

One of the main problems with the SSWL database is that the binary
variables of syntactic structures are very non-uniformly mapped across
languages. In order to use the data for phylogenetic reconstruction, it
is necessary to restrict to only those variables that are completely
mapped for all the languages considered. In our present case, some
of the languages are very poorly mapped in the SSWL database: Tocharian A
is only 19$\%$ mapped, Tocharian B 18$\%$, Hittite is 32$\%$ mapped, 
Albanian 69$\%$, Armenian 89$\%$ and (Ancient) Greek is also 89$\%$
mapped. Moreover, not all the 29 binary syntactic variables that are mapped
for Tocharian A are also among the variables mapped for Hittite. This reduces
the list of syntactic variables that are completely mapped for all five of these
languages to a total of only 22 variables. The variables (listed with the
name used in the SSWL database) and the resulting values are given in the table below. 
Based on these data, the boundary distribution for the two cases considered above
is given by the following. In the first case the frequencies are given by
$$ \begin{array}{ccc} p_{00000}= 4/11, & p_{11111}=3/11, & 
p_{11101} =2/11, \\  p_{11011}=1/22, & p_{10111}=1/11, & p_{01000}=1/22 \end{array}
$$
with $p_{i_1,\ldots, i_5}=0$ for all the remaining binary vectors in $\{ 0,1 \}^5$.
In the second case we have frequencies
$$ \begin{array}{ccc} p_{00000}= 4/11, & p_{11111}=3/11, & 
p_{11011} =2/11, \\  p_{10111}=1/22, & p_{11101}=1/11, & p_{00010}=1/22 \end{array}
$$
with $p_{i_1,\ldots, i_5}=0$ for all the remaining binary vectors in $\{ 0,1 \}^5$.

\begin{center}
 \begin{tabular}{| l ||c|}
\hline 
 P & [Tocharian A, Hittite, Albanian, Armenian, A.Greek]  \\
  \hline 
01 & [1,1,1,1,1]  \\
06 & [1,1,0,1,1] \\
11 &  [1,0,1,1,1]\\
12 &  [1,1,1,1,1] \\
13 &  [1,1,0,1,1] \\
15 & [1,1,1,1,1] \\
17 &  [1,1,1,1,1] \\
19 & [1,1,0,1,1] \\
21 & [1,1,0,1,1] \\
A01 & [1,1,1,0,1] \\
A02  & [1,1,1,0,1] \\
Neg 01 &  [1,1,1,1,1] \\
Neg 03  &  [0,0,0,1,0] \\
Neg 04  &  [0,0,0,0,0] \\
Neg 07 &  [0,0,0,0,0] \\
Neg 08 & [0,0,0,0,0] \\
Neg 09 & [0,0,0,0,0] \\
Neg 10 & [0,0,0,0,0] \\
Neg 12 &  [0,0,0,0,0] \\
Neg 13 &  [0,0,0,0,0] \\
Neg 14 &  [0,0,0,0,0] \\
Order N3 01 &  [1,1,1,1,1] \\
\hline 
\end{tabular}
\end{center}

For the first case, the flattening matrices evaluated at the boundary distribution $P$ give
the matrices
$$ \left(\begin{array}{cccccccc}
\frac{4}{11} & 0 & 0 &  0  &  0  & 0  &  0 & 0    \\ [2mm]
\frac{1}{22}   &  0   &  0  &  0  & 0    &   0 &  0 &     0 \\ [2mm]
           0       &  0  & 0   &  0  &  0  &  0  & 0 &   \frac{1}{11}   \\ [2mm]
 	0	&   0  &  0  &  \frac{1}{22}  &    0 &  \frac{2}{11} &  0 & \frac{3}{11}
\end{array}\right) $$
$$ \left( \begin{array}{cccc}
\frac{4}{11} & 0   &  0  &   0  \\ [2mm]
	0	 &  0  &  0  &   0   \\ [2mm]
\frac{	1}{22}  & 0   &  0  &   0   \\ [2mm]
	0	 & 0   & 0   &  0    \\ [2mm]
	0	 &  0  & 0   &   0   \\ [2mm]
	0	 &  0  &  0  &   \frac{1}{11}  \\ [2mm]
	0	 &  0  &   0 &   \frac{1}{22}   \\ [2mm]
	0	 &  \frac{2}{11}  &  0  &   \frac{3}{11}   
\end{array}\right) $$
For the second case, on the other hand, we obtain 
the matrices
$$ \left(\begin{array}{cccccccc}
\frac{4}{11} & 0   & \frac{1}{22}   &  0   &  0  &    0  &  0  & 0    \\ [2mm]
             0   &  0   &  0  &  0   &  0  &   0  &  0  &  0 \\ [2mm]
             0   &  0  &   0  &  0   &  0  &   0  &  0  &  \frac{1}{22}  \\ [2mm]
 	     0	 &  0  &   0  &  \frac{2}{11}   &  0 &  \frac{1}{11}  &  0  & \frac{3}{11}
\end{array}\right) $$
$$ \left( \begin{array}{cccc}
\frac{4}{11} & 0   &  \frac{1}{22}  &   0  \\ [2mm]
	0	 &  0  &  0  &   0   \\ [2mm]
         0       & 0   &  0  &   0   \\ [2mm]
	0	 & 0   & 0   &  0    \\ [2mm]
	0	 &  0  & 0   &   0   \\ [2mm]
	0	 &  0  &  0  &   \frac{1}{22}  \\ [2mm]
	0	 &  0  &   0 &   \frac{2}{11}  \\ [2mm]
	0	 &  \frac{1}{11}  &  0  &   \frac{3}{11}   
\end{array}\right) $$

\smallskip
\subsection{Phylogenetic invariants}

The evaluation of the phylogenetic invariants on these two boundary distributions 
by evaluating the $3\times 3$ minors of the matrices above gives
\begin{enumerate}
\item For the Gray-Atkins tree $T_1$:
$$ \| \Phi_{T_1}(P) \|_{\ell^\infty} =\max_{\substack{\phi \in 3\times 3 \text{minors} \\ \text{ of flattenings of } T_1}} |\phi(P)| = \frac{8}{1331} $$
$$ \| \Phi_{T_1}(P) \|_{\ell^1} =\sum_{\substack{\phi \in 3\times 3 \text{minors} \\ \text{ of flattenings of } T_1}} |\phi(P)| =\frac{61}{2662} $$
\item For the Ringe--Warnow--Taylor tree $T_2$:
$$ \| \Phi_{T_1}(P) \|_{\ell^\infty} =\max_{\substack{\phi \in 3\times 3 \text{minors} \\ \text{ of flattenings of } T_1}} |\phi(P)| = \frac{8}{1331} $$
$$ \| \Phi_{T_1}(P) \|_{\ell^1} =\sum_{\substack{\phi \in 3\times 3 \text{minors} \\ \text{ of flattenings of } T_1}} |\phi(P)| =\frac{18}{1331} $$
\end{enumerate}

On the basis of this naive test of evaluation of the phylogenetic invariants, the $\ell^\infty$ norm
does not distinguish the two trees while the $\ell^1$ norm prefers the 
Ringe--Warnow--Taylor tree $T_2$. We show below that
this preference is also confirmed by an estimation of the Euclidean distance.

\smallskip
\subsection{Estimate of the Euclidean distance function}

In this case, in order to obtain a lower bound estimate of the Euclidean distance as
an estimate of likelihood of the two trees $T_1$ and $T_2$, we compute the
distances
$$ D_{1,1}= {\rm dist}({\rm Flat}_{e_1,T_1}(P), \cD_2(4,8)), \ \ \  D_{1,2}={\rm dist}({\rm Flat}_{e_2,T_2}(P), \cD_2(8,4)) $$
with the Euclidean distance estimate of $T_1$ given by $L_1=\max\{ D_{1,1}, D_{1,2} \}$ and
$$ D_{2,1}= {\rm dist}({\rm Flat}_{e_1,T_2}(P), \cD_2(4,8)), \ \ \  D_{2,1}={\rm dist}({\rm Flat}_{e_2,T_2}(P), \cD_2(8,4)) $$
with the Euclidean distance estimate of $T_2$ given by $L_2=\max\{ D_{2,1}, D_{2,2} \}$.

\smallskip

The computation of the singular values $\Sigma=(\sigma_1,\ldots, \sigma_4)$ of the flattening 
matrices ${\rm Flat}_{e_i,T_j}(P)$ gives
$$ \Sigma({\rm Flat}_{e_1,T_1}(P))={\rm diag}( 0.3664662612  ,   0.3394847389 ,  0.5018672314 \times 10^{-1}, 0 ) $$
$$ \Sigma({\rm Flat}_{e_2,T_1}(P))={\rm diag}( 0.3664662612, 0.3388120907, 0.5454321492  \times 10^{-1}, 0  ) $$
$$ \Sigma({\rm Flat}_{e_1,T_2}(P))={\rm diag}( 0.3664662613, 0.3421098124 , 0.2700872640  \times 10^{-1}, 0  ) $$
$$ \Sigma({\rm Flat}_{e_2,T_2}(P))={\rm diag}( 0.3664662613 , 0.3394847388 , 0.5018672301 \times 10^{-1}, 0 ). $$
Since the last singular value is always zero, the Euclidean distances are given by the $\sigma_3$ value
$$ D_{1,1}  = 0.5018672314 \times 10^{-1}, \ \ \  D_{1,2} =   0.5454321492  \times 10^{-1}, $$
$$ D_{2,1}  =  0.2700872640  \times 10^{-1}, \ \ \ D_{2,1}  = 0.5018672301 \times 10^{-1} $$
This gives $L_1 =0.5454321492  \times 10^{-1}$ and $L_2= 0.5018672301 \times 10^{-1}$.

\smallskip

Thus, the Euclidean distance estimate also favors the Ringe--Warnow--Taylor tree $T_2$ over the 
Gray-Atkins tree $T_1$. 
The fact that there are very few parameters that are mapped (at present time) for all of these
languages in the SSWL database, and that these parameters largely agree on this set of languages, 
however make this analysis not fully reliable. A more extensive set of syntactic data for these
languages would be needed to confirm whether the phylogenetic reconstruction based on
syntactic data and the algebro-geometric method is reliable. 

\medskip

\section{Towards larger phylogenetic trees: grafting}

As we have seen in the previous sections, Phylogenetic Algebraic Geometry is
a procedure that associates to a given language family
$\cL =\{ \ell_1, \ldots, \ell_n \}$ an algebraic variety $Y=Y(\cL,P)$ constructed
on the basis of the syntactic variables (listed in the distribution $P$). 

\smallskip

A possible geometric viewpoint on comparative historical linguistics can
then be developed, by considering the geometry of the varieties
$Y(\cL,P)$ for different language families. This contains more information
than the topology of the tree by itself, in the sense that one can, for example,
look more specifically for the position of the point $P$ on the variety. The
point $P$ contains precise information on how the binary syntactic
variables change across the languages in the family. For example,
in the case of the six Germanic languages in the set $\cS_1(G)$, 
we see from our table of occurrences that only very few possibilities
for the binary vector $(i_1,\ldots, i_6)$ occur for these six languages.
We also see that, apart from the cases where the value of a syntactic
variable agrees in all six languages (40 occurrences where the feature
is not expressed, and 22 where it is), we find that it is more likely for
Icelandic to have a feature that differs from the other
languages in the group (4 occurrences of $(0,0,0,0,1,0)$ of lacking
a features the others have and 3 occurrences of $(1,1,1,1,0,1)$ for
having a feature that the others lack). Thus, the location of the point $P$
on the variety contains information that is related to the spreading of
syntactic features across the language family considered. This
geometric way of thinking may be compared with the coding theory
approach of \cite{Mar}, \cite{ShuMa2} to measuring the spread of syntactic 
features across a language family. 

\smallskip

As we have seen in the example discussed above of a small set of Germanic
languages, as well as in the examples with Romance and Slavic languages,
the use of SSWL data is suitable for phylogenetic reconstruction, provided only
the subset of the completely mapped syntactic variables (for the given set of languages)
is used and the candidate phylogenetic trees are selected through the computation
of phylogenetic invariants, and their evaluation at the boundary distribution
determined by the syntactic variables. 

\smallskip

This method works very well for small trees and for a set of languages
that is well mapped in the available databases (with enough binary syntactic 
variables that are mapped for all the languages in the given set). However,
one then needs a way to combine phylogenetic trees of smaller subfamilies
into those of larger families. 

\smallskip

In terms of Phylogenetic Algebraic Geometry, this procedure can be articulated as
follows, see \S 5--8 of \cite{AllRho}. 
Given two binary trees $T'$ and $T''$, respectively with $n$ and $m$ leaves, 
the grafting $T=T' \star_\ell T''$ at a leaf $\ell$ is the binary tree obtained
by gluing together a leaf of $T'$ with marking $\ell$ to a leaf of $T''$ with
the same marking. The resulting tree $T$ has $n+m-2$ leaves. It is shown
in \cite{AllRho} how the phylogenetic invariants of $T$ depend on the
invariants of $T'$ and $T''$. Consider the maps $\Phi_{T'}$ and $\Phi_{T''}$,
defined as in \eqref{PhiT} using \eqref{Ppol}, with values in $\C^{2^n}$ and
$\C^{2^m}$, respectively. We identify $\C^{2^n}=\C^{2^{n-1}}\otimes \C^2$,
where the last binary variable corresponds to the leaf $\ell$. We then identify 
the affine space $\C^{2^{n-1}}\otimes \C^2\simeq {\rm Hom}({\C^{2^{n-1}}}^\vee,\C^2)$ 
with the space of matrices $M_{2^{n-1}\times 2}(\C)$, and similarly with
$\C^{2^m}\simeq M_{2\times 2^{m-1}}(\C)$. One then defines $\Phi_T = \Phi_{T'} \star \Phi_{T''}$
as the matrix product of the elements in the range of $\Phi_{T'}$, seen as
matrices in $M_{2^{n-1}\times 2}(\C)$ with the elements in the range of $\Phi_{T''}$,
seen as matrices in $M_{2\times 2^{m-1}}(\C)$. This results in a matrix in $M_{2^{n-1}\times 2^{m-1}}(\C))$, 
which gives a map $\Psi_T$ with values in $\C^{n+m-2}$. The domain variables of $\Psi_T$
are obtained as follows. For those edges of $T$ not involved in the grafting
operation, we define the $2\times 2$ matrices $M^e$ to be the same as those
originally associated to the edges of $T'$ or $T''$, respectively. For the edge of $T'$ 
and the edge of $T''$ that are glued together in the grafting, we
replace the respective matrices $M^{e'}$ and $M^{e''}$ by their product $M^e=M^{e'}M^{e''}$.
Dually, as in \eqref{PsiT}, this
determines the map $\Psi_T$ of polynomial rings, whose kernel is the
phylogenetic ideal of $T$. The closure in $\C^{n+m-2}$ of the image of $\Psi_T$ is
the phylogenetic algebraic variety of the grafted tree $T=T' \star_\ell T''$.

\smallskip

Suppose we are interested in the phylogenetic tree of a language family $\cL$,
for which we assume that we already know (from other linguistic input) a
subdivision into several subfamilies $\cL=\cL_1 \cup \cdots \cup \cL_N$.
Suppose also that for the language families taken into consideration there are sufficient data
available about the ancient languages. (This requirement will limit the applicability of the
algorithm discussed here to families like the Indo-European, where significant amount
of data about ancient languages is available.) We can then follow the following procedure
to graft phylogenetic trees of the subfamilies $\cL_k$ into a larger phylogenetic tree for the
family $\cL$.
\begin{enumerate}
\item For each subfamily $\cL_k=\{ \ell_{k,1}, \ldots, \ell_{k,n_k} \}$, consider the list of 
SSWL data that are completely mapped for all the languages $\ell_{k,j}$ in the 
subfamily $\cL_k$. 
\item On the basis of that set of binary syntactic variables, a
preferred candidate phylogenetic tree $T_k$ is constructed based on the method
illustrated above in the example of the Germanic languages. 
\item Us the procedure discussed in  \S \ref{rootSec} above to identify the
best location of the root vertex for each tree $T_k$, and regard
each tree $T_k$ as a tree with $n_k+1$ leaves, including one leaf attached
to the root vertex. 
\item Let $\{ \lambda_1, \ldots, \lambda_N \}$ be the ancient
languages located at the root vertex of each tree $T_1, \ldots, T_N$. 
Consider the
list of SSWL parameters that are completely mapped for all the
ancient languages $\lambda_k$.
\item On the basis of that set of binary syntactic variables, select 
preferred candidate phylogenetic tree $T$ with $N$ leaves, by
evaluating the phylogenetic invariants of these trees on the
boundary distribution given by this set of binary syntactic variables.
\item Graft the best candidate tree $T$ to the trees $T_k$ by gluing 
the leaf $\lambda_k$ of $T$ to the root of $T_k$.
\item The phylogenetic invariants of the resulting grafted tree $T' = T\star_{k=1}^N T_k$
can be computed with the grafting procedure of \cite{AllRho} described above and
evaluation at the boundary distribution given by the leaves 
$\{ \ell_{k,j}\,|\, j=1,\ldots, n_k, \, k=1,\ldots,N \}$ of $T'$ (coming from the smaller set
of syntactic variables that are completely mapped for all the $\ell_{k,j}$) can
confirm the selected tree topology $T'$.
\end{enumerate}

The advantage of this procedure is that it is going to work even in the absence of 
a sufficient number of binary syntactic variables in the SSWL
database that are completely mapped for all of the languages $\ell_{k,j}$
at the same time, provided there are enough for each subset $\cL_k$
and for the $\lambda_k$. In cases where the number of variables
that are completely mapped for all the $\ell_{k,j}$ is significantly smaller 
compared to those that are mapped within each group, the last test on
the tree $T'$ becomes less significant. This method also 
has the advantage that one works with the
smaller subtrees $T_k$ and $T$, rather than with the bigger tree given by
their grafting, so that the computations of phylogenetic invariants 
is more tractable. 

\smallskip

In the case of language families where one does not have syntactic
data of ancient languages available, one can still adapt the procedure
described above, provided there is a reasonable number of SSWL variables that
are completely mapped for all the languages $\ell_{k,j}$ in $\cL$. One
can proceed as follows.
\begin{enumerate}
\item For each subfamily $\cL_k=\{ \ell_{k,1}, \ldots, \ell_{k,n_k} \}$, consider the list of 
data that are completely mapped for all the languages $\ell_{k,j}$ in the 
subfamily $\cL_k$. 
\item On the basis of that set of binary syntactic variables, a
preferred candidate phylogenetic tree $T_k$ is constructed based on the method
illustrated above in the example of the Germanic languages. 
\item Consider all possible choices of a root vertex for each $T_k$ (there are
as many choices as the number of internal edges of $T_k$).
\item Consider all the possible candidate tree topologies $T$ with $N$ leaves.
\item For each choice of a root vertex in each $T_k$ graft a choice of $T$ to
the give roots of the trees $T_k$ to obtain a candidate tree $T' = T\star_{k=1}^N T_k$.
\item Compute the phylogenetic invariants of $T' = T\star_{k=1}^N T_k$ using the
procedure of \cite{AllRho} recalled above.
\item Evaluate the phylogenetic invariants of each candidate $T'$ on the
boundary distribution determined by the binary syntactic variables that are
completely mapped for all the languages $\{ \ell_{k,j} \,|\, j=1,\ldots, n_k, \, k=1,\ldots,N \}$,
to select the best candidate among the $T'$.
\end{enumerate}

\smallskip

There are serious computational limitations to this procedure, however, 
because of how fast the number of trees on $N$ leaves grows.
While the grafting procedure discussed above makes it possible to
work with smaller trees and then consider the problem of grafting
them into a larger tree, this would still only work computationally
for small size trees, and cannot be expected to handle, for example,
the entire set of languages recorded in the SSWL database.

\medskip
\section{Modifying the setting to account for syntactic relations}\label{RelationsSec}

In a followup to this paper, based on the ongoing analysis of \cite{OBM}, we will
discuss how to adjust these phylogenetic models to incorporate deviations from
the assumption that the syntactic parameters are i.i.d.~random variables evolving 
according to the same Markov model on a tree.

\smallskip

Indeed, we know from various data analysis of the syntactic variables, including
topological data analysis \cite{PortMa1}, \cite{PortMa2}, methods of
coding theory \cite{ShuMa2}, and recoverability in Kanerva networks \cite{ParkMa},
that the syntactic parameters are certainly not i.i.d.~variables. 
Thus, it is likely that some discrepancies we observed in this paper,
in the application of the Phylogenetic Algebraic Geometry method (for example in
the case of the Romance languages or the early Indo-European languages
where the tree selected by the Euclidean distance is not the same as the
tree favored by the phylogenetic invariants) may be an effect of the use of this
overly simplified assumption. 

\smallskip

The approach we plan to follow 
to at least partially correct for this problem, is to modify the boundary distribution
on the tree by attaching to the different syntactic parameters a weight that comes
from some measure of its dependence from other parameters, in such a way that
parameters that are more likely to be dependent variables according to one of these
tests will weight less in the boundary distribution than parameters that are more
likely to be truly statistically independent variables.

\smallskip

The main idea on how to achieve this gola is to modify the boundary distribution 
$P$ by counting occurrences $n_{i_1,\ldots,i_n}$
of parameter values $(i_1,\ldots,i_n)$ at the $n$ leaves of the tree by introducing weights
for different parameters that measure their degree of independence. An example of
such a weight would be the degree of recoverability in a 
Kanerva network, as in \cite{ParkMa}, or a computation of clustering coefficients as in \cite{OBM}.

\smallskip

This means that, instead of assigning to a given binary vector $(i_1,\ldots,i_n)$
the frequency $$p_{i_1,\ldots, i_n}=\frac{n_{i_1,\ldots,i_n}}{N}$$ with $N$ total number of parameters and
$n_{i_1,\ldots,i_n}$ number of parameters that have values $(i_1,\ldots,i_n)$
on the $n$ languages at the leaves of the tree, we replace this by a new distribution
$$ p'_{i_1,\ldots, i_n}=Z^{-1} \sum_{r=1}^{n_{i_1,\ldots,i_n}} w(\pi_r) $$
where for a syntactic parameter $\pi$ the weight $w(\pi)$ measures the degree
of independence of $\pi$, for example with $w(\pi)$ close to $1$ the more $\pi$
can be regarded as an independent variable and close to $0$ the more $\pi$
is recoverable from the other variables, with $Z$ a normalization factor so that
$p'_{i_1,\ldots, i_n}$ is again a probability distribution. 

\smallskip

With this new boundary distribution $P'$ we will recompute the Euclidean
distances of the flattening matrices ${\rm Flat}_e(P')$ from the
varieties $\cD_2(a,b)$ by computing the singular values $(\sigma_1,\ldots,\sigma_a)$
of ${\rm Flat}_e(P')$ and computing the square-distance as $\sigma_3^2+\cdots +\sigma_a^2$,
and compare the new distances obtained in this way with those of the original boundary distribution $P$.

\smallskip

Results on this approach will be presented in forthcoming work.

\newpage 

\section*{Appendix A: SSWL syntactic variables of the set $\cS_1(G)$ of Germanic languages}

We list here the 90 binary syntactic variables of the SSWL  database that are completely
mapped for the six Germanic languages $\ell_1=$Dutch, $\ell_2=$German, $\ell_3=$English, 
 $\ell_4=$Faroese, $\ell_5=$Icelandic, $\ell_6=$Swedish. The column on the left in the tables
 lists the SSWL parameters $P$ as labeled in the database. 

\begin{center}
\includegraphics[scale=0.35]{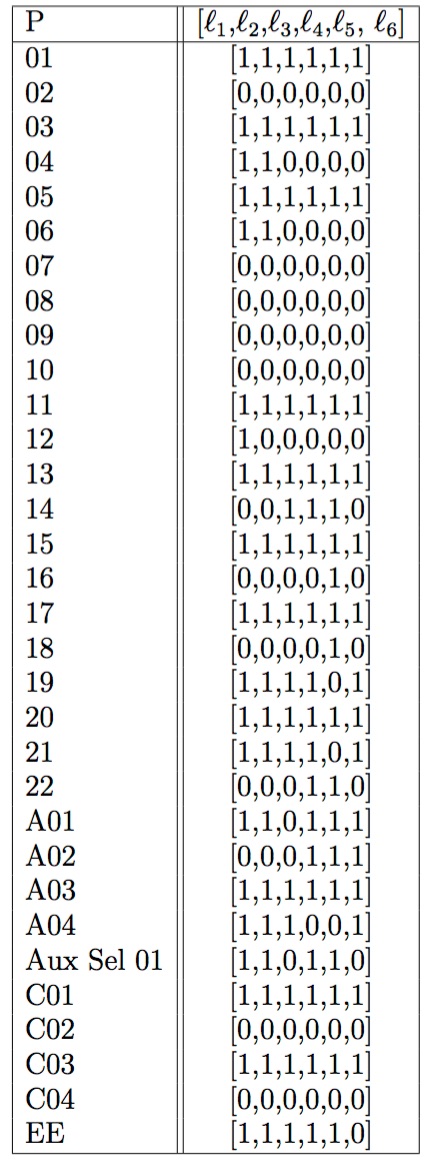}
\includegraphics[scale=0.35]{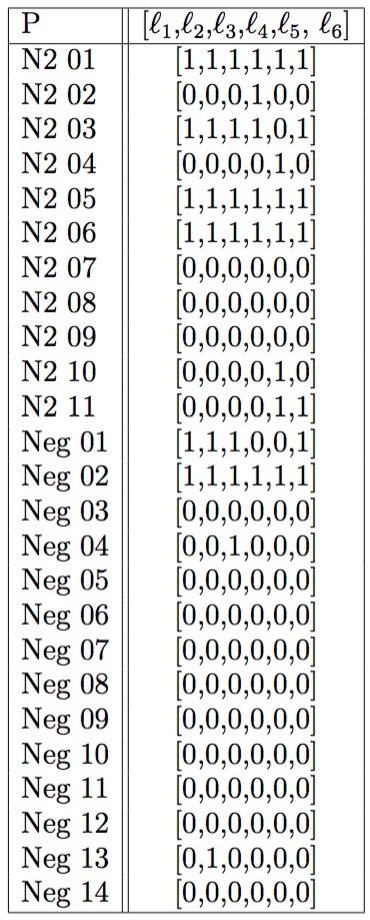}
\includegraphics[scale=0.35]{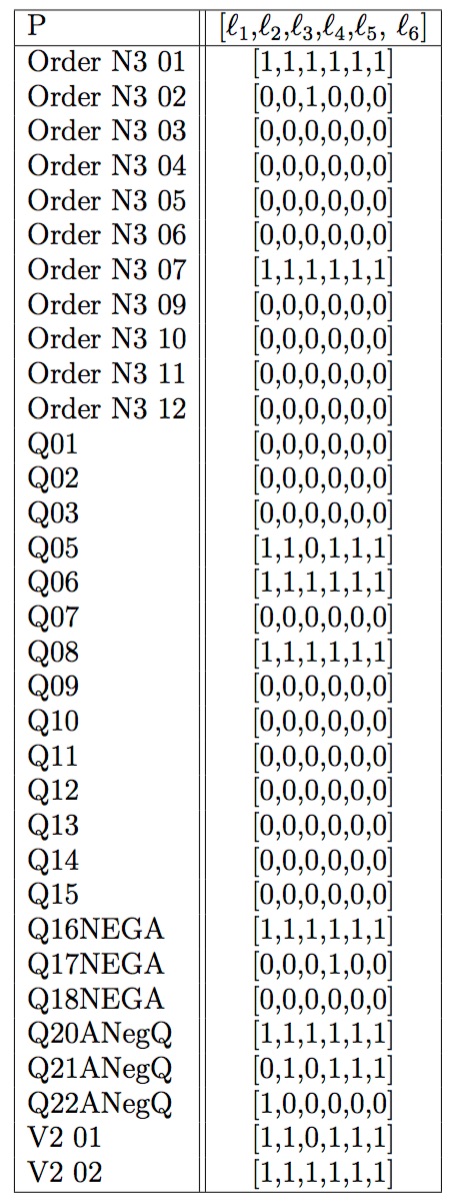}
\end{center}

\newpage 

\section*{Appendix B: SSWL syntactic variables of the set $\cS_2(G)$ of Germanic languages}

We list here the 90 binary syntactic variables of the SSWL  database that are completely
mapped for the seven Germanic languages $\ell_1=$Norwegian, $\ell_2=$Danish, $\ell_3=$Gothic, 
 $\ell_4=$Old English, $\ell_5=$Icelandic, $\ell_6=$English, $\ell_7=$German. The column on the left in the tables
 lists the SSWL parameters $P$ as labeled in the database. 

\begin{center}
\includegraphics[scale=0.2]{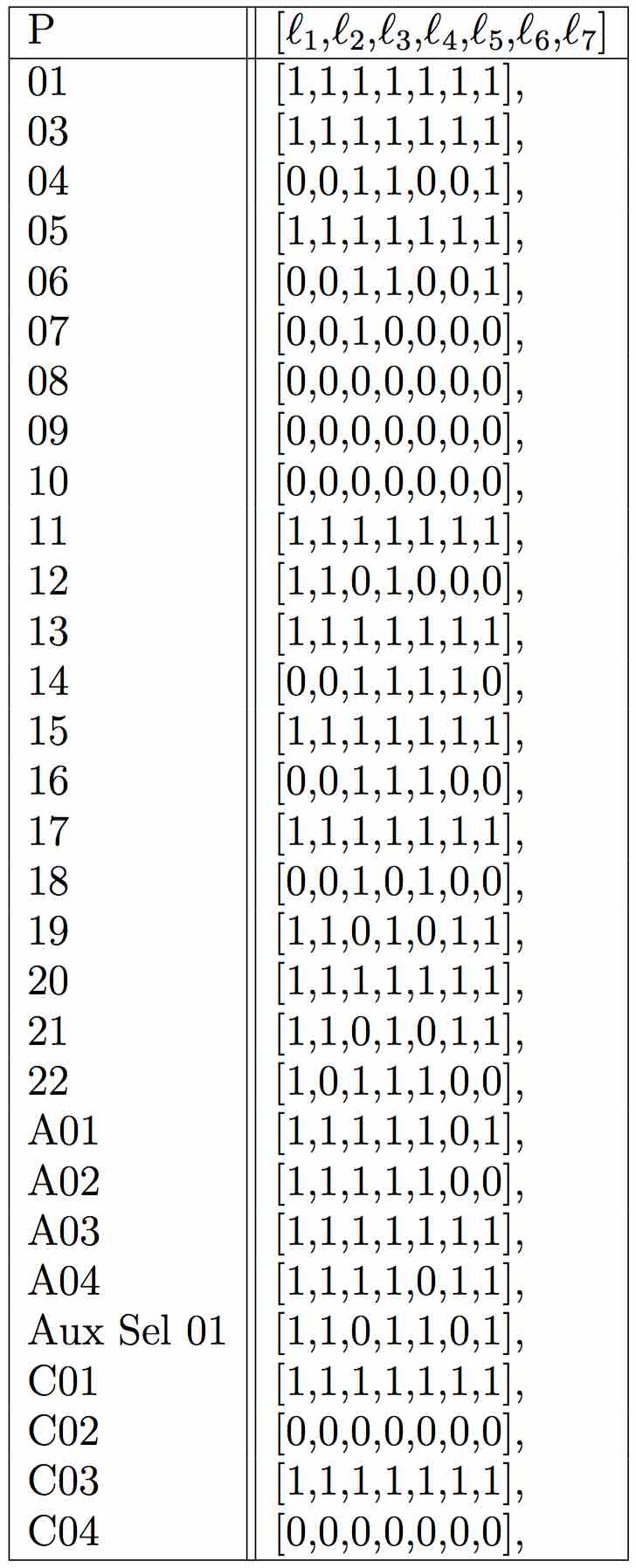}
\includegraphics[scale=0.2]{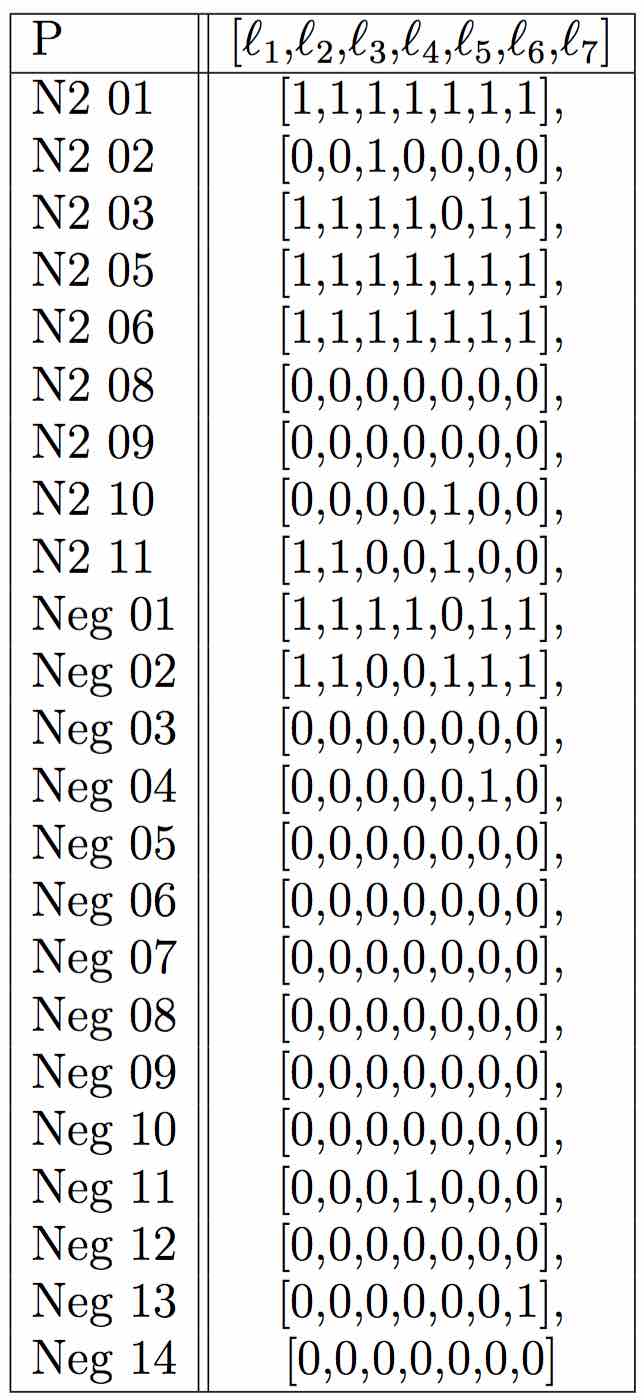}
\includegraphics[scale=0.2]{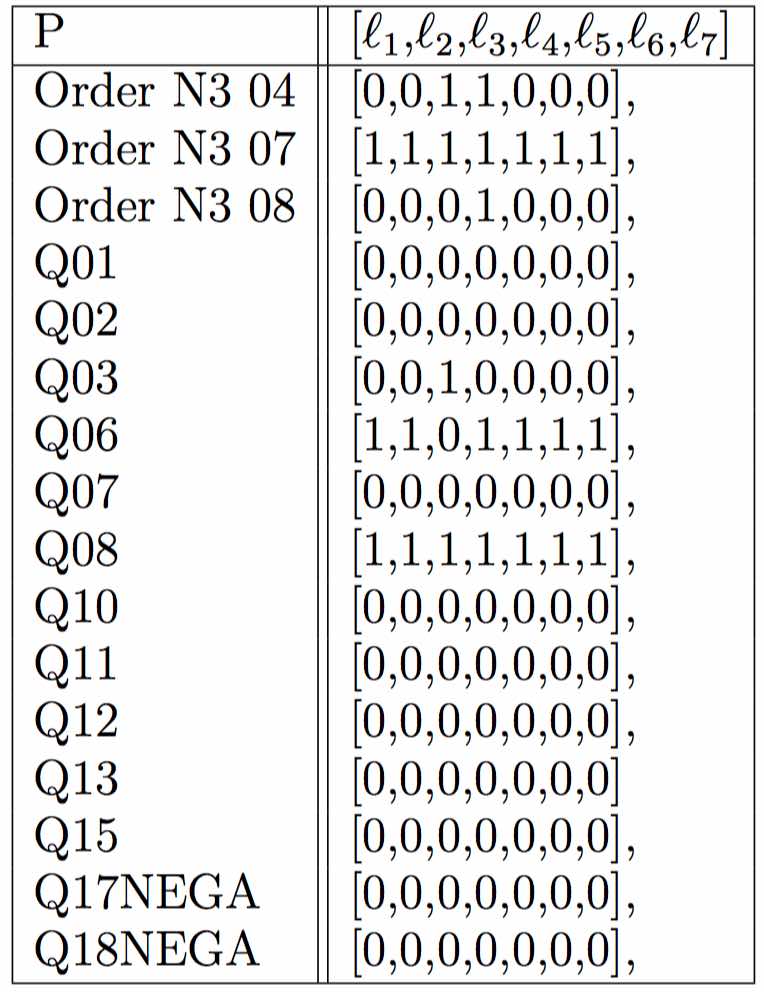}
\end{center}

\newpage
\section*{Appendix C: Flattening matrices $F_5$ and $F_6$}

The remaining flattening matrices (written in transpose form for convenience)
for the $T_5$ and $T_6$ trees, in the case of the Longobardi data are given by
the following:
$$ F_5^t = \left( \begin{array}{cccc} \frac{4}{7} & 0 & 0 & 0 \\
0 & 0 & 0 & 0 \\
0 & 0 & 0 & 0 \\
0 & 0 & 0 & 0 \\
0 & 0 & 0 & 0 \\
0 & 0 & 0 & 0 \\
0 & 0 & 0 & 0 \\
0 & 0 & 0 & \frac{1}{42} \\
0 & 0 & 0 & 0 \\
0 & 0 & 0 & 0 \\
0 & 0 & 0 & 0 \\
0 & 0 & 0 & 0 \\
0 & 0 & 0 & 0 \\
0 & 0 & 0 & 0 \\
0 & 0 & 0 & 0 \\
0 & 0 & 0 & \frac{1}{42} \\
0 & 0 & 0 & 0 \\
0 & 0 & 0 & 0 \\
0 & 0 & 0 & 0 \\
0 & 0 & 0 & 0 \\
0 & 0 & 0 & 0 \\
0 & 0 & 0 & 0 \\
0 & 0 & 0 & 0 \\
0 & 0 & 0 & 0 \\
0 & 0 & 0 & 0 \\
0 & 0 & 0 & 0 \\
0 & 0 & 0 & 0 \\
0 &  \frac{1}{42} & 0 &  \frac{1}{42} \\
0 & 0 & 0 &  \frac{1}{42} \\
0 & 0 & 0 & 0 \\
0 & 0 & 0 & 0 \\
0 & 0 & \frac{1}{42} & \frac{2}{7}
\end{array} \right)  \ \ \ \ \  
F_6^t = \left( \begin{array}{cccc} \frac{4}{7} & 0 & 0 & 0 \\
0 & 0 & 0 & 0 \\
0 & 0 & 0 & 0 \\
0 & 0 & 0 & 0 \\
0 & 0 & 0 & 0 \\
0 & 0 & 0 & 0 \\
0 & 0 & 0 & 0 \\
0 & 0 & 0 & \frac{1}{42} \\
0 & 0 & 0 & 0 \\
0 & 0 & 0 & 0 \\
0 & 0 & 0 & 0 \\
0 & 0 & 0 & 0 \\
0 & 0 & 0 & 0 \\
0 & 0 & 0 & 0 \\
0 & 0 & 0 & 0 \\
0 & 0 & 0 & \frac{1}{42} \\
0 & 0 & 0 & 0 \\
0 & 0 & 0 & 0 \\
0 & 0 & 0 & 0 \\
0 & 0 & 0 & 0 \\
0 & 0 & 0 & 0 \\
0 & 0 & 0 & 0 \\
0 & 0 & 0 & 0 \\
0 & 0 & 0 & 0 \\
0 & 0 & 0 & 0 \\
0 & 0 & 0 & 0 \\
0 & 0 & 0 & 0 \\
0 & \frac{1}{42} & 0 & 0 \\
0 & 0 & 0 & \frac{1}{42} \\
0 & 0 & 0 & 0 \\
0 & 0 & 0 & 0 \\
0 & \frac{1}{42} & \frac{1}{42} & \frac{2}{7}
\end{array} \right) $$

\newpage

The same flattening matrices for the SSWL data are given by the following.
$$ F_5^t = \left( \begin{array}{cccc} 
\frac{13}{34} & \frac{1}{68} & \frac{1}{34} & 0 \\
0 & 0 & 0 & 0 \\
0 & 0 & 0 & 0 \\
0 & 0 & \frac{1}{68} & 0 \\
\frac{3}{68} & \frac{1}{68} & \frac{1}{68} & \frac{1}{68} \\
0 & 0 & 0 & 0 \\
0 & 0 & 0 & \frac{1}{68} \\
0 & 0 & 0 & \frac{1}{68} \\
\frac{1}{68} & 0&0&0 \\
0 & 0 & 0 & 0 \\
0 & 0 & 0 & 0 \\
0 & 0 & 0 & \frac{1}{68} \\
0 & 0 & \frac{1}{34} & 0 \\
0 & 0 & 0 & 0 \\
0 & 0 & 0 & 0 \\
0 & 0 & 0 & \frac{1}{68} \\
\frac{1}{68} & 0&0&0 \\
0 & 0 & 0 & 0 \\
0 & 0 & 0 & 0 \\
0 & 0 & 0 & 0 \\
0 & 0 & 0 & \frac{1}{68} \\
0 & 0 & 0 & 0 \\
0 & 0 & 0 & 0 \\
0 & 0 & 0 & 0 \\
0 & 0 & 0 & 0 \\
0 & 0 & 0 & 0 \\
0 & 0 & 0 & 0 \\
0 & \frac{1}{68} & \frac{1}{34} & \frac{1}{68} \\
0 & 0 & 0 & 0 \\
0 & 0 & 0 & 0 \\
0 & 0 & 0 & 0 \\
0 & 0 & \frac{3}{68} & \frac{4}{17} 
\end{array} \right) \ \ \ \ \
 F_6^t = \left( \begin{array}{cccc} 
\frac{13}{34} & \frac{1}{68} & \frac{3}{68} & \frac{1}{68} \\
0 & 0 & 0 & 0 \\
0 & 0 & 0 & 0 \\
0 & 0 & 0 & 0 \\
\frac{1}{34} & 0 & \frac{1}{68} & \frac{1}{68} \\
0 & 0 & 0 & 0 \\
0 & 0 & 0 & \frac{1}{68} \\
\frac{1}{68} & 0 & 0 & \frac{1}{68} \\
\frac{1}{68} & 0&0&0 \\
0 & 0 & 0 & 0 \\
0 & 0 & 0 & 0 \\
0 & 0 & 0 & 0 \\
0 & 0 & \frac{1}{34} & 0 \\
0 & 0 & 0 & 0 \\
0 & 0 & 0 & 0 \\
0 & \frac{1}{68} & 0 & \frac{1}{68} \\
\frac{1}{68} & 0&0&0 \\
0 & 0 & 0 & 0 \\
0 & 0 & 0 & 0 \\
0 & 0 & 0 & 0 \\
0 & 0 & 0 & \frac{1}{68} \\
0 & 0 & 0 & 0 \\
0 & 0 & 0 & 0 \\
0 & 0 & 0 & 0 \\
0 & 0 & 0 & 0 \\
0 & 0 & 0 & 0 \\
0 & 0 & 0 & 0 \\
0 & \frac{1}{68} & 0 & 0 \\
0 & 0 & 0 & 0 \\
0 & 0 & 0 & 0 \\
0 & 0 & 0 & 0 \\
\frac{1}{34} & \frac{1}{68} & \frac{3}{68} & \frac{4}{17}
\end{array} \right) $$
\newpage

\section*{Appendix D: list of LanGeLin syntactic parameters}

\begin{center}
\includegraphics[scale=0.65]{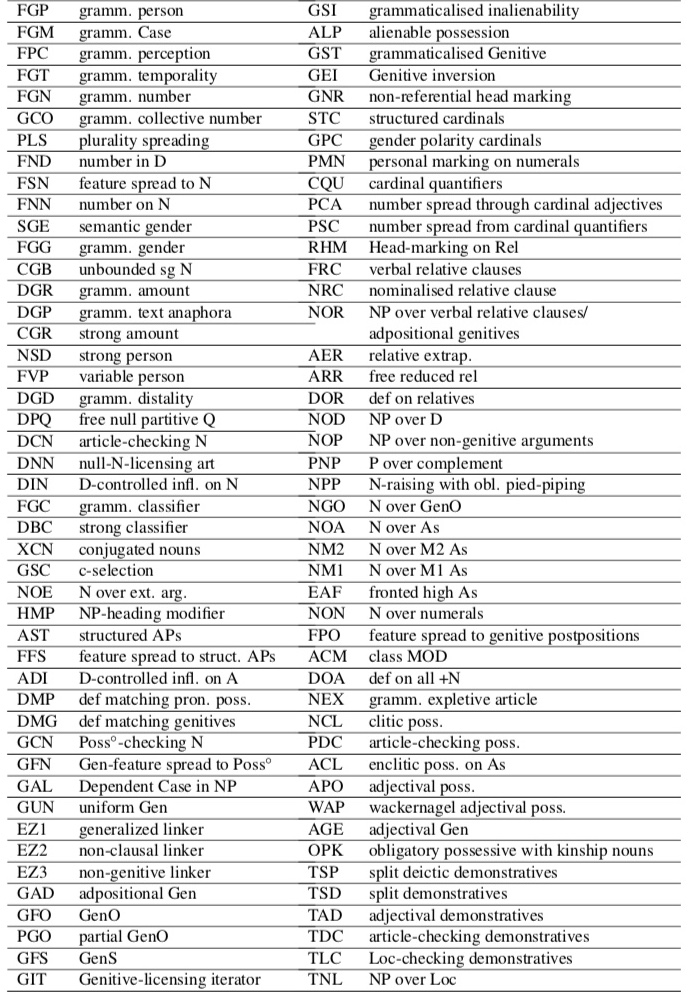}
\end{center}

\newpage

\subsection*{Acknowledgment} The first and second author were partially supported by a
Summer Undergraduate Research Fellowship at Caltech. The last author is partially supported by
NSF grant DMS-1707882, NSERC Discovery Grant RGPIN-2018-04937, Accelerator Supplement grant RGPAS-2018-522593, and by the Perimeter Institute for Theoretical Physics. We are very grateful to
the two anonymous referees for many very useful comments, corrections, and suggestions that greatly improved the paper.

\newpage

\end{document}